\documentclass{article}

\usepackage[nonatbib, preprint]{neurips_2025}

\usepackage[utf8]{inputenc} 
\usepackage[T1]{fontenc}    
\usepackage{hyperref}       
\usepackage{url}            
\usepackage{booktabs}       
\usepackage{amsfonts}       
\usepackage{nicefrac}       
\usepackage{microtype}      
\usepackage{xcolor}         

\usepackage{amsmath}
\newtheorem{theorem}{Theorem}[section]
\newtheorem{remark}[theorem]{Remark}
\newtheorem{lemma}[theorem]{Lemma}
\newtheorem{corollary}[theorem]{Corollary}
\newtheorem{assumption}[theorem]{Assumption}
\newtheorem{definition}[theorem]{Definition}

\usepackage{enumitem}
\setlist[itemize]{noitemsep}
\usepackage{subfig}
\usepackage{graphicx,sidecap}
\usepackage{wrapfig}

\DeclareMathOperator{\Tr}{Tr}
\DeclareMathOperator*{\argmin}{argmin}

\renewcommand{\vec}[1]{\mathbf{#1}}
\newcommand{\mat}[1]{\mathbf{#1}}
\usepackage{dsfont}
\usepackage{float}
\usepackage[
backend=biber,
style=numeric,
]{biblatex}
\title{\textbf{Impact of Bottleneck Layers and Skip Connections on the Generalization of Linear Denoising Autoencoders}}

\author{%
  Jonghyun Ham\thanks{\texttt{Contact: hamj@informatik.uni-freiburg.de}} \\
  University of Freiburg\\
  \And
  Maximilian Fleissner \\
  Technical University of Munich \\
  \AND
  Debarghya Ghoshdastidar \\
  Technical University of Munich \\
}

\begin{document}

\maketitle

\begin{abstract}
Modern deep neural networks exhibit strong generalization even in highly overparameterized regimes. Significant progress has been made to understand this phenomenon in the context of supervised learning, but for unsupervised tasks such as denoising, several open questions remain. While some recent works have successfully characterized the test error of the linear denoising problem, they are limited to linear models (one-layer network). 
In this work, we focus on two-layer linear denoising autoencoders trained under gradient flow, incorporating two key ingredients of modern deep learning architectures: A low-dimensional bottleneck layer that effectively enforces a rank constraint on the learned solution, as well as the possibility of a skip connection that bypasses the bottleneck. We derive closed-form expressions for all critical points of this model under product regularization, and in particular describe its global minimizer under the minimum-norm principle. From there, we derive the test
risk formula in the overparameterized regime, both for models with and without skip connections. 
Our analysis reveals two interesting phenomena: Firstly, the bottleneck layer introduces an additional complexity measure akin to the classical bias--variance trade-off---increasing the bottleneck width reduces bias but introduces variance, and vice versa. Secondly, skip connection can mitigate the variance in denoising autoencoders---especially when the model is mildly overparameterized. 
We further analyze the impact of skip connections in denoising autoencoder using random matrix theory and support our claims with numerical evidence.
\end{abstract}

\section{Introduction}

Despite having a large number of parameters and achieving nearly zero training error, modern neural networks generalize remarkably well to unseen data. This phenomenon, often referred to as \textit{benign overfitting} \cite{Bartlett_2020}, challenges the classical understanding of generalization characterized by a U-shaped risk curve, where increasing model complexity is expected to eventually harm test performance. Extensive theoretical efforts have sought to explain this behavior---albeit almost exclusively in supervised learning. In contrast, little attention has been paid to understanding generalization in \textit{unsupervised learning}, where contradictory statements are made based on numerical studies \cite{jayalath2023no,rahimi2024multiple}. 

A prominent example is the denoising autoencoder (DAE) \cite{vincent2010stacked}. Despite its distinct setting---which differs significantly from standard regression in that \textit{noise is added to the input} rather than the output---and its widespread use in unsupervised representation learning, the generalization properties of DAEs remain underexplored.
A pioneering study by \cite{sonthalia2022training} initiated the theoretical analysis of the denoising problem under a rank-1 data setting. This work was later extended to general low-rank data by \cite{kausik2024doubledescentoverfittingnoisy}. \textit{However, both works are confined to single-layer linear architectures}, limiting their applicability to modern neural networks. More specifically, a notable architectural feature of contemporary neural networks is the presence of \textit{bottleneck structures}, where intermediate layers have significantly lower dimensionality than the overall parameter count. This form of architectural complexity is not adequately addressed by existing theory,  which tends to treat input dimensionality as the \textit{only} measure of model complexity. Our goal is to investigate the role of bottleneck layers as a complementary complexity measure and examine their impact on generalization behavior, specifically in the context of DAEs.

\begin{SCfigure}[50]
\includegraphics[scale=0.25]{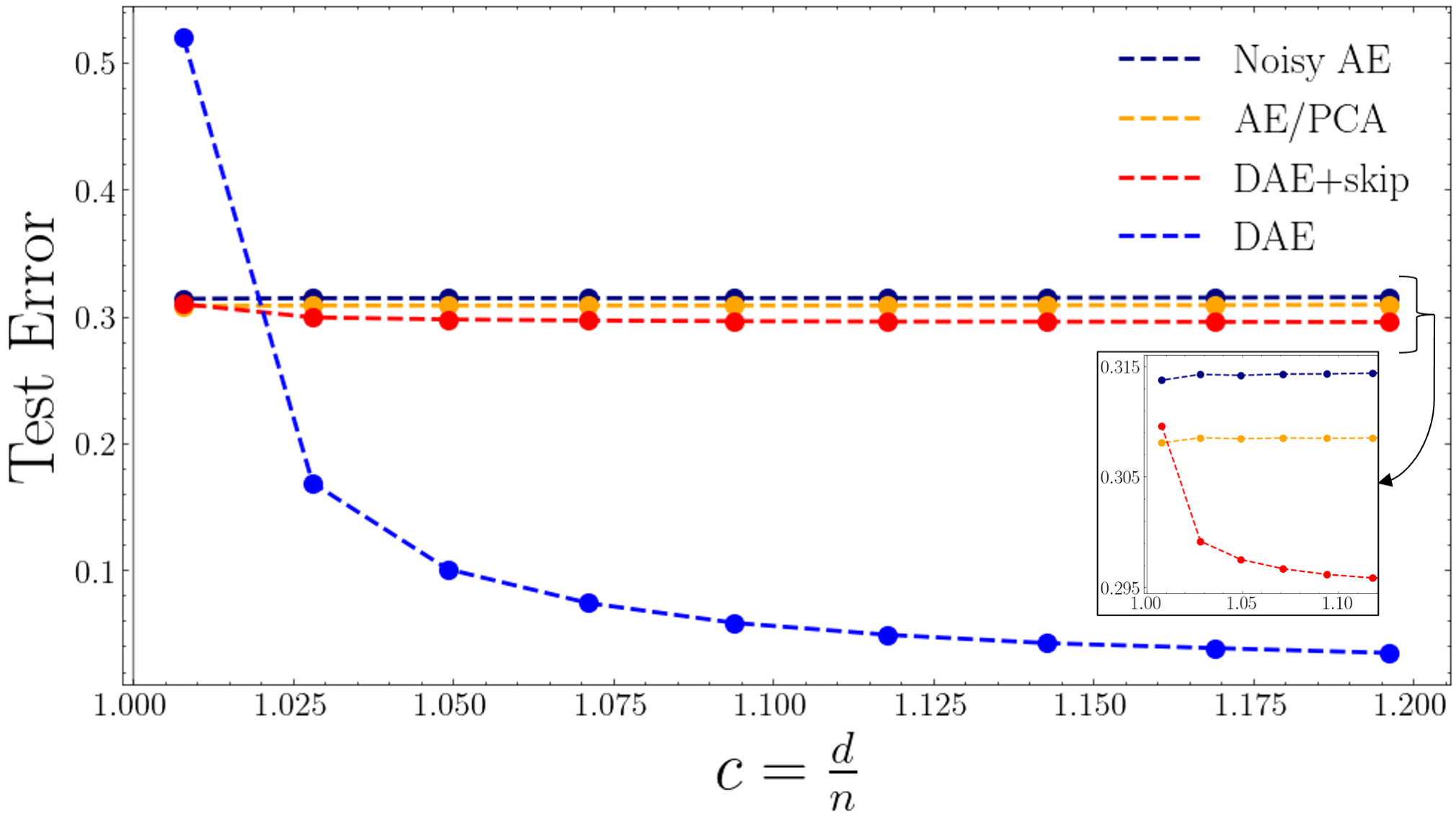}
\caption{
Test error curves for variants of linear autoencoders (AE). 
Unsupervised \textit{AE} learns to reconstruct the input and is equivalent to PCA.
\textit{Noisy AE} maps clean input to a noisy version \cite{adlam22a}. 
We study the generalization error in \textit{denoising AE (DAE)} \cite{vincent2010stacked} that  reconstructs clean samples from noisy version, and \textit{DAE with skip connection} \cite{ronneberger2015unetconvolutionalnetworksbiomedical} that implicitly learns to generate pure noise from noisy data. 
While the generalization error of AE and noisy AE are not affected by over-parameterisation (in the linear case), linear DAE exhibits pronounced peak near $c \approx 1$, which is partly dampened in DAE with skip connection.
}\label{fig:compare_test_risk}
\end{SCfigure}

In general, there is little understanding of how the training dynamics in the presence of bottleneck layers influence the generalization error of neural networks. Several works study random feature models for noisy autoencoders \cite{adlam22a} or investigate the effect of principal component analysis (PCA)-based dimensionality reduction in linear (one-layer) regression settings \cite{Teresa_2022, pmlr-v235-gedon24a, curth2023a}. With dimensionality reduction, the number of retained principal components acts as a form of regularization and an appropriate small dimension suppresses the double descent phenomenon. In contrast, 
fully trained architectures with a bottleneck layer can still exhibit double descent behavior as illustrated in the generalization error curves of over-parameterized two-layer linear DAEs in Figure \ref{fig:compare_test_risk}.

In addition, practical implementations of DAE (such as the U-Net architecture \cite{ronneberger2015unetconvolutionalnetworksbiomedical} that serves as a de facto denoiser in diffusion models \cite{ho2020denoisingdiffusionprobabilisticmodels}) routinely incorporate skip connections as a core architectural feature. While skip connections are widely acknowledged for enhancing training stability by mitigating vanishing or exploding gradients \cite{he2015deepresiduallearningimage}, their impact on test error remains poorly understood. \cite{cui2023high} investigate the role of skip connections in improving generalization performance in undercomplete DAEs using two-layer nonlinear neural networks. However, their analysis is conducted under restrictive assumptions, including Gaussian input data and tied weights. Our work extends the more realistic data assumptions used in \cite{kausik2024doubledescentoverfittingnoisy} and does not impose tied-weight constraints.

To this end, we analyze linear DAEs modeled as two-layer linear networks in the high-dimensional regime, where the input dimension $d$ exceeds the number of training samples $n$. Our model includes a low-dimensional bottleneck layer of size $k \ll n < d$ with or without a skip connection. To address the effect of bottleneck layers and skip connections, we extend the theoretical frameworks for DAEs \cite{sonthalia2022training, kausik2024doubledescentoverfittingnoisy}, which are built upon the assumption of low rank data, and derive analytical expressions that characterize the generalization error in these settings. Moreover, we offer a deeper theoretical understanding of the role of skip connections by explicitly performing a bias--variance decomposition, which was absent in previous studies. Our contributions are summarized as follows.
\begin{enumerate} [left=1em]
    \item In Section \ref{section:preliminaries}, we obtain closed-form expressions for the critical points and global minimizers of linear DAEs with bottleneck layers, both with and without skip connections, under a reconstruction loss with product regularization. Furthermore, we derive the minimum-norm solution, and find that it is approached by the global minimizer of the regularized loss in the ridgeless limit.
    \item In Section \ref{section:generalization}, we leverage the closed-form expressions obtained for the learned model to derive expressions for the test risk. To better understand this result, we perform a bias–variance decomposition that quantifies how the bottleneck dimension influences generalization. In particular, increasing the bottleneck dimension reduces bias but increases variance, and vice versa. While this trade-off echoes the classical interpretation of the bias--variance relationship, it arises within the modern high-dimensional regime, by virtue of the bottleneck layer.
    \item We extend our analysis of the test risk to DAEs with skip connections. Notably, this results in a significantly smoother variance curve as a function of $d/n$, compared to the model without skip connections. The effect is particularly pronounced when the model is only mildly overparameterized.
    \item In Section \ref{section:rmt}, we provide further insights into the origin of the smoother variance for models with skip connections, using tools from random matrix theory. By analyzing a slightly simplified model, we uncover the origin of the smoother variance curve induced by skip connections.
\end{enumerate}

\section{Setting}\label{section:preliminaries}

We begin by introducing the denoising autoencoders(DAEs) and specifying their associated loss functions. We then characterize the solutions learned under gradient flow by deriving all critical points and providing an explicit expression for the global minimizer. This analysis lays the groundwork for our subsequent study of generalization error in Section~\ref{section:generalization}.

\subsection{Training Setup}\label{subsection:training_setup}

We consider two variants of two-layer linear networks. The first model contains the bottleneck structure, and does not include skip connections. Given an input matrix $\mat{Z} \in \mathbb{R}^{d \times n}$, where $n$ denotes the number of training samples and each column of $\mat{Z}$ represents a $d$-dimensional data point, the model is defined by two weight matrices: $\mat{W}_1 \in \mathbb{R}^{k \times d}$ (the encoder) and $\mat{W}_2 \in \mathbb{R}^{d \times k}$ (the decoder). The output of this model is given by $\mat{W_2}\mat{W_1}\mat{Z}$.
The second model includes a skip connection. Its output is defined as
$(\mat{W_2}\mat{W_1} + \mathbb{I})\mat{Z}$,
where the skip connection is implemented as an identity map $\mathbb{I} \in \mathbb{R}^{d \times d}$, directly linking the input to the output layer and bypassing the trainable weight matrices.

\paragraph{Loss Functions for the Denoising Setup}

In the denoising setting, $\mat{X} \in \mathbb{R}^{d \times n}$ denotes the clean input data matrix, and $\mat{A} \in \mathbb{R}^{d \times n}$ is the noise matrix. Then, the input to the network is the corrupted matrix $\mat{X + A}$, while the target output is the clean matrix $\mat{X}$. The networks with and without skip connection are trained to minimize the reconstruction loss of $\mat{X}$ from $\mat{X+A}$. Additionally, a  product regularization term \cite{pretorius2018learning} with regularization strength $\lambda$ is added to the loss function. For the model without skip connections, the loss function $\mathcal{L}_{\text{DAE}}$ is given by
\begin{equation} \label{eq:loss_dae}
\mathcal{L}_{\text{DAE}}(\mat{W}_2, \mat{W}_1, \lambda) := \frac{1}{n}\|\mat{X} - \mat{W}_2 \mat{W}_1(\mat{X + A})\|_F^2 + \lambda \| \mat{W}_2 \mat{W}_1 \|_F^2.
\end{equation}
For the model with skip connections, the corresponding loss function $\mathcal{L}_{\text{DAE+SC}}$ is given by
\begin{align}
\label{eq:loss_dae_skip}
\mathcal{L}_{\text{DAE+SC}}(\mat{W}_2, \mat{W}_1, \lambda) :&= \frac{1}{n}\|\mat{X} - (\mat{W}_2 \mat{W}_1 + \mathbb{I})(\mat{X + A})\|_F^2 + \lambda \| \mat{W}_2 \mat{W}_1 \|_F^2 \nonumber \\ 
&= \frac{1}{n}\|- \mat{A} - \mat{W}_2 \mat{W}_1 (\mat{X + A})\|_F^2 + \lambda \| \mat{W}_2 \mat{W}_1 \|_F^2.
\end{align}
Hence, the loss function for the skip-connected model, $\mathcal{L}_{\text{DAE+SC}}$, can be interpreted as a variant of $\mathcal{L}_{\text{DAE}}$ where the target output $\mat{X}$ is replaced by $- \mat{A}$. In other words, the model learns to reconstruct the noise from the noisy input—a setup commonly used in diffusion models \cite{ho2020denoisingdiffusionprobabilisticmodels}. This observation naturally leads to a more general formulation of the training objective, which we present next.

\paragraph{Training with a General Input-Output Pair}

Both \eqref{eq:loss_dae} and \eqref{eq:loss_dae_skip} can be viewed more generally as training a two-layer linear network to predict an output $\mat{Y} \in \mathbb{R}^{d \times n}$ from inputs $\mat{Z} \in \mathbb{R}^{d \times n}$. In that case, the loss function $\mathcal{L}$ is given by
\begin{equation} \label{section2:general_loss}
\mathcal{L}(\mat{W}_2, \mat{W}_1, \lambda) := \frac{1}{n}\|\mat{Y} - \mat{W}_2 \mat{W}_1 \mat{Z}\|_F^2 + \lambda \| \mat{W}_2 \mat{W}_1 \|_F^2.
\end{equation}
To analyze the generalization error in DAEs, we first need to characterize the critical points $\hat{\mat{W}}_2 \in \mathbb{R}^{d \times k}$ and $\hat{\mat{W}}_1 \in \mathbb{R}^{k \times d}$ that satisfy
\begin{equation*}
 \frac{d}{d \hat{\mat{W}}_2} \mathcal{L}(\hat{\mat{W}}_2, \hat{\mat{W}}_1, \lambda) = 0 \quad \text{and} \quad \frac{d}{d\hat{\mat{W}}_1} \mathcal{L}(\hat{\mat{W}}_2, \hat{\mat{W}}_1, \lambda) = 0.
\end{equation*}
Particularly, we analyze the ridgeless limit of the critical points $\hat{\mat{W}}_c := \hat{\mat{W}}_2 \hat{\mat{W}}_1$, that is $\mat{W}_{\text{c}} = \lim\limits_{\lambda \rightarrow 0} \hat{\mat{W}}_c$.
Ultimately, for $\hat{\mat{W}}^*_2, \hat{\mat{W}}^*_1 = \argmin_{\hat{\mat{W}_1}, \hat{\mat{W}_2}} \mathcal{L}(\hat{\mat{W}}_2, \hat{\mat{W}}_1, \lambda)$, we are interested in the \textit{minimum-norm global minimizer} $\mat{W}_*$ which is given by $\mat{W}_* := \lim\limits_{\lambda \rightarrow 0} \hat{\mat{W}}^*_2 \hat{\mat{W}}^*_1$.

\subsection{General Expressions for Critical Points}
We begin by introducing several notations. Given an input matrix $\mat{Z} \in \mathbb{R}^{d \times n}$, let $\tilde{\mat{Z}} := \mat{Z}\mat{Z}^\top + \lambda \mathbb{I}$ and $\mat{G} := \mat{Y} \mat{Z}^\top \tilde{\mat{Z}}^{-1} \mat{Z} \mat{Y}^\top$. 
Let $\mat{G} = \mat{U}_{\mat{G}} \Lambda_{\mat{G}} \mat{U}_{\mat{G}}^\top$ denote the eigendecomposition of $\mat{G}$. We use the shorthand $[k]$ to denote the set of natural numbers $\{1, 2, \dots, k\}$. For any matrix $\mat{M}$, we write $r_{\mat{M}}$ for its rank and $\mat{M}^{\dagger}$ for its Moore-Penrose pseudo-inverse. For $k \leq r_{\mat{M}}$, let $\mathcal{I}_{\mat{M}, k}$ denote the collection of ordered index sets, where each $I \in \mathcal{I}_{\mat{M}, k}$ satisfies $I \subseteq [r_\mat{M}]$ such that $|I| \leq k$. That is, for $I = \{j_1, \dots, j_{|I|} \}$, we require $1 \leq j_1 < j_2 < \cdots < j_{|I|} \leq r_\mat{M}$. Then, we define the projection onto the corresponding rank-one components of $\mat{M}$ by $P_{I}(\mat{M})  := \sum_{j \in I} \sigma^{\mat{M}}_j \textbf{u}^{\mat{M}}_j {\textbf{v}^{\mat{M}}_j}^\top$, where $\sigma^{\mat{M}}_j$, $\textbf{u}^{\mat{M}}_j$, and $\textbf{v}^{\mat{M}}_j$ denote $j$-th singular value, left singular vector, and right singular vector of $\mat{M}$, respectively. With this notation in place, we now state a simplified version of the critical point characterization; the full version and its proof are provided in Appendix~\ref{appendix:critical_points}.
\begin{theorem}\textnormal{(Critical Points for General Input-Output Pairs).}\label{section:solutions:theorem1}
Assume that
the multiplicity of non-zero eigenvalues of $G$ is $1$, i.e., $\lambda^{\mat{G}}_1 > \lambda^{\mat{G}}_2 > \cdots > \lambda^{\mat{G}}_{r_{\mat{G}}} > 0$.
Further, suppose that the bottleneck dimension $k$ satisfies $k \leq r_{\mat{G}}$.
Then, each index set $I_{\mat{G}} \in \mathcal{I}_{\mat{G}, k}$ characterizes a critical point $\hat{\mat{W}}_c$ of the loss function $\mathcal{L}$, given by
\begin{equation}\label{equation:regularized_critical_points}
            \hat{\mat{W}}_c = \hat{\mat{W}}_2 \hat{\mat{W}}_1 = \mat{U}_{\mat{G}, I_{\mat{G}}} \mat{U}_{\mat{G}, I_{\mat{G}}}^\top \mat{Y} \mat{Z}^\top \tilde{\mat{Z}}^{-1}.
\end{equation}
where $\mat{U}_{\mat{G}, I_{\mat{G}}}$ denotes the submatrix of $\mat{U}_{\mat{G}}$ consisting of the columns indexed by $I_{\mat{G}}$. Moreover, assume that $r_{\mat{Z}} = n$. Then, in the ridgeless limit $\lambda \rightarrow 0$, it holds that
\begin{equation}
    \mat{W}_{\text{c}} := \lim\limits_{\lambda \rightarrow 0} \hat{\mat{W}}_c = P_{I_\mat{G}}(\mat{Y}) \mat{Z}^{\dagger}.
\end{equation} 
Furthermore, the minimum-norm global minimizer $W_*$ is given by 
\begin{equation}
 \mat{W}_* = P_{[k]}(\mat{Y}) \mat{Z}^{\dagger}.
\end{equation}
\end{theorem}
Note that $\mat{W}_{\text{c}}$ is constructed by selecting a subset of rank-one components from the singular value decomposition of $\mat{Y}$. The global minimizer corresponds to the choice $I_{\mat{G}} = [k]$.
\begin{remark}\textnormal{(Comparison with \cite{baldi1989neural}).}
    This result is reminiscent of the classical analysis in \cite{baldi1989neural}, which characterizes critical points in the underparameterized regime ($d < n$). In contrast, our theorem extends this line of work to the overparameterized setting, while additionally incorporating a regularization term to find the minimum-norm solution.
\end{remark}

\begin{remark}\textnormal{(Effect of Overparameterization and Minimum-Norm Principle).} Observe that Eq.~(\ref{equation:regularized_critical_points}) involves the matrix $U_{G, I}$, whose columns are eigenvectors of $\mat{G}$. Its dependence on both $Y$ and $Z$ makes direct analysis challenging. However, in the overparameterized regime and in the ridgeless limit, $\mat{G}$ simplifies to $\mat{G} = \mat{Y}\mat{Y}^\top$. This leads to a more tractable expression for $\mat{W}_{\text{c}}$, which now depends only on a projection of $\mat{Y}$ and the pseudo-inverse of $\mat{Z}$. This enable us to build on technical tools developed in \cite{sonthalia2022training, kausik2024doubledescentoverfittingnoisy} to characterize the generalization risk, as detailed in Section \ref{section:generalization}.
\end{remark}
\subsection{Specification to the Denoising Setup}\label{subsection:denoising_setup}

We now specialize the result of Theorem~\ref{section:solutions:theorem1} to the denoising autoencoder setting, yielding closed-form solutions for the models introduced earlier.
We assume that the eigenvalues of both $\mat{X}\mat{X}^\top$ and $\mat{A}\mat{A}^\top$ have multiplicity one. We also assume that the matrix $\mat{X} + \mat{A}$ is full-rank, which holds almost surely (see \cite{feng2007rank}). Under this setup, we obtain the following corollaries for the two DAE models \eqref{eq:loss_dae} and \eqref{eq:loss_dae_skip}.

\begin{corollary}\textnormal{(Critical Points of the Model without Skip Connections).} \label{theorem:solution_noskip}
    Let $\mathcal{I}_{\mat{X}}$ be a family of index sets, where each element is an ordered set of distinct natural numbers from $[k]$, where $k \leq r_{\mat{X}}$. Then, for each $I^x \in \mathcal{I}_{\mat{X}}$, a critical point is given by $\mat{W}_{\text{c}} = P_{I^x}(\mat{X})(\mat{X + A})^{\dagger}$. Moreover, the global minimizer $\mat{W}_*$ is given by $\mat{W}_* = P_{[k]}(\mat{X})(\mat{X + A})^{\dagger}$.
\end{corollary}
\begin{corollary}\textnormal{(Critical Points of the Model with Skip Connections).} \label{theorem:solution_skip}
    Let $\mathcal{I}_{\mat{A}}$ be a family of index sets, where each element is an ordered set of distinct natural numbers from $[k]$, where $k \leq r_{\mat{A}}$. Then, for each $I^a \in \mathcal{I}_{\mat{A}}$, a critical point is given by $\mat{W}^{\text{sc}}_{c} = - P_{I^a}(\mat{A})(\mat{X + A})^{\dagger}$. Moreover, the global minimizer $\mat{W}^{\text{\text{sc}}}_*$ is given by $\mat{W}^{\text{\text{sc}}}_{*} = - P_{[k]}(\mat{A})(\mat{X + A})^{\dagger}$.
\end{corollary}
These closed-form expressions enable us to derive formulas for the test risk of both models. This is the topic of the following section.
\section{The Generalization Error of Linear DAEs}\label{section:generalization}
In this section, we analyze the test error corresponding to the critical points derived in Section~\ref{subsection:denoising_setup}. We begin by outlining the data assumptions of our analysis and then derive expressions for the test error of the models with and without skip connections. Next, we introduce a natural bias--variance decomposition that arises from the bottleneck structure, and examine how varying the bottleneck dimension influences bias and variance in both models. Finally, we highlight the surprising effect that adding skip connections produces a smoother test error curve.
We first present the data model that forms the basis of our analysis. For training, we consider 
$\mat{X} \in \mathbb{R}^{d \times n}$ as the clean (noise-free) data matrix and $\mat{A} \in \mathbb{R}^{d \times n}$ as the associated additive Gaussian noise matrix. Likewise, $\mat{X}_{\text{\text{tst}}} \in \mathbb{R}^{d \times N_{\text{\text{tst}}}}$ and $\mat{A}_{\text{\text{tst}}} \in \mathbb{R}^{d \times N_{\text{\text{tst}}}}$ denote the clean and noise matrices used for testing, respectively. Note that in this section, we work on non-asymptotic setting where $d, n$ are high-dimensional but finite. 
\begin{assumption}\textnormal{(Data Assumptions).}\label{data_assumption}
\begin{enumerate}[left=1em]
    \item Normalized Low Rank: $\mat{X}$ is normalized such that $\|\mat{X}\|_2 = \Theta(1)$. Its rank, denoted as $r$, satisfies $r \ll d, n$. 
    \item Well Conditioned: The ratio between the largest singular value and the smallest nonzero singular value of $\mat{X}$ is $\Theta(1)$.
    \item Noise: The entries of $\mat{A}, \mat{A}_{\text{\text{tst}}}$, are sampled independently from $\mathcal{N}(0, \frac{\eta_{\text{\text{trn}}}^2}{d})$, $ \mathcal{N}(0, \frac{\eta_{\text{\text{tst}}}^2}{d})$ respectively, where $\eta_{\text{\text{trn}}}, \eta_{\text{\text{tst}}} = \Theta(1)$.
    \item Test Data: $\mat{X}_{\text{\text{tst}}}$ is assumed to lie in the same low dimensional subspace as the training data. In other words, the test data $\mat{X}_{\text{\text{tst}}}$ satisfies $\mat{X}_{\text{\text{tst}}} = \mat{U} \mat{L}$, for $\mat{U} \in \mathbb{R}^{d \times r}$ the left singular vectors of $\mat{X}$ and for some non-zero coefficient matrix $\mat{L} \in \mathbb{R}^{r \times N_{\text{\text{tst}}}}$. 
\end{enumerate}
\end{assumption}
The data scaling assumption for $X$ ensures that the signal matrix and the noise matrix are comparable in magnitude. This is motivated by the fact that the spectral norm of the noise matrix satisfies $\|\mat{A}\|_2 = O(1)$ with high probability, as established in \cite[Theorem 4.4.5]{vershynin2018high}. The low-rank assumption on $\mat{X}$ is supported by empirical evidence that real-world datasets are approximately low-rank, as argued in \cite{udell2018bigdatamatricesapproximately} and adopted by \cite{kausik2024doubledescentoverfittingnoisy}. A crucial point emphasized in \cite{kausik2024doubledescentoverfittingnoisy} is that the training data $X$ is treated as an arbitrary but deterministic low-rank matrix \textit{without any distributional assumptions}. In particular, the observations need not be independent. 

Given the critical points $\mat{W}_{\text{c}}$ and $\mat{W}_{\text{c}}^{\text{\text{sc}}}$, we evaluate the test error. Following \cite{sonthalia2022training}, the test error of the model without skip connections is given by
\begin{equation}\label{test_metric}
    R(\mat{W}_{\text{c}}, \mat{X}_{\text{\text{tst}}}) := \frac{1}{N_{\text{\text{tst}}}} \mathbb{E}_{\mat{A}_{\text{\text{trn}}}, \mat{A}_{\text{\text{tst}}}}\left[\|\mat{X}_{\text{\text{tst}}} - \mat{W}_{\text{c}} (\mat{X}_{\text{\text{tst}}} + \mat{A}_{\text{\text{tst}}})\|_F^2\right].
\end{equation}
For the model with skip connections, it is given by
\begin{equation}\label{test_metric_skip}
    R_{\text{\text{sc}}}(\mat{W}_{\text{c}}^{\text{sc}}, \mat{X}_{\text{\text{tst}}}) := \frac{1}{N_{\text{\text{tst}}}} \mathbb{E}_{\mat{A}_{\text{\text{trn}}}, \mat{A}_{\text{\text{tst}}}}\left[\|\mat{X}_{\text{\text{tst}}} - (\mat{W}_{\text{c}}^{\text{\text{sc}}} + \mathbb{I}) (\mat{X}_{\text{\text{tst}}} + \mat{A}_{\text{\text{tst}}})\|_F^2\right].
\end{equation}

\subsection{Effect of Bottleneck Layers on Generalization Error}
To understand how the bottleneck dimension influences generalization, we start by analyzing the test error for the model without skip connections \eqref{test_metric}. To this end, we plug in the the critical points $\mat{W}_c$ obtained in Corollary~\ref{theorem:solution_noskip}. We then explore a natural bias–variance decomposition of the generalization error, focusing on the global minimizer $\mat{W_*}$.
\begin{theorem}\textnormal{(Test Error for the Model without Skip Connections).}\label{theorem:test_risk_noskip}
    Let $\alpha_i := \sigma_i \eta_{\text{\text{trn}}}^{-1}$, where $\sigma_i$ denotes the $i$-th singular value of $\mat{X}$.
    Let $d \geq n + r$, and $c := \frac{d}{n}$. Let $\mat{J} \in \mathbb{R}^{r \times r}$ be the diagonal matrix
    \begin{align*}
        \mat{J}_{ii} = \left(\alpha_i^2 + 1\right)^{-2} \cdot \mathds{1}_{i \in I^x} 
             + \mathds{1}_{i \notin I^x}
    \end{align*} where $\mathds{1}_{(\cdot)}$ denotes the indicator function.
    Then, for a critical point $\mat{W}_{\text{c}}$ we have that
    \begin{align}\label{equation_noskip}
    R(\mat{W}_{\text{c}}, \mat{X}_{\text{\text{tst}}})
    &= \frac{1}{N_{\text{\text{tst}}}}\Tr(\mat{J} \mat{L}\mat{L}^\top) + \frac{\eta_{\text{\text{tst}}}^2 c}{d(c - 1)} \sum\limits_{j \in I^x} \frac{\alpha_j^2}{1 + \alpha_j^2} + o\left(\frac{1}{d}\right).
    \end{align}
\end{theorem}
See Appendix~\ref{appendix:section3:proofs} for the proof of this theorem.
\begin{remark}\textnormal{(Global Minimizer).} The test risk for $\mat{W}_*$ is obtained by plugging in $[k]$ for $I^x$.
\end{remark}
\begin{remark}\label{remark:bias}\textnormal{(Bias Term).}
Since $\mat{L}$ depends only on the test data, the overall magnitude of $\Tr(\mat{J}\mat{L}\mat{L}^\top)$ is mainly influenced by the size of the diagonal entries of $\mat{J}$.
\end{remark}

\paragraph{Bottleneck Dimension as a Complexity Measure}
Consider the case of the global minimizer, where $I^x = [k]$, and all parameters except $k$ are fixed. Then, the first term of Eq. (\ref{equation_noskip}) \textit{decreases} as $k$ increases towards $r$. This is because the diagonal entries of 
$\mat{J}$ are either $1$ or of the form $(\alpha_i^2 + 1)^{-2} < 1$. 
As $k$ increases, the number of $1$s decreases, leading to a lower value. 
Conversely, the second term \textit{increases} as $k$ grows, 
due to the increase of the number of summands. 
This trade-off behavior aligns with the \textit{classical understanding of bias--variance trade-off}, as adjusting $k$ is directly related to varying the model complexity. 

In light of this, we interpret the first term of Eq.~(\ref{equation_noskip}) as the bias component and the second term as the variance. Our notion of bias and variance aligns with that of \cite{sonthalia2022training}: the bias term is derived from the expected reconstruction error on clean data, given by $N_{\text{\text{tst}}}^{-1} \mathbb{E}[\|\mat{X}_{\text{\text{tst}}} - \mat{W}_{\text{c}} \mat{X}_{\text{\text{tst}}}\|_F^2]$ , while the variance term arises from $d^{-1} \eta_{\text{\text{tst}}}^2 \mathbb{E}[\|\mat{W}_{\text{c}}\|_F^2]$, which relates to the norm of the estimator (see Appendix~\ref{appendix:section3:bias_variance_decomposition}). 

Under this decomposition, fixing $k$ and $n$ while increasing $d$ (thus increasing overparameterization) leads to a decrease in the overall test error due to the reduced variance term. Notably, this occurs without a corresponding increase in bias, and therefore, without a trade-off. This absence of trade-off reflects the \textit{modern understanding} of the bias--variance relationship, in which larger models can generalize better. In this sense, Theorem~\ref{theorem:test_risk_noskip} illustrates the \textit{coexistence} of both classical and modern perspectives on the interplay between bias and variance in the DAE setting. 

\begin{figure}
    \centering
    {{\includegraphics[width=4.4cm, height=3.2cm]{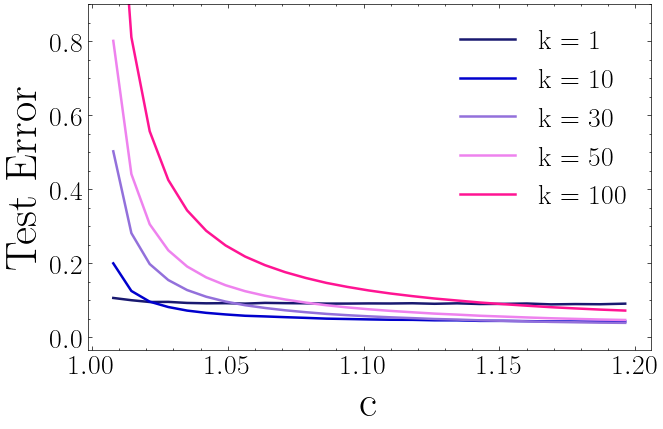} }}%
    {{\includegraphics[width=4.4cm, height=3.2cm]{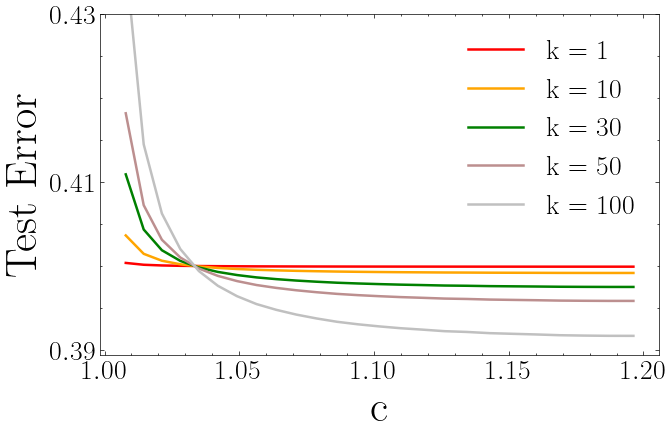} }}%
    {{\includegraphics[width=4.99cm, height=3.2cm]{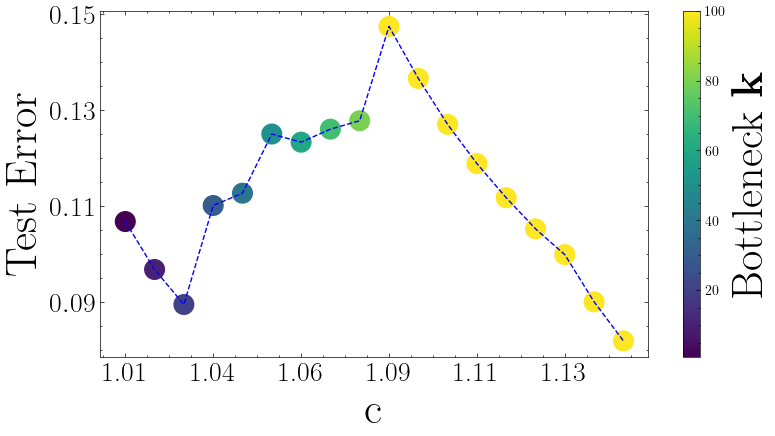} }}%
\caption{\textbf{Effect of Bottleneck.} Test errors on CIFAR-10 illustrating how the bottleneck dimension $k$ influences generalization. The left plot corresponds to the model without skip connections. The center shows results for the model with skip connections. The right subfigure is constructed by jointly increasing both $k$ and $d$, using the corresponding test errors from the left plot. As seen in the left and center plots, the optimal choice of $k$ depends on the level of overparameterization, reflecting a distinct bias--variance trade-off in different regimes.
}
\label{fig:bottleneck}
\end{figure}

This trade-off is further illustrated in Figure~\ref{fig:bottleneck}, which shows that for fixed input dimension $d$, a smaller bottleneck dimension $k$ can improve test error in certain regimes by reducing the variance term---dominant in the mildly overparameterized setting due to peaking behavior. Moreover, the right plot of Figure~\ref{fig:bottleneck} demonstrates that jointly increasing input dimension $d$ and bottleneck dimension $k$ leads to a \textit{second peak} in the test curve within the overparameterized regime. These findings underscore that both the degree of overparameterization in $d$ and the choice of $k$ shape the generalization of DAEs.

\subsection{Including Additional Skip Connections} 
Having examined the effect of bottleneck layers, we now investigate how the inclusion of skip connections influences test performance, particularly through its effect on variance. As in the previous subsection, we first present a theoretical result and then interpret it through numerical experiments. Consider any critical point $\mat{W}^{\text{\text{sc}}}_{\text{c}}$ from Corollary \ref{theorem:solution_skip}, with the corresponding index set $I^a$. Then, for $d \geq n + r$, we have the following theorem.
\begin{figure}
    \centering
    {{\includegraphics[width=4.5cm, height=3.2cm]{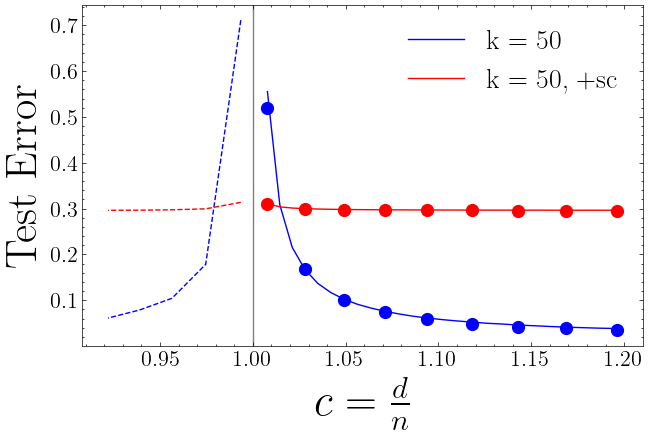} }}%
    {{\includegraphics[width=4.5cm, height=3.2cm]{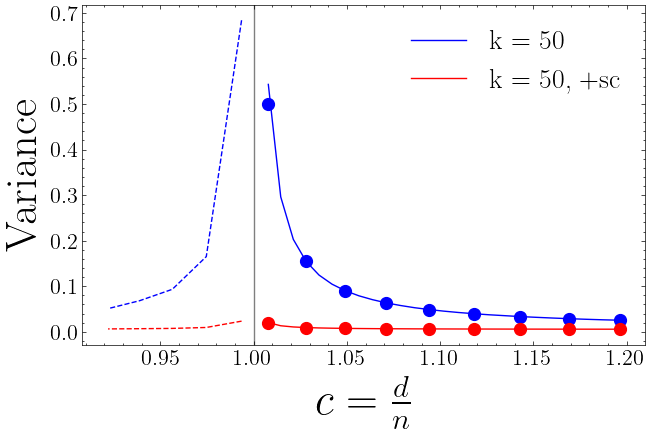} }}%
    {{\includegraphics[width=4.8cm, height=3.2cm]{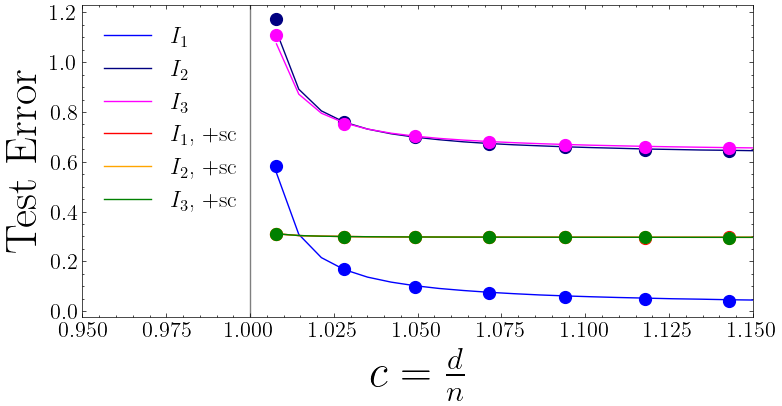} }}%
    \caption{\textbf{Effect of Skip Connections.} Experiments on CIFAR-10 (Data rank is fixed at $r = 100$). Solid lines represent theoretical predictions, while cross markers indicate empirical results. Dotted lines correspond to empirical values in the underparameterized regime (solutions derived from \cite{baldi1989neural}). Red lines and markers denote results for the model with skip connections. The left subfigure shows the test error; the center subfigure displays the corresponding variance curve. The right subfigure compares test errors across different critical points: $I_1 = [50]$, $I_2 = [11, 60]$, and $I_3 = [31, 80]$.  }
    \label{fig:skip_connection}
\end{figure} 
\begin{theorem}\textnormal{(Test Error for the Model with Skip Connections).}\label{theorem:test_risk_skip} 
     Let $\mat{J}^{\text{\text{sc}}}$ be a diagonal matrix defined as, 
        $\mat{J}^{\text{\text{sc}}}_{ii} = 
            \frac{c + (c - 1)\sigma_i^2}{c(1 + \eta_{\text{\text{trn}}}^{-2} \sigma_i^2)^2}$, for each $i \in [r]$.
    Then, for a critical point $\mat{W}^{\text{\text{sc}}}_{\text{c}}$, we have that
    \begin{align} \label{equation_skip}
        R_{\text{\text{sc}}}(\mat{W}^{\text{\text{sc}}}_{\text{c}}, \mat{X}_{\text{\text{tst}}})
        &= \eta_{\text{\text{tst}}}^2 \left(1 - \frac{|I^a|}{d}\right) + \frac{|I^a|}{d N_{\text{\text{tst}}}} \Tr\left(\mat{J}^{\text{\text{sc}}} \mat{L}\mat{L}^\top\right) + \frac{\eta_{\text{\text{tst}}}^2 |I^a|}{d^2} \frac{c}{c - 1}\sum\limits_{i=1}^r \frac{\sigma_i^2}{(\eta_{\text{\text{trn}}}^2 + \sigma_i^2)} \nonumber \\
        &+ \frac{3\eta_{\text{\text{tst}}}^2 |I^a|}{dn} \frac{1}{c} \sum\limits_{i=1}^r \frac{\eta_{\text{\text{trn}}}^2 \sigma_i^2}{(\eta_{\text{\text{trn}}}^2 + \sigma_i^2)} + O\left(\frac{1}{dN_{\text{\text{tst}}}}\right). 
    \end{align}
\end{theorem}
Similar to the previous theorem, with the global minimizer $\mat{W}_*^{\text{\text{sc}}}$, we replace $I^a$ to $[k]$. The proofs of this theorem can be found in Appendix~\ref{appendix:section3:proofs}.
\begin{remark}\textnormal{(Bias and Variance Terms).}\label{remark:variance_skip}
Following the approach in the previous subsection, we interpret the norm term, i.e., $\eta_{\text{\text{tst}}}^2 d^{-1} \|\mat{W}_*^{\text{\text{sc}}}\|_F^2$, as the variance term, with the remaining terms attributed to bias. Importantly, the third term involving $(c - 1)^{-1}$ arises directly from the norm term. This decomposition is consistent with a unified definition of bias and variance across the models with\&without skip connections, where bias decreases with increasing model complexity (measured by bottleneck size), and variance increases. 
We formalize this interpretation in Appendix~\ref{appendix:section3:bias_variance_decomposition}.
\end{remark}
\begin{remark}\textnormal{(Small Difference in Test Error Among Critical Points).} From Eq.(\ref{equation_skip}), the first term dominates the test error, while the remaining terms are comparatively small. Based on the decomposition in Remark~\ref{remark:variance_skip}, the leading term reflects the bias, with the remaining terms contributing to the variance (see Appendix~\ref{appendix:section3:bias_variance_decomposition} for details). As a result, the overall test error exhibits only minor variation across different critical points. This contrasts with the model with no skip connections, where different critical points can significantly influence both the bias and variance components. The right subfigure of Figure~\ref{fig:skip_connection} illustrates this.
\end{remark}
\paragraph{The Impact of Skip Connections on the Test Error Curve}
Observe first that the variance term in Eq.(\ref{equation_noskip}) is responsible for the sharp increase in the test curves as the ratio 
$c$ approaches $1$, due to the inclusion of $(c - 1)^{-1}$. A similar albeit less pronounced trend is observed in the model with skip connections. The term
$\frac{\eta_{\text{\text{tst}}}^2 |I^a|}{d^2} \frac{c}{c - 1}\sum_{i=1}^r \frac{\sigma_i^2}{(\eta_{\text{\text{trn}}}^2 + \sigma_i^2)}$ also includes the factor $(c - 1)^{-1}$. However, in contrast to the model without skip connections, the expression is multiplied with an additional factor of $d^{-1}$. This suggests that \textit{skip connections help mitigate the sharp rise in variance} that typically occurs when the model is in the moderately overparameterized regime, leading to more \textit{stable generalization performance} even in this regime. 
We now examine this intriguing phenomenon more closely to better understand its origin. 

\section{Explaining the Variance Discrepancy Between DAEs With and Without Skip Connections}\label{section:rmt}
As seen in the previous section, the variance term is scaled by an additional $d^{-1}$ factor for models with skip connections. As Figure \ref{fig:skip_connection} illustrates, this difference becomes particularly pronounced as $c = d/n$ approaches 1. However, the underlying cause of this behavior is not immediately clear. 

In this section, we identify the source of the discrepancy. We remind the reader of Eq.~(\ref{eq:loss_dae_skip}), which shows that the skip connection effectively cancels the signal component, reorienting the learning task toward predicting the noise. We demonstrate that this shift leads to weaker alignment between certain singular vector pairs, which in turn yields a substantially smaller variance.

For the remainder of this section, let $\bar{\lambda}_j$, $\lambda_j$, and $\lambda^{\mat{A}}_j$ denote the squared $j$-th singular values of the matrices $\mat{X + A}$, $\mat{X}$, and $\mat{A}$, respectively. Similarly, let $\bar{\mat{V}}$, $\mat{V}$, and $\mat{V}_{\mat{A}}$ denote the corresponding matrices of right singular vectors for $\mat{X + A}$, $\mat{X}$, and $\mat{A}$. Then,
the Frobenius norms of the global minimizers with and without skip connections are given by:
\begin{align}\label{equation:norm_noskip}
\quad\|\mat{W}_*^{\text{\text{sc}}}\|_F^2 = \sum_{j = 1}^n \bar{\lambda}_j^{-1} \sum_{i = 1}^k \lambda_i^{\mat{A}} (\mat{V}_{\mat{A}}^\top \bar{\mat{V}})_{ij}^2
 \quad \text{and} \quad \|\mat{W}_*\|_F^2 = \sum_{j = 1}^n \bar{\lambda}_j^{-1} \sum_{i = 1}^k \lambda_i (\mat{V}^\top \bar{\mat{V}})_{ij}^2.
\end{align}
Note that $\bar{\lambda}_j$ is shared between both models. Moreover, under data assumptions~\ref{data_assumption}, the $k$ first squared singular values $\lambda_i$ and $\lambda^{\mat{A}}_i$ scale at the same rate. Therefore, any substantial difference between the two must be attributed to what we refer to as the  \textbf{alignment terms} $(\mat{V}^\top \bar{\mat{V}})_{ij}^2$ and $(\mat{V}_{\mat{A}}^\top \bar{\mat{V}})_{ij}^2$. 

Specifically, for $i \in \{ 1, \dots, k\}$ and $j \in \{1, \dots, n\}$, the term $(\mat{V}^\top \bar{\mat{V}})_{ij}^2$ quantifies the squared inner product between the $i$-th singular vector of the signal matrix $\mat{X}$ and the $j$-th singular vector of the signal-plus-noise matrix $\mat{X + A}$. In contrast, $(\mat{V}_{\mat{A}}^\top \bar{\mat{V}})_{ij}^2$ captures the corresponding alignment between the noise matrix $\mat{A}$ and $\mat{X + A}$. Ideally, we would study the alignment behavior through the "Information-plus-Noise" model \cite{dozier2007empirical}, defined by $(\mat{X} + \mat{A})(\mat{X} + \mat{A})^\top$. However, this is technically challenging. Therefore, we instead focus on a simpler, but conceptually closely related model\footnote{This simplified setting belongs to a broader class of models where a low-rank signal is perturbed by additive noise, commonly referred to as "spiked models" \cite{baik2004eigenvalueslargesamplecovariance}. We formally define this model class in Appendix~\ref{appendix:rmt_background:spiked_models}. For a comprehensive overview, see \cite{couillet_liao_2022}.}.

\begin{definition}\textnormal{(Rank-1 Additive Model).}\label{def:rank1_add_model}
Let $\mat{X}$ and $\mat{A}$ satisfy Assumptions~\ref{data_assumption}. Further assume that $\mat{X}$ is rank-1, i.e., $\mat{X} = \sqrt{\lambda_1} \mathbf{u}_1 \mathbf{v}_1^\top$, where $\sqrt{\lambda_1}$ is the largest singular value of $\mat{X}$, and $\mathbf{u}_1$, $\mathbf{v}_1^\top$ are the corresponding singular vectors. We then define the additive model
$\mat{S} := \mat{X}\mat{X}^\top + \mat{A}\mat{A}^\top.$ 
\end{definition}

\begin{remark}\textnormal{(Connection between Additive Models and Information-plus-noise Models).}
The expected value of the Information-plus-Noise model coincides with that of the Additive model: $\mathbb{E}[(\mat{X} + \mat{A})(\mat{X} + \mat{A})^\top] = \mathbb{E}[\mat{X}\mat{X}^\top + \mat{A}\mat{A}^\top]$, since $\mathbb{E}[\mat{X}\mat{A}^\top] = \mathbb{E}[\mat{A}\mat{X}^\top] = 0$.
\end{remark}
With this simplified model, our focus is on comparing the alignment between the top eigenvector of $\mat{X}\mat{X}^\top$ and the $j$-th eigenvector of $\mat{S}$ (Denoted as $\vec{u}_j^{\mat{S}}$), and the alignment between the top eigenvectors of $\mat{A}\mat{A}^\top$ (Denoted as $\vec{u}_i^{\mat{A}}$ for $i \in [k]$) and the same $\vec{u}_j^{\mat{S}}$. Denote by $\lambda_j^{\mat{S}}$ the $j$-th eigenvalue of $\mat{S}$. The corresponding proofs of the theorem below are given in Appendix~\ref{appendix:rmt_results:main_proof}.
\begin{theorem}\label{theorem:alignment}\textnormal{(Skip Connections Cause Weaker Alignment).} For $c \in (0, \infty)$, $i \in [k]$, and $j \in [2,n]\backslash \{i-1, i\}$,
    \begin{equation}
    \mathbb{E}\left[\frac{\langle \mathbf{u}^{\mat{A}}_{i}, \mathbf{u}^{\mat{S}}_j\rangle^2}{\langle \mathbf{u}_1, \mathbf{u}^{\mat{S}}_j \rangle^2}\right] = \Theta\left(\frac{1}{d(\lambda_i^{\mat{A}} - \lambda_j^{\mat{S}})^2}\right).
    \end{equation}
\end{theorem}   

This theorem suggests that when the eigenvalue gap is large, the alignment term in models with skip connections becomes significantly weaker than in models without them. To see this, the first point to notice is that the eigenvalues of $\mat{S}$ follow the \textit{Marchenko–Pastur distribution}. This implies that as $c$ approaches 1, \textit{an increasing number of eigenvalues of $\mat{S}$ concentrate near zero} (see Appendix~F.1).

Now consider the term $(\lambda_i^{\mat{A}} - \lambda_j^{\mat{S}})^2$. For example, take $i = 1$. It is known that $\lambda_1^{\mat{A}}$ converges almost surely to $\eta_{\text{\text{tst}}}^2 c^{-1}(1 + \sqrt{c})^2$ as $d, n \to \infty$ \cite[Theorem 5.8]{bai2010spectral}. For small $\lambda_j^{\mat{S}}$, which care plentiful when $c \approx 1$ and concentrate around $0$, we obtain $(\lambda_i^{\mat{A}} - \lambda_j^{\mat{S}})^2 = \Theta(1)$. As a result, the ratio between the alignment terms is $\Theta(d^{-1})$. 
This is important because the Frobenius norms in Eq.~(\ref{equation:norm_noskip}) involve the \textit{inverses} of the eigenvalues of the corrupted input covariance matrix. Hence, small eigenvalues \textit{dominate} the variance. More precisely, it is known that with high probability, the smallest eigenvalue of $\mat{A}\mat{A}^\top$ scales as $\Theta((\sqrt{d} - \sqrt{n - 1})^2)$ \cite{rudelson2009smallest}. At $c = d/n = 1$ this scales as $\Theta(d^{-2})$, implying that its inverse is of order $\Theta(d^2)$. 
Accordingly, the Frobenius norms are largely influenced by the smallest eigenvalues.
Importantly, for those small eigenvalues, Theorem~\ref{theorem:alignment} shows that the corresponding alignment terms in skip-connected models are \textit{suppressed} by a factor of $\Theta(d^{-1})$ relative to models without skip connections. This directly accounts for the reduced variance and explains the smoother generalization curves observed in the previous section.

\section{Discussion}\label{section:discussion}
\paragraph{Bottleneck Dimension as an Additional Complexity Measure}
In supervised settings, \cite{curth2023a} empirically show that by controlling an additional complexity measure along with the input parameter count, the test error curve can take on diverse shapes, ranging from the traditional U-shaped curve to ones exhibiting multiple descents. Our work provides concrete theoretical evidence that this interpretation extends to unsupervised scenarios, identifying the bottleneck dimension as a key complexity measure in DAEs. Beyond this, our findings uncover a subtle yet crucial distinction for denoising autoencoders: the emergence of a bias–variance trade-off, governed by the number of neurons in the bottleneck layer. This phenomenon is not considered in \cite{curth2023a}, whose focus is on Principal Component Regression (PCR) rather than denoising. We elaborate on this further in the next paragraph.

\paragraph{PCA-based Methods vs. Two-Layer Linear DAEs}
The key difference lies in where the dimensionality reduction happens. While PCA-based methods (including PCR \cite{Teresa_2022, pmlr-v235-gedon24a, curth2023a} and PCA-denoising \cite{cui2023high}) identify the top-$k$ \textit{input} components, Theorem~\ref{theorem:solution_noskip} shows that two-layer linear DAEs align with the top-$k$ directions of the \textit{output}. This is a consequence of the critical points identified in Corollaries \ref{theorem:solution_noskip}--\ref{theorem:solution_skip}, which lead to generalization behavior distinct from PCA (cf. Figure \ref{fig:compare_test_risk}). Theorem \ref{theorem:test_risk_noskip} and Figure \ref{fig:skip_connection} highlight that the double descent phenomenon as a function of $d/n$ persists, even for small bottleneck dimension $k$. In particular, the variance term becomes dominant near $c \approx 1$, significantly influencing the test error---an effect that diminishes with increasing overparameterization.
Conversely, PCA-based methods suppress this peaking behavior by discarding small eigenvalues of the input data.
\paragraph{The Role of Skip Connections \& Diffusion Models}  Instead of eliminating small eigenvalues, skip connections attenuate their contribution.
This improves generalization in certain regimes of $d / n$, which is in line with previous works \cite{cui2023high}. Theorems \ref{theorem:test_risk_skip} and \ref{theorem:alignment} extend this understanding by identifying the variance term as the primary driver of the improvement. Interestingly, the input--output structure in Eq.~(\ref{eq:loss_dae_skip}) mirrors that of diffusion models \cite{ho2020denoisingdiffusionprobabilisticmodels}, where the network is trained to predict noise rather than clean signals---a design choice shown to improve performance empirically \cite[Sec. 3.2]{ho2020denoisingdiffusionprobabilisticmodels}. In relation to this, our results suggest two explanations: first, reduced variance; and second, as shown in the right plot of Figure~\ref{fig:skip_connection}, the presence of non-global critical points that perform comparably to the global minimizer, potentially easing optimization.


\paragraph{Interpolation \& Double Descent Phenomenon} 

Broadly, the double descent phenomenon suggests that generalization error peaks near the "interpolation threshold" \cite{Belkin_2019} where a network attains (nearly) zero training error, before decreasing again as the model enters the overparameterized regime. Empirical works such as \cite{Belkin_2019}, \cite{nakkiran2019deepdoubledescentbigger}, as well as
prior theoretical works such as \cite{hastie2020surpriseshighdimensionalridgelesssquares}, \cite{bach2023highdimensionalanalysisdoubledescent} show that this second
descent occurs after the model interpolates the training data. However, when the bottleneck dimension is lower than the rank of the input and output data, \textit{exact interpolation
is not possible}. However, our results reveal that a peak can still exist in the regime $d/n \approx 1$ (cf. $k<r$ in Figure \ref{fig:bottleneck}).


\paragraph{Future Work \& Limitations} Our paper opens several pathways for future work to explore. While our setting strictly extends prior works on single-layer networks \cite{sonthalia2022training, kausik2024doubledescentoverfittingnoisy} to two-layer networks with bottleneck (and skip connection), it remains restricted to the linear regime. Although \cite{cui2023high} explores a non-linear setting, their analysis relies on a tied-weights assumption, and their resulting formulas do not capture the mechanisms we highlight. Extending our work to non-linear models is therefore an important direction for future work. 
Furthermore, our study focuses on two-layer networks with a single skip connection from input to output. Investigating deeper architectures with skip connections between intermediate layers may yield deeper insights into their role in shaping generalization.

\printbibliography


\appendix
\section*{Appendix}
We include additional material in the appendix. Section~\ref{appendix:related_works} discusses further related work. Section~\ref{Appendix:notations} introduces general notational conventions. Section~\ref{appendix:rmt_background} provides background on Random Matrix Theory to support the results in Section~\ref{appendix:rmt_results}. Section~\ref{appendix:critical_points} presents the proofs for Section~\ref{section:generalization} and proposes a definition of bias--variance tailored to our setting. Section~\ref{appendix:rmt_results} proves the results from Section~\ref{section:rmt} and includes additional supporting results. Section~\ref{appendix:experiments} describes the experimental settings used to generate the figures in the main text.

\section{Related Works}\label{appendix:related_works}
In this work, we focus on two-layer linear networks and emphasize that the current progress remains limited to linear models. Prior studies of two-layer non-linear architectures typically impose additional architectural constraints, such as random projection methods (where one layer is fixed) or weight-tying assumptions between the layers. These constraints hinder a complete analysis of two-layer networks. To the best of our knowledge, our work is the first to address the full two-layer linear network architecture in the overparameterized setting without such restrictions, although our analysis is confined to the denoising setting. Additional related works are discussed in detail, where we also clarify how our approach compares to prior literature.

\paragraph*{Loss Landscape and Critical Points of Linear Neural Network} 
While the loss landscape properties of training metric are well studied using standard linear algebra \cite{baldi1989neural, kawaguchi2016deeplearningpoorlocal}, second-order analysis \cite{achour2024losslandscapedeeplinear}, and algebraic geometry \cite{trager2020purespuriouscriticalpoints}, comparatively few works focused on the analytical characterization of solutions and the
generalization error. \cite{baldi1989neural} provided an analytical description of the
critical points in two-layer linear networks under the underparameterized setting,
and \cite{zhou2017criticalpointsneuralnetworks} extended this to multi-layer linear networks. Yet these
efforts leave open questions regarding regularized solutions, and the corresponding
generalization error. This paper builds on this body of work by deriving regularized
solutions for two layer linear DAEs and analyzing their generalization error in the context of bottleneck layers.

\paragraph{Characterizing Generalization Behavior in Overparameterized Setting} The double descent phenomenon has emerged as a key insight in understanding generalization in overparameterized models, prompting significant interest in analyzing simple models within this regime. This behavior has been particularly well studied in linear regression with Gaussian inputs, often through the lens of minimum-norm solutions and regularized settings such as Lasso~\cite{gerbelot2020asymptoticerrorsconvexpenalized} and ridge regression~\cite{ho2020denoisingdiffusionprobabilisticmodels, bach2023highdimensionalanalysisdoubledescent}. Extensions to nonlinear two-layer neural networks have been explored in~\cite{bach2023highdimensionalanalysisdoubledescent, ba2020generalization}, where one of the weight matrices is trained while the other is fixed randomly. These studies show that double descent arises when the hidden layer dimension scales proportionally with the number of data points, in both nonlinear and linear settings--mirroring similar findings in the PCA setting \cite{pmlr-v235-gedon24a, Teresa_2022}.
In contrast, our work takes the opposite approach in the context of denoising autoencoders (DAEs): we analyze the two-layer linear case without fixing either weight matrix and without scaling the hidden layer width with the dataset size. This allows for a complete characterization of the generalization behavior in a fully trainable two-layer linear architecture. 

\paragraph{Discussion of Double Descent Phenomenon in Auto-Encoder Setting} There is ongoing debate about the presence of the double descent phenomenon in autoencoder models. For example, \cite{lupidi2023does} conjectured its absence in self-supervised learning tasks and provided empirical evidence using reconstruction autoencoders (RAEs). In contrast, \cite{rahimi2024multiple} demonstrated that double descent does occur in deep undercomplete RAEs, particularly when the input data includes noisy measurements. Although these studies focus on RAEs, their findings---alongside ours---suggest that input noise plays a critical role in shaping the generalization behavior of undercomplete autoencoders.

\section{Notations}
\label{Appendix:notations}
\subsection{General Notations}
For a matrix $\mat{M} \in \mathbb{R}^{d \times n}$, we use
\begin{itemize}
    \item $\Tr(\mat{M})$, $\|\mat{M}\|_F$, $\|\mat{M}\|_2$, $\mat{M}^{\dagger}$, and $\vec{m}_i$ to denote the Trace, Frobenius norm, spectral norm, Moore-Penrose pseudoinverse, and $i$-th column vector of $\mat{M}$, respectively.
    \item $r_\mat{M}$, $\mathbb{I}_d$ to denote the rank of $M$, and the identity matrix of size $d \times d$.
    \item For an index set $I \subset \{1, \dots, n\}$, $\mat{M}_I$ denotes the submatrix of $\mat{M}$ with its columns indexed by $I$. $\mat{M}_p$ denotes the submatrix of $\mat{M}$ with its first $p$ columns.
\end{itemize}
Some set notations:
\begin{itemize}
    \item $[p] = \{1, 2, \dots, p\}$ denotes the set of natural numbers from $1$ to $p$.
    \item Similarly, $[p, q]$ denotes $\{p, p+1, \dots, q\}$, for $p, q \in \mathbb{N}$.
    \item $|I|$ denotes the cardinality of a set $I$.
\end{itemize}
Additionally,
\begin{itemize}
    \item $\vec{v}$ denotes a vector, and $\|\vec{v}\|_2$ denotes its Euclidean norm.
    \item For some vector $\vec{v}$ and $\vec{u}$, $\langle \vec{v}, \vec{u} \rangle$ denotes the inner product of $\vec{v}$ and $\vec{u}$.
    \item For $a, b \in \mathbb{R}$, we denote $(a, b)^+ = \max(a, b)$ and $(a, b)^- = \min(a, b)$.
    \item $Im(z)$ for $z \in \mathbb{C}$ denotes the imaginary part of $z$. 
    \item $a \simeq b$ denotes $a$ converges almost surely to $b$.
    \item $\mathds{1}_{\text{condition}}$ denotes the indicator function, which is $1$ if the condition is satisfied, and $0$ otherwise.
\end{itemize} 
\paragraph{Big-O Notation} Throughout this work, we use standard asymptotic notation. Specifically, $O(\cdot)$ denotes an upper bound up to constant factors---that is, a quantity that grows no faster than the reference function. In contrast, $\Theta(\cdot)$ denotes a tight asymptotic bound, meaning the quantity grows at the same rate as the reference function, up to constant factors. Finally, $o(\cdot)$ denotes a lower-order term that becomes negligible compared to the reference term in the asymptotic regime.

\subsection{Singular Value Decomposition(SVD)}
For an arbitrary matrix $\mat{M} \in \mathbb{R}^{d \times n}$, 
\begin{itemize}
    \item $\mat{U}_\mat{M} \mat{\Sigma_M V_M}^\top$ denotes the SVD of $\mat{M}$, where $\mat{U_M} \in \mathbb{R}^{d \times d}$, $\mat{\Sigma_M} \in \mathbb{R}^{d \times n}$, and $\mat{V_M} \in \mathbb{R}^{n \times n}$.
    \item $\vec{u}^\mat{M}_i$ and $\sigma^\mat{M}_i$ are the $i$-th column vector of $\mat{U_M}$ and the $i$-th singular value of $\mat{M}$, respectively.
    \item $\mat{\tilde{\mat{U}}_M D_M \tilde{\mat{V}}_M}^\top = \mat{U}_{\mat{M}, r_\mat{M}} \mat{D}_\mat{M} \mat{V}_{\mat{M}, r_\mat{M}}$ is the \textit{reduced SVD} of $\mat{M}$, where $\tilde{\mat{U}}_M := \mat{U}_{\mat{M}, r_\mat{M}}$, $\tilde{\mat{V}}_M := \mat{V}_{\mat{M}, r_\mat{M}}$, 
    and $\mat{D}_\mat{M} \in \mathbb{R}^{r_\mat{M} \times r_\mat{M}}$ is the diagonal matrix with the singular values of $\mat{M}$, i.e., $(\mat{D}_\mat{M})_{ij} = \sigma^\mat{M}_i \delta_{ij}$, for $\delta$ the Dirac Delta function.
\end{itemize}

For an index set $I \subset \{1, \dots, r_\mat{M}\}$, we use 
\begin{itemize}
    \item $\hat{\mat{\Sigma}}_{\mat{M}, I} \in \mathbb{R}^{d \times n}$ to denote the matrix with its diagonal part consists of the singular values of $\mat{M}$ indexed by $I$. 
    Precisely, 
    $$ (\hat{\mat{\Sigma}}_{\mat{M}, I})_{ij} = \begin{cases}
        \sigma^\mat{M}_i & \text{if } i = j \text{ and } i \in I \\
        0 & \text{otherwise}
    \end{cases}
    $$
    \item $P_{I}(\mat{M}) := \mat{U}_{\mat{M}} \hat{\mat{\Sigma}}_{\mat{M}, I} \mat{V}_{\mat{M}}^\top = \mat{U}_{\mat{M}, I} \mat{D}_{\hat{\mat{\Sigma}}_\mat{M}} \mat{V}_{\mat{M}, I}^\top$ .
\end{itemize}
For example, when applying Eckart-Young Theorem \cite{eckart1936approximation}, we need the best rank-$q$ approximation of a matrix $\mat{M}$. 
For this special situation, we denote $P_{[q]}(\mat{M})$ as $P_q(\mat{M})$.
Then, it can be written as $P_q(\mat{M}) = \mat{U}_\mat{M} \hat{\mat{\Sigma}}_{\mat{M}, q} \mat{V}_\mat{M}^\top = \mat{U}_{\mat{M}, q} \mat{D}_{\mat{M}, q} \mat{V}_{\mat{M}, q}^\top$. \\
Additionally, for a symmetric matrix $\mat{S} \in \mathbb{R}^{d \times d}$, 
\begin{itemize}
    \item $\mat{S} = \mat{U}_\mat{S} \mat{\Lambda_S U_S}^\top$ denotes the eigendecomposition of $\mat{S}$, where $\mat{U_S} \in \mathbb{R}^{d \times d}$ and $\mat{\Lambda_S} \in \mathbb{R}^{d \times d}$.
    \item $\lambda_i(\mat{S})$ or $\lambda^\mat{S}_i$ denotes the $i$-th eigenvalue of $\mat{S}$.
\end{itemize}
\section{Background on Random Matrix Theory}\label{appendix:rmt_background}
Here, we present the notations and key tools from Random Matrix Theory that are frequently referenced in Appendix~\ref{appendix:rmt_results}.
We include only the essential definitions and theorems, and refer the reader to~\cite{couillet_liao_2022, bai2010spectral} for more comprehensive treatments.

\subsection{Getting Eigenvalue Information}
We consider a Gaussian random matrix $\tilde{\mat{A}} \in \mathbb{R}^{d \times n}$, whose elements are independently and identically distributed(i.i.d) following 
a gaussian distribution of mean $0$ and variance $\sigma^2$, i.e., $\mathcal{N}(0, \sigma^2)$. 
The goal is to analyze \textit{the behavior of the eigenvalues} of $\frac{1}{n} \tilde{\mat{A}} \tilde{\mat{A}}^T$ as $d, n$ grows together toward infinity, 
with their ratio satisfying $\frac{d}{n} \rightarrow c \in (0, \infty)$.
Notably, the normalized histogram of the eigenvalues of $\frac{1}{n} \tilde{\mat{A}} \tilde{\mat{A}}^T$ exhibits \textit{deterministic} behavior 
as $d, n$ grows, converging weakly to the Marchenko-Pastur distribution, which is denoted as $\mu_{MP}$. 
This can be written as follows.
\begin{align}\label{appendix:rmt_background:histogram}
    \frac{1}{d} \sum\limits_{i = 1}^{d} \delta_{\lambda_i(\frac{1}{n} \tilde{\mat{A}} \tilde{\mat{A}}^T)} \rightarrow \mu_{MP} \text{ weakly}.
\end{align}
To characterize this distribution, the \textit{Stieltjes transform} is one of the most commonly used tools.
We introduce the following definitions, following~\cite[Chapter 2]{couillet_liao_2022}.
\begin{definition} \label{appendix:rmt_background:definitions}
    \begin{enumerate}
        \item \textnormal{(Resolvent).} For a symmetric matrix $\mat{M} \in \mathbb{R}^{d \times d}$, the resolvent of $\mat{M}$ is defined as 
        \begin{align*}
            \mat{Q}_\mat{M}(\alpha) = (\mat{M} - \alpha \mathbb{I}_d)^{-1}, \quad \alpha \in \mathbb{C} \backslash \{\lambda_1^\mat{M}, \dots , \lambda_d^\mat{M}\}.
        \end{align*}
        \item \textnormal{(Empirical Spectral Measure).} The above eq.~(\ref{appendix:rmt_background:histogram}) is formally known as Empirical Spectral Measure, and 
        for a symmetric matrix $\mat{M} \in \mathbb{R}^{d \times d}$, it is defined as
        \begin{align*}
            \mu_\mat{M} = \frac{1}{d} \sum\limits_{i = 1}^{d} \delta_{\lambda_i^\mat{M}}.
        \end{align*}
        \item \textnormal{(Stieltjes Transform).} For a real probability measure $\mu$ with support $\textnormal{supp}(\mu)$, 
        the Stieltjes Transform of $\mu$ is defined as 
        \begin{align*}
            m_{\mu}(\alpha) = \int \frac{1}{x - \alpha} \,d\mu(x), \quad \alpha \in \mathbb{C} \backslash \textnormal{supp}(\mu).
        \end{align*}
    \end{enumerate}
\end{definition}

Note that the stieltjes transform of the \textit{empirical spectral measure} is given by,
\begin{align*}
    m_{\mu_{\mat{M}}}(\alpha) &= \int \frac{1}{x - \alpha} \,d\mu_\mat{M}(x) \\
    &= \frac{1}{d} \sum\limits_{j = 1}^d \frac{1}{\lambda_j^\mat{M} - \alpha} \\
    &= \frac{1}{d} \Tr(\mat{Q}_\mat{M}).
\end{align*}

The stieltjes transform is used to characterize 
the following convergence.
\begin{align*}
    \frac{1}{d} \Tr(\mat{Q}_{\frac{1}{n}\tilde{\mat{A}}\tilde{\mat{A}}^T}) \rightarrow \int \frac{1}{z - \alpha} \,d\mu_{MP}(z) =: m_{\mu_{MP}}.
\end{align*}
Once we characterize how the stieltjes transform $m_{\mu_{MP}}$ looks like, 
using the \textit{Inverse Stieltjes Transform} (\cite[Theorem 2.1]{couillet_liao_2022}), the Marchenko-Pastur density $\mu_{MP}$ can be retrieved.
\begin{theorem}\textnormal{(Marcenko-Pastur Law, \cite[Theorem 3.1.1]{bai2010spectral}).}
    \label{appendix:rmt_background:mp_law}
    For a random matrix $\tilde{\mat{A}} \in \mathbb{R}^{d \times n}$, whose entries are i.i.d random variables with mean $0$ and finite variance $\sigma^2$, 
    the empirical spectral measure of $\frac{1}{n} \tilde{\mat{A}} \tilde{\mat{A}}^T$ converges weakly to $\mu_{MP}$, 
    where 
    \begin{align*}
        \mu_{MP}(dx) = (1 - \frac{1}{c})^+\delta_0(x) + \frac{1}{2\pi \sigma^2 c x}\sqrt{(b - x)(x - a)}\mathds{1}_{x \in [a, b]},
    \end{align*}
    for the ratio $c$ satisfies $\frac{d}{n} \rightarrow c \in (0, \infty)$, $a := \sigma^2(1 - \sqrt{c})^2$, $b := \sigma^2(1 + \sqrt{c})^2$. 
\end{theorem}
The corresponding stieltjes transform $m_{\mu_{MP}}$ satisfies the following quadratic equation: 
\begin{equation}
    c \alpha \sigma^2 m_{\mu_{MP}}^2 - (\sigma^2(1 - c) - \alpha)\ m_{\mu_{MP}} + 1 = 0.
\end{equation}
The following lemma connects the resolvent of the sample covariance matrix $\mat{Q}_{\frac{1}{n}\tilde{\mat{A}}\tilde{\mat{A}}^T}$ to $m_{\mu_{MP}}$.
\begin{lemma}(Due to \cite[Theorem 2.4]{couillet_liao_2022})
    \label{chap:background:rmt:connection_lemma}
    For unit vectors $\vec{a}, \vec{b} \in \mathbb{R}^d$, it satisfies that
    \begin{align*}
        \vec{a}^T \mat{Q}_{\frac{1}{n}\tilde{\mat{A}}\tilde{\mat{A}}^T} \vec{b} \simeq m_{\mu_{MP}} \vec{a}^T \vec{b}.
    \end{align*}
\end{lemma}

We state important lemmas further that are frequently referred in Appendix~\ref{appendix:rmt_results}. 
\begin{lemma}
    \label{appendix:rmt_background:lemmata} 
    \begin{enumerate}
        \item \textnormal{(Resolvent Identity, \cite[Chapter 2]{couillet_liao_2022}).}
            Let $\mat{A}, \mat{B}$ invertible matrices. Then, $\mat{A}^{-1} - \mat{B}^{-1} = \mat{A}^{-1}(\mat{B} - \mat{A})\mat{B}^{-1}$. 
        \item \textnormal{(Sherman-Morrison Formula).}
            Let $\mat{A} \in \mathbb{R}^{p \times p}$, and $\vec{x}, \vec{y} \in \mathbb{R}^p$. For $1 + \vec{y}^T\mat{A} \vec{x} \neq 0$, we have that 
            $(\mat{A} + \vec{x}\vec{y}^T)^{-1} = \mat{A}^{-1} - \frac{\mat{A}^{-1}\vec{x}\vec{y}^T\mat{A}^{-1}}{1 + \vec{y}^T\mat{A}\vec{x}}$.
        \item \textnormal{(Concentration of Quadratic Forms, Gaussian case).} 
            For $\alpha \in \mathbb{C} \backslash \mathbb{R}_+$, let $\tilde{\mat{Q}}(\alpha) = \left(\tilde{\mat{A}}\tilde{\mat{A}}^T - \alpha \mathbb{I}_d\right)^{-1} = \left(\sum\limits_{i=1}^N \tilde{\vec{a}}_i \tilde{\vec{a}}_i^T - \alpha \mathbb{I}_d\right)^{-1}$, 
            where $\tilde{A} \in \mathbb{R}^{d \times N}$ is a i.i.d real gaussian random matrix, whose entries are sampled from $\mathcal{N}(0, 1)$.
            Let $\tilde{\mat{Q}}_{-j}(\alpha)$ be $\tilde{\mat{Q}}(\alpha)$ with $j$-th row and column removed. Then, $\frac{1}{d}\tilde{\vec{a}}_j^T \tilde{\mat{Q}}_{-j} \tilde{\vec{a}}_j \xrightarrow{a.s} \frac{1}{d}\Tr(\tilde{\mat{Q}}_{-j})$.

            (Proof.) \\
            We have from Hanson-Wright Inequality \cite{vershynin2018high}, that for $\epsilon > 0$, there exists $C > 0$ such that
            \begin{align*}
                \mathbb{P}\left(\left|\frac{1}{d}\tilde{\vec{a}}_j^T \tilde{\mat{Q}}_{-j} \tilde{\vec{a}}_j - \frac{1}{d}\mathbb{E}\left[\tilde{\vec{a}}_j ^T \tilde{\mat{Q}}_{-j} \tilde{\vec{a}}_j \right]\right| > \epsilon\right) 
                &\leq 2\exp(-\epsilon d C \|\tilde{\mat{Q}}_{-j}\|_2^{-1}).
            \end{align*}
            Due to $\|\tilde{\mat{Q}}_{-j}\|_2^{-1} = \left(\min\limits_{1 \leq k \leq d}|\lambda_k^{\tilde{\mat{A}}\tilde{\mat{A}}^T} - \alpha|\right)^{-1} < \infty$, 
            \begin{align*}
                &\lim\limits_{d \rightarrow \infty} \sum\limits_{n=1}^d \mathbb{P}\left(\left|\frac{1}{n}\tilde{\vec{a}}_j^T \tilde{\mat{Q}}_{-j} \tilde{\vec{a}}_j - \frac{1}{n}\mathbb{E}[\tilde{\vec{a}}_j ^T \tilde{\mat{Q}}_{-j} \tilde{\vec{a}}_j ]\right| > \epsilon\right) \\
                &= \lim\limits_{d \rightarrow \infty} \sum\limits_{n=1}^d \left(2\exp(-\epsilon n C \|\tilde{\mat{Q}}_{-j}\|_2^{-1})\right) < \infty.
            \end{align*}
            The application of Borel-Cantelli Lemma\cite[Theorem 2.3.1]{durrett2019probability} gives us that $\frac{1}{d} \tilde{\vec{a}}_j^T \tilde{\mat{Q}}_{-j} \tilde{\vec{a}}_j \xrightarrow{a.s} \frac{1}{d}\mathbb{E}[\tilde{\vec{a}}_j^T \tilde{\mat{Q}}_{-j} \tilde{\vec{a}}_j] = \frac{1}{d}\Tr(\tilde{\mat{Q}}_{-j})$.
            
            \rightline{$\square$}
        \item \textnormal{(Minimal Effect of Finite-Rank Perturbations Inside a Trace, due to~\cite[Lemma 2.6]{SILVERSTEIN1995175}).} \\
            For symmetric and positive semi-definite $\mat{B}, \mat{M} \in \mathbb{R}^{d \times d}$, and $\alpha \in \mathbb{C} \backslash \mathbb{R}_+$, let $\mat{M} := \sum\limits_{j=1}^h l_i \mat{x}_i \mat{x}_i^T$ for fixed $h$, 
            and $l_i \in \mathbb{R}$, $\mat{x}_i \in \mathbb{R}^d$. Then, we have that 
            \begin{align*}
                |\Tr\left((\mat{B} - \alpha \mathbb{I})^{-1} - (\mat{B} + \mat{M} - \alpha \mathbb{I})^{-1}\right)| \leq \frac{h}{Im(\alpha)}.
            \end{align*}

            (Proof.) \\
            The proof is a simple recursive argument of \cite[Lemma 2.6]{SILVERSTEIN1995175}, which states that 
            \begin{align*}
                \left|\Tr\left((\mat{B} - \alpha \mathbb{I})^{-1} - (\mat{B} + l \mathbf{x} \mathbf{x}^T - \alpha \mathbb{I})^{-1}\right)\right| \leq \frac{1}{Im(\alpha)},
            \end{align*}
            for some $l \in \mathbb{R}$, and $\mathbf{x} \in \mathbb{R}^d$. We repeat this $h$ times to get the result.

            \rightline{$\square$}

        \item \textnormal{(Weinstein-Aronszajn Identity).} 
            For $\mat{A} \in \mathbb{R}^{p \times q}, \mat{B} \in \mathbb{R}^{q \times p}$, and $\lambda \in \mathbb{R} \backslash \{0\}$. Then, we have that
            \begin{align*}
                det(\mat{A}\mat{B} - \lambda \mathbb{I}_p) = (- \lambda)^{p - q} det(\mat{B}\mat{A} - \lambda \mathbb{I}_q).
            \end{align*}
        
        \item \textnormal{(Woodbury Identity).}
            For $\mat{A} \in \mathbb{R}^{p \times p}$, $\mat{U} \in \mathbb{R}^{p \times q}$, $\mat{V} \in \mathbb{R}^{q \times p}$,
            we have that 
            $$
            (\mat{A} + \mat{U}\mat{V}^T)^{-1} = \mat{A}^{-1} - \mat{A}^{-1}\mat{U}(\mathbb{I}_q + \mat{V}^T\mat{A}^{-1}\mat{U})^{-1}\mat{V}^T\mat{A}^{-1}.
            $$
    \end{enumerate}
\end{lemma}

\subsection{Getting Subspace Information}\label{appendix:rmt_background:subspace_information}
One useful property of the resolvent and the Stieltjes transform is their connection to the eigenstructure of a random matrix.
This connection stems from their resemblance to the Cauchy integral formula.

\begin{theorem}\textnormal{(Cauchy Integral Formula \cite[Thm. 6]{ahlfors1979complex}).}
    For a complex analytic function $f(z)$ in a simply connected domain $D$ with a simple pole at $z_0 \in D$, 
    it satisfies that 
    \begin{align*}
        f(z_0) = \frac{1}{2\pi i} \oint_{\gamma} \frac{f(z)}{z - z_0} \,dz,
    \end{align*}
    where $\gamma$ is a positively oriented simple closed curve around $z_0$.
\end{theorem}
By using the orthogonal decomposition of $\frac{1}{n} \tilde{\mat{A}} \tilde{\mat{A}}^T = \mat{U}_{\tilde{\mat{A}}} \mat{\Lambda}_{\tilde{\mat{A}}} \mat{U}_{\tilde{\mat{A}}}^T$,
 $\mat{Q}_{\frac{1}{n} \tilde{\mat{A}} \tilde{\mat{A}}^T}$ can be written down as 
\begin{align*}
    \mat{Q}_{\frac{1}{n} \tilde{\mat{A}} \tilde{\mat{A}}^T} &= \sum\limits_{j = 1}^{d} \frac{1}{\lambda_j\left(\frac{1}{n} \tilde{\mat{A}} \tilde{\mat{A}}^T\right) - \alpha} \vec{u}^{\tilde{\mat{A}}}_j (\vec{u}^{\tilde{\mat{A}}}_j)^T.
\end{align*}
By defining $\Gamma_{\lambda_j}$ as a positively oriented simple closed curve that 
\textit{only} encloses $\lambda_j\left(\frac{1}{n} \tilde{\mat{A}} \tilde{\mat{A}}^T\right)$, 
we can express the eigenvector information of the random matrix as follows. 
\begin{align*}
    \vec{u}^{\tilde{\mat{A}}}_j (\vec{u}^{\tilde{\mat{A}}}_j)^T = - \frac{1}{2\pi i} \oint_{\Gamma_{\lambda_j}} \mat{Q}_{\frac{1}{n} \tilde{\mat{A}} \tilde{\mat{A}}^T} \,d\alpha.
\end{align*}
For example, for a vector $\vec{v} \in \mathbb{R}^d$, the projection information of $\vec{u}^{\tilde{\mat{A}}}_j$ onto $\vec{v}$ is given by
\begin{align*}
    \vec{v}^T \vec{u}^{\tilde{\mat{A}}}_j  (\vec{u}^{\tilde{\mat{A}}}_j)^T \vec{v} = \langle \vec{v}, \vec{u}^{\tilde{\mat{A}}}_j \rangle^2= - \frac{1}{2\pi i} \oint_{\Gamma_{\lambda_j}} \vec{v}^T \mat{Q}_{\frac{1}{n} \tilde{\mat{A}} \tilde{\mat{A}}^T} \vec{v} \,d\alpha.
\end{align*}

\subsection{Spiked Models}
\label{appendix:rmt_background:spiked_models}
The denoising problem considered in this work is closely related to the \textit{Spiked Models} in Random Matrix Theory. 
The authors \cite{couillet_liao_2022} have categorized the following models(among others) as \textit{Additive Models}, 
emphasizing their common \textit{additive structures.}

\begin{definition}\textnormal{(Spiked-Covariance Model, due to \cite{johnstone}).}
    For a rank $r$ deterministic, and symmetric matrix $\mat{P} \in \mathbb{R}^{d \times d}$, 
    we consider a covariance matrix $\mat{C} \in \mathbb{R}^{d \times d}$, which is defined as:
    \begin{align*}
        \mat{C} = \sum\limits_{j = 1}^r \lambda_i \vec{u}^\mat{P}_j (\vec{u}^\mat{P}_j)^T + \mathbb{I}_d.
    \end{align*}
    Then, for a random matrix $\mat{A}_1 \in \mathbb{R}^{d \times n}$, whose columns are i.i.d random vectors sampled from a distribution 
    with mean $0$ and covariance $\mat{C}$, 
    the spiked-covariance model is defined as 
    $$\frac{1}{n} \mat{A}_1 \mat{A}_1^T.$$
\end{definition}

\begin{definition}(Information-plus-Noise Model)\label{appendix:rmt_background:information_plus_noise_model}
    For a rank $r$ deterministic matrix $\mat{X}_2 \in \mathbb{R}^{d \times n}$, and an i.i.d random matrix $\mat{A}_2 \in \mathbb{R}^{d \times n}$,
    the Information-plus-Noise model is defined as $$\frac{1}{n} (\mat{X}_2 + \mat{A}_2)(\mat{X}_2 + \mat{A}_2)^T.$$
\end{definition}

\begin{definition}(Additive Model)\label{appendix:rmt_background:additive_model}
    For a rank $r$ deterministic, and symmetric matrix $\mat{P}_3 \in \mathbb{R}^{d \times d}$, and an i.i.d random matrix $\mat{A}_3 \in \mathbb{R}^{d \times n}$, 
    the Additive Model is defined as $$\mat{P}_3 + \frac{1}{n} \mat{A}_3 \mat{A}_3^T.$$
\end{definition}

These three models share a common structure, in that they all involve \textit{adding the low-rank data to the random matrices}. 
Our denoising problem corresponds precisely to the Information-plus-Noise model. 
In Appendix \ref{appendix:rmt_results:main_proof}, we will leverage this structural similarity between these models to derive an interesting result, 
by alternatively considering the Additive Model.


\section{Characterization of Critical Points}\label{appendix:critical_points}
In this section, we first state and prove a detailed version of Theorem~\ref{section:solutions:theorem1} (see Appendix~\ref{appendix:section2:subsection1}).
In addition, we show that the global minimizer it identifies is equivalent to the solution obtained via reduced-rank regression under the minimum-norm principle, as shown in Appendix~\ref{appendix:section2:equivalence_rrr}.
\subsection{Proofs of Section \ref{section:preliminaries}}
\label{appendix:section2:subsection1}
Before stating the main theorem, we present two auxiliary lemmas.  
The first lemma establishes necessary conditions for critical points and their consequences, while the second lemma is important for characterizing each critical point.
To facilitate the proofs of these lemmas, we introduce the following notation:  
let \(\mat{U}_{\mat{G}} = \begin{bmatrix} \mat{U}_1 & \mat{U}_2 \end{bmatrix}\), where \(\mat{U}_1 \in \mathbb{R}^{d \times r_{\mat{G}}}\) and \(\mat{U}_2 \in \mathbb{R}^{d \times (\mat{D} - r_{\mat{G}})}\).
Additionally, for any matrix \(\mat{M}\), we denote by \(P_{\mat{M}}\) the orthogonal projection matrix onto the column space of \(\mat{M}\).

\begin{lemma}\textnormal{(Necessary Conditions for Critical Points).}\label{appendix:section2:lemma1}
    Consider the objective of minimizing (\ref{section2:general_loss}). 
    Then, the necessary conditions for critical points $\hat{\mat{W}}_2, \hat{\mat{W}}_1$ are given as:
    \begin{equation}\label{appendix:section2:lemma1:necessary1}
        \mat{Y} \mat{Z}^\top \hat{\mat{W}}_1^\top = \hat{\mat{W}}_2\hat{\mat{W}}_1\tilde{\mat{Z}}\tilde{\mat{Z}}^\top\hat{\mat{W}}_1^\top.
    \end{equation}
    \begin{equation}\label{appendix:section2:lemma1:necessary2}
        \hat{\mat{W}}_2^\top \mat{Y} \mat{Z}^\top = \hat{\mat{W}}_2^\top\hat{\mat{W}}_2\hat{\mat{W}}_1 \tilde{\mat{Z}} \tilde{\mat{Z}}^\top.
    \end{equation}
   The following equations are implied by the necessary conditions: \begin{equation}\label{appendix:section2:lemma1:consequence1}
        \hat{\mat{W}} := \hat{\mat{W}}_2 \hat{\mat{W}}_1 = P_{\hat{\mat{W}}_2} \mat{Y} \mat{Z}^\top (\tilde{\mat{Z}} \tilde{\mat{Z}}^\top)^{-1},
    \end{equation}
    \begin{equation}\label{appendix:section2:lemma1:consequence2}
        P_{\hat{\mat{W}}_2} \mat{G} = \mat{G} P_{\hat{\mat{W}}_2} = P_{\hat{\mat{W}}_2} \mat{G} P_{\hat{\mat{W}}_2}.
    \end{equation}
\end{lemma}

\textit{Proof.\quad}
The gradient flows of weight matrices $\mat{W}_1, \mat{W}_2$ are:
\begin{align}
    \frac{d\mat{W}_1}{dt} &= \mat{W}_2^\top(\mat{Y}\mat{Z}^\top - \mat{W}_2\mat{W}_1\mat{Z} \mat{Z}^\top - \lambda \mat{W}_2\mat{W}_1) \nonumber \\
    &= \mat{W}_2^\top \mat{Y} \mat{Z}^\top - \mat{W}_2^\top\mat{W}_2\mat{W}_1 \tilde{\mat{Z}} \tilde{\mat{Z}}^\top, \nonumber
\end{align}
\begin{align}
    \frac{d\mat{W}_2}{dt} &= (\mat{Y}\mat{Z}^\top - \mat{W}_2\mat{W}_1 \mat{Z} \mat{Z}^\top - \lambda \mat{W}_2\mat{W}_1)\mat{W}_1^\top \nonumber \\
    &= \mat{Y} \mat{Z}^\top \mat{W}_1^\top - \mat{W}_2\mat{W}_1\tilde{\mat{Z}}\tilde{\mat{Z}}^\top\mat{W}_1^\top. \nonumber
\end{align}
In order to find critical points, we set both of the gradient flows to zero.
For the first condition, by taking the generalized inverse of $\mat{W}_2^\top\mat{W}_2$, we have that 
for any $\mat{W}_1 \in \mathbb{R}^{k \times d}$, the following holds:
\begin{align}
    \mat{W}_1 = (\mat{W}_2^\top\mat{W}_2)^{\dagger}\mat{W}_2^\top \mat{Y} \mat{Z}^\top(\tilde{\mat{Z}} \tilde{\mat{Z}}^\top)^{-1} 
    + (\mathbb{I} - (\mat{W}_2^\top\mat{W}_2)^{\dagger} \mat{W}_2^\top\mat{W}_2)\mat{L}_1 \nonumber
\end{align}
For the second condition $\frac{d\mat{W}_2}{dt} = 0$, this is equivalent to 
\begin{align}\label{lemma1:second_necessary_condition}
    \mat{Y} \mat{Z}^\top \mat{W}_1^\top = \mat{W}_2\mat{W}_1 \tilde{\mat{Z}} \tilde{\mat{Z}}^\top \mat{W}_1^\top.
\end{align} 
For a critical point $\hat{\mat{W}} := \hat{\mat{W}}_2 \hat{\mat{W}}_1$, it holds that 
\begin{align}\label{lemma1:critical_points}
    \hat{\mat{W}} &= \hat{\mat{W}}_2 \hat{\mat{W}}_1 \nonumber \\
    &= \hat{\mat{W}}_2(\hat{\mat{W}}_2^\top\hat{\mat{W}}_2)^{\dagger}\hat{\mat{W}}_2^\top \mat{Y} \mat{Z}^\top (\tilde{\mat{Z}} \tilde{\mat{Z}}^\top)^{-1} + (\hat{\mat{W}}_2 - \hat{\mat{W}}_2(\hat{\mat{W}}_2^\top\hat{\mat{W}}_2)^{\dagger} \hat{\mat{W}}_2^\top\hat{\mat{W}}_2)\mat{L}_1 \nonumber \\
    &= P_{\hat{\mat{W}}_2} \mat{Y} \mat{Z}^\top (\tilde{\mat{Z}} \tilde{\mat{Z}}^\top)^{-1}.
\end{align}
Then, it follows from the second condition (\ref{lemma1:second_necessary_condition}) that
$\mat{Y} \mat{Z}^\top \hat{\mat{W}}^\top = \hat{\mat{W}} (\tilde{\mat{Z}} \tilde{\mat{Z}}^\top)^{-1} \hat{\mat{W}}^\top$, by multiplying $\hat{\mat{W}}_2^\top$ on both sides. 
From eq.~(\ref{lemma1:critical_points}), we have that 
\begin{align*}
    P_{\hat{\mat{W}}_2}\mat{G} P_{\hat{\mat{W}}_2} = P_{\hat{\mat{W}}_2} \mat{G} = \mat{G} P_{\hat{\mat{W}}_2}.
\end{align*} $P_{\hat{\mat{W}}_2} \mat{G} = \mat{G} P_{\hat{\mat{W}}_2}$ is due to $P_{\hat{\mat{W}}_2} = P_{\hat{\mat{W}}_2}^\top$.

\rightline{$\square$}

\begin{lemma} \label{appendix:section2:lemma2}
    Following from Lemma \ref{appendix:section2:lemma1}, it holds that
    \begin{equation}
        P_{\mat{U}_\mat{G}^\top \hat{\mat{W}}_2} = \begin{bmatrix}
            \mathcal{D} & \mat{0} \\
            \mat{0} & \mat{B}
        \end{bmatrix},
    \end{equation} where $\mathcal{D}$ and $\mat{W}$ satisfy the followings.
    \begin{enumerate}
        \item $\mat{B} \in \mathbb{R}^{(\mat{D} - r_\mat{G}) \times (\mat{D} - r_\mat{G})}$ is defined as $\mat{B} := \mat{B}_1 \mat{B}_1^\top$, where 
        \begin{equation*}
        \mat{B}_1 = ((\mat{U}_\mat{G}^\top\mat{U}_{\hat{\mat{W}}_2})_{i, j})_ {r_{\mat{G}} + 1 \leq i \leq d, r_{\mat{G}} + 1 \leq j \leq d}. 
        \end{equation*}
        Furthermore, $\mat{B}$ has $r_\mat{B}$ eigenvalues equal to $1$ and the rest equal to $0$. 
        \item $\mathcal{D} \in \mathbb{R}^{r_\mat{G} \times r_\mat{G}}$ is a diagonal matrix with its diagonal elements consist of $r_{\mathcal{D}}$ number of $1$'s and $r_\mat{G} - r_{\mathcal{D}}$ number of $0$'s.
        \item $r_{\mathcal{D}} + r_\mat{B} = r_{\hat{\mat{W}}_2}$.
    \end{enumerate}
\end{lemma}

\textit{Proof.\quad}
We analyze the conditions (\ref{appendix:section2:lemma1:consequence1}) and (\ref{appendix:section2:lemma1:consequence2}) further. 
Consider the eigendecomposition of $\mat{G}$, given by $\mat{G} = \mat{U}_\mat{G} \Lambda_\mat{G} \mat{U}_\mat{G}^\top$, and the singular value decomposition of $\hat{\mat{W}}_2$, given by $\hat{\mat{W}}_2 = \mat{U}_{\hat{\mat{W}}_2} \Sigma_{\hat{\mat{W}}_2} \mat{V}_{\hat{W}_2}^\top.$
Then, 
\begin{equation}\label{appendix:section2:lemma2:eq1}
P_{\mat{U}_\mat{G}^\top\hat{\mat{W}}_2} = \mat{U}_\mat{G}^\top\hat{\mat{W}}_2 (\mat{U}_\mat{G}^\top\hat{\mat{W}}_2)^{\dagger} = \mat{U}_\mat{G}^\top \hat{\mat{W}}_2\hat{\mat{W}}_2^{\dagger} \mat{U}_\mat{G} = \mat{U}_\mat{G}^\top P_{\hat{\mat{W}}_2} \mat{U}_\mat{G}.
\end{equation}
Then, it satisfies that $P_{\hat{\mat{W}}_2} = \hat{\mat{W}}_2\hat{\mat{W}}_2^{\dagger} = \mat{U}_\mat{G} P_{\mat{U}_\mat{G}^\top\hat{\mat{W}}_2} \mat{U}_\mat{G}^\top$. From (\ref{appendix:section2:lemma1:consequence2}), we have 
$P_{\hat{\mat{W}}_2} \mat{G} = \mat{G} P_{\hat{\mat{W}}_2}$. By combining the two equations, we obtain $P_{\mat{U}_\mat{G}^\top\hat{\mat{W}}_2} \Lambda = \Lambda P_{\mat{U}_\mat{G}^\top\hat{\mat{W}}_2}$. 
Combining this with the fact that $P_{\hat{\mat{W}}_2} = U_{\hat{\mat{W}}_2}\begin{bmatrix}
    \mathbb{I}_{r_{\hat{\mat{W}}_2}} & 0 \\
    0 & 0
  \end{bmatrix} U_{\hat{\mat{W}}_2}^\top$, we obtain that $P_{\mat{U}_\mat{G}^\top\hat{\mat{W}}_2} = \begin{bmatrix}
    \mathcal{D} & 0 \\
    0 & \mat{B}
  \end{bmatrix}$, where $\mathcal{D} \in \mathbb{R}^{r_{\mat{G}} \times r_{\mat{G}}}, \mat{B} \in \mathbb{R}^{(\mat{D} - r_\mat{G}) \times (\mat{D} - r_{\mat{G}})}$, 
  and $r_{\mathcal{D}} + r_\mat{B} = r_{\mat{W}_2} \leq k$. Note that $P_{\mat{U}_{\mat{G}}^\top \hat{\mat{W}}_2}$ has eigenvalues consist of $r_{\hat{\mat{W}}_2}$ number of $1$s and the rest $0$. Using this fact, a straightforward calculation shows that $\mathcal{D}$ is a diagonal matrix with $r_{\mathcal{D}}$ number of $1$s, 
  and the rest $0$s. For $\mat{B} = \mat{B}_1 \mat{B}_1^\top$ where $\mat{B}_1 = ((\mat{U}_{\mat{G}}^\top\mat{U}_{\hat{W}_2})_{i, j})_ {r_{\mat{G}} + 1 \leq i \leq d, r_{\mat{G}} + 1 \leq j \leq d}$, 
   its eigenvalues consist of $r_\mat{B}$ number of $1$s, and the rest $0$s. 

\rightline{$\square$}

\paragraph{Full Statement of Theorem \ref{section:solutions:theorem1}}
From Lemma \ref{appendix:section2:lemma2}, we established that $\mat{B}$ is a symmetric positive semi-definite matrix with its eigenvalues $1$ or $0$. Thus we can write down its eigendecomposition as $\mat{B} = \tilde{\mat{U}}_{\mat{B}_1}\tilde{\mat{U}}_{\mat{B}_1}^\top$. With this notation, we state the following theorem.

\begin{theorem}\textnormal{(Full Version of Theorem \ref{section:solutions:theorem1}).}
    \label{appendix:section2:full_theorem}
    Assume that the multiplicity of non-zero eigenvalues of $\mat{G}$ is $1$, 
    i.e., $\sigma^\mat{G}_1 > \sigma^\mat{G}_2 > \cdots > \sigma^\mat{G}_{r_\mat{G}} > 0$. 
    We further assume that the bottleneck dimension $k$ is chosen such that $k \leq r_\mat{G}$\footnote{The condition $k \leq r_\mat{G}$
      is imposed to analyze the effect of the bottleneck layer, which is the focus of this work. 
      This assumption can be relaxed to the general case, albeit at the cost of losing the uniqueness of the global minimizer.}.
    Let $\mathcal{I}$ be a family of index sets, where each element is an ordered set of distinct natural numbers from $[k]$. 
    Then, we establish the following results. 
    \begin{enumerate}
        \item Let $(I, \mat{B})$ a tuple, such that $|I| + r_\mat{B} = k$, for $I \in \mathcal{I}$.
        Then, for $\hat{\mat{W}}_2$ and $\hat{\mat{W}}_1$ define critical points for (\ref{section2:general_loss}) 
        if and only if there exist an invertible matrix $\mat{C} \in \mathbb{R}^{k \times k}$
        and a tuple $(I, \mat{B})$ such that $\hat{\mat{W}}_2$ and $\hat{\mat{W}}_1$ satisfy that: \\
        for $|I| < k$,
        \begin{align*}
            &\hat{\mat{W}}_2 = \begin{bmatrix} \mat{U}_{\mat{G}, I} & \mat{U}_2 \tilde{\mat{U}}_{\mat{B}_1} \end{bmatrix} \mat{C}, \\ 
            &\hat{\mat{W}}_1 = \mat{C}^{-1} \begin{bmatrix} \mat{U}_{\mat{G}, I} & \mat{U}_2\tilde{\mat{U}}_{\mat{B}_1} \end{bmatrix}^\top \mat{Y} \mat{Z}^\top (\tilde{\mat{Z}}\tilde{\mat{Z}}^\top)^{-1},
        \end{align*} 
        and for $|I| = k$,
        \begin{align*}
            &\hat{\mat{W}}_2 = \mat{U}_{\mat{G}, I} \mat{C}, \\
            &\hat{\mat{W}}_1 = \mat{C}^{-1} \mat{U}_{\mat{G}, I}^\top \mat{Y} \mat{Z}^\top (\tilde{\mat{Z}}\tilde{\mat{Z}}^\top)^{-1}.
        \end{align*}
        For the both cases, assume $r_{\mat{Z}} = n$. Then, in $\lambda \rightarrow 0$ limit, it satisfies that 
        \begin{align*}
            \mat{W}_c := \lim\limits_{\lambda \rightarrow 0} \hat{\mat{W}}_2 \hat{\mat{W}}_1 = P_I(\mat{Y}) \mat{Z}^{\dagger}.
        \end{align*}
        \item For some invertible matrix $\mat{C} \in \mathbb{R}^{r_{k} \times r_{k}}$, global minimizers $\hat{\mat{W}}_2^*, \hat{\mat{W}}_1^*$ are given by 
        \begin{align*}
            &\hat{\mat{W}}_2^* = \mat{U}_{\mat{G}, k} \mat{C} \\
            &\hat{\mat{W}}_1^* = \mat{C}^{-1} \mat{U}_{\mat{G}, k}^\top \mat{Y}  \mat{Z}^\top (\tilde{\mat{Z}}\tilde{\mat{Z}}^\top)^{-1}.
        \end{align*}
        Furthermore, assume $r_\mat{Z} = n$. Then the global minimizer $\mat{W}_*$ in the ridgeless case is given uniquely\footnote{Uniqueness in terms of there is an unique $\hat{\mat{W}}_*$.} by,
        $$\mat{W}_* = \lim\limits_{\lambda \rightarrow 0} \hat{\mat{W}}_2^* \hat{\mat{W}}_1^* = P_k(\mat{Y}) \mat{Z}^{\dagger}.$$ 
        \item For $\lambda > 0$, we have an unique global minimizer in terms of $\hat{\mat{W}}_* = \hat{\mat{W}}^*_2 \hat{\mat{W}}^*_1$. 
        On the other hand, at $\lambda = 0$, there are multiple global minimizers. But for the both cases, 
        it satisfies that all the critical points other than global minima are saddle points. In other words, all the local minima are global minima and 
        other critical points are saddle points. 
    \end{enumerate}
\end{theorem}

\textit{Proof.\quad} 
We begin by proving the first result, which characterizes general critical points.
Note that the "$\Leftarrow$" direction of the proof follows from a straightforward calculation, and we therefore omit the details.
For the "$\Rightarrow$" part, observe that for any critical point, the conditions (\ref{appendix:section2:lemma1:consequence1}) and (\ref{appendix:section2:lemma1:consequence2}) must be satisfied, 
and we have that $P_{\mat{W}_2} = \mat{U}_\mat{G} P_{\mat{U}_\mat{G}^\top \mat{W}_2} \mat{U}_\mat{G}^\top$, where $P_{\mat{U}_\mat{G}^\top \mat{W}_2} = \begin{bmatrix}
    \mathcal{D} & 0 \\
    0 & \mat{B}
  \end{bmatrix}.$ From Lemma \ref{appendix:section2:lemma2}, we have that $\mat{B}$ is a symmetric positive semi-definite matrix with eigenvalues $1$ or $0$. Thus its eigendecomposition can be represented as $\mat{B} = \tilde{\mat{U}}_{\mat{B}_1} \tilde{\mat{U}}_{\mat{B}_1}^\top$. 
With $\mat{U}_\mat{G} = \begin{bmatrix} \mat{U}_1 & \mat{U}_2 \end{bmatrix}$, for $\mat{U}_1 \in \mathbb{R}^{d \times r_\mat{G}}$, $\mat{U}_2 \in \mathbb{R}^{d \times (\mat{D} - r_\mat{G})}$. 
With $I_{\mathcal{D}} := \{i ; \mathcal{D}_{ii} = 1\}$, 
we denote $\mat{M}_{\hat{\mat{W}}_2} := \begin{bmatrix}\mat{U}_{\mat{G}, I_{\mathcal{D}}}  & \mat{U}_2 \tilde{\mat{U}}_{\mat{B}_1}\end{bmatrix}$. Then,
$P_{\hat{\mat{W}}_2} = \mat{U}_\mat{G}P_{\mat{U}_\mat{G}^\top \hat{\mat{W}}_2} \mat{U}_\mat{G}^\top = \mat{M}_{\hat{\mat{W}}_2} \mat{M}_{\hat{\mat{W}}_2}^\top$. Thus $\hat{\mat{W}}_2$ is spanned by the columns of $\mat{M}_{\hat{\mat{W}}_2}$. 
Then, there exist an invertible coefficient matrix $\mat{C}$ such that $\hat{\mat{W}}_2 = \mat{M}_{\hat{\mat{W}}_2}\mat{C}$. 
From this, we have that $\hat{\mat{W}}_1 = \mat{C}^{-1} \mat{M}_{\hat{\mat{W}}_2}^\top \mat{Y} \mat{Z}^\top (\tilde{\mat{Z}}\tilde{\mat{Z}}^\top)^{-1}$, from (\ref{appendix:section2:lemma1:consequence1}). For the full rank $\mat{Z}$(i.e., $r_{\mat{Z}} = n$), it satisfies that $\hat{\mat{W}}_1 = \mat{C}^{-1} \begin{bmatrix} \mat{U}_{I_{\mathcal{D}}} & \mat{0} \end{bmatrix}^\top \mat{Y} \mat{Z}^\top (\tilde{\mat{Z}}\tilde{\mat{Z}}^\top)^{-1}$. To see this, consider $\mat{G} = \mat{Y}\mat{Z}^\top (\mat{Z}\mat{Z}^\top + \lambda \mathbb{I})^{-1} \mat{Z}^\top \mat{Y}$. Let $\mat{Z} = \mat{U}_\mat{Z} \mat{\Sigma_\mat{Z}} \mat{V_Z}^\top$ be the SVD of $\mat{Z}$. Then, we have that 
$\mat{G} = \mat{Y} \mat{V}_\mat{Z} \mat{\Sigma_Z}^\top (\mat{\Sigma_Z \Sigma_Z}^\top + \lambda \mathbb{I}_d)^{-1}\mat{\Sigma_Z V_Z^\top Y}^\top$. 
Let $\mat{T := \Sigma_Z^\top (\Sigma_Z \Sigma_Z^\top + \lambda \mathbb{I})^{-1}\Sigma_Z}$. Then $\mat{T}$ is a diagonal matrix with $\mat{T}_{ij} = (\sigma^Z_i)^2((\sigma^Z_i)^2 + \lambda)^{-1} \mathds{1}_{i = j}$ as its diagonal elements. 
Thus, for $\mat{G = Y V_Z T V_Z^\top Y^\top = U_G \Lambda_G U_G^\top}$, $\mat{U_G}$ is a matrix of the left singular vectors of $\mat{Y V_Z \sqrt{T}}$. Then, $\mat{U}_2\mat{Y} = \mat{U}_2 \mat{Y}\mat{V}_\mat{Z} \sqrt{\mat{T}} \sqrt{\mat{T}}^{-1} \mat{V}_\mat{Z}^\top$, and $\mat{U}_2 \mat{Y}\mat{V}_\mat{Z} \sqrt{\mat{T}} = 0$. Therefore, in any case of $|I|$, it holds that $\hat{\mat{W}}_c = \mat{\hat{W}_2 \hat{W}_1} = \mat{U}_{\mat{G}, I}\mat{U}_{\mat{G}, I}^\top \mat{YZ^\top (\tilde{Z}\tilde{Z}^\top)}^{-1}$. 
Finally, in the ridgeless limit $\lambda \rightarrow 0$, it holds that $\mat{G} = \mat{Y} \mat{Y}^\top$. Therefore, it satisfies that:
\begin{align*}
    \mat{W}_c = \lim\limits_{\lambda \rightarrow 0} \mat{\hat{W}}_c
    &= \lim\limits_{\lambda \rightarrow 0} \mat{U}_{\mat{G}, I} \mat{U}_{\mat{G}, I}^\top \mat{YZ^\top (\tilde{Z}\tilde{Z}^\top)}^{-1} \\
    &= \mat{U}_{\mat{Y}, I} \mat{U}_{\mat{Y}, I}^\top \mat{Y Z}^{\dagger} = P_I(\mat{Y}) \mat{Z}^\dagger.
\end{align*}

We now prove the second claim about the global minimizer. To do this, we analyze the loss function (\ref{section2:general_loss}) further.
The loss function with $\lambda \rightarrow 0$ is given by
\begin{align*}
    &n \mathcal{L}(\mat{W}_2, \mat{W}_1) = \|\mat{Y} - \mat{W}_2 \mat{W}_1\mat{Z}\|_F^2 \\
    &= \Tr(\mat{Y}\mat{Y}^\top) - 2\Tr(\mat{W}_2\mat{W}_1 \mat{Z} \mat{Y}^\top) + \Tr(\mat{W}_2\mat{W}_1 \mat{Z}\mat{Z}^\top\mat{W}_1^\top\mat{W}_2^\top) \\
    &=\Tr(\mat{Y}\mat{Y}^\top) - 2\Tr(P_{\mat{W}_2} \mat{G}) + \Tr(P_{\mat{W}_2} \mat{G} P_{\mat{W}_2}) \\
    &=\Tr(\mat{Y}\mat{Y}^\top) - \Tr(P_{\mat{U}_\mat{G}^\top\mat{W}_2} \mat{G}) \\
    &=\Tr(\mat{Y}\mat{Y}^\top) - \Tr(\mathcal{D} \Lambda_{\mat{G}}).
\end{align*}
This result implies that the loss function depends solely on $\Tr(\mathcal{D} \mat{\Lambda}_{\mat{G}})$. Consequently, minimizing the loss is equivalent to maximizing $\Tr(\mathcal{D} \mathbf{\Lambda})$.
Then, for the optimal $\mathcal{D}^*$, $r_{\mathcal{D}^*}$ must be $k$, which is its upper bound. 
Furthermore, $\mathcal{D}^*_{ii} = \mathds{1}_{i \leq k}$, as the first $k$ components of $\mat{\Lambda}_\mat{G}$ will give the biggest trace. 
This is our global minimum. Thus, our optimal index set $I^* = [k]$.
Note that in this case, where all the budget $k$ is spent on $\mathcal{D}^*$, $\mat{B}$ becomes $0$, which follows from the fact that all the eigenvalues of $\mat{B}$ should be $0$. As a result, $P_{\mat{U}_\mat{G}^\top \mat{W}^*_2}$ becomes a diagonal matrix, 
with its $i$-th diagonal element $(P_{\mat{U}_\mat{G}^\top \mat{W}^*_2})_{ii} = \mathds{1}_{i \leq k}$. 
Thus, 
\begin{align}
    P_{\mat{W}^*_2} &= \mat{U}_\mat{G} P_{\mat{U}_\mat{G}^\top \mat{W}_2} \mat{U}_\mat{G}^\top = \mat{U}_{\mat{G}, k} \mat{U}_{\mat{G}, k}^\top,
\end{align}
where $\mat{U}_{\mat{G}, k}$ consists of the first $k$ columns of $\mat{U}_\mat{G}$, 
which correspond to the top-$k$ eigenvalues of $\mat{G}$.
This leads us to the conclusion that for the full rank $\mat{Z}$, the global minimum of our ridgeless estimator is indeed
\begin{align*} 
    \mat{W}_* &= \lim\limits_{\lambda \rightarrow 0} P_{\mat{W}_2} \mat{Y} \mat{Z}^\top (\tilde{\mat{Z}} \tilde{\mat{Z}}^\top)^{-1} \\
    &= \lim\limits_{\lambda \rightarrow 0} \mat{U}_{\mat{G}, k} \mat{U}_{\mat{G}, k}^\top \mat{Y} \mat{Z}^\top (\tilde{\mat{Z}} \tilde{\mat{Z}}^\top)^{-1} \xrightarrow{\lambda \rightarrow 0} P_k(\mat{Y}) \mat{Z}^{\dagger}.
\end{align*}

The last part is to prove that these critical points are saddle points with descent direction. 
If $\lambda > 0$, then the global minimum is uniquely defined with $\mat{\hat{W}^*}$. All the critical points other than 
the global minimum are saddle points, and it turns out that we can prove this by following the proof of \cite{baldi1989neural}. 
We give a brief summary of this proof. The main idea of this proof is basically perturbing a column vector of $\mat{U}_{\mat{G}, I_{\mathcal{D}}}$ a bit into the direction of eigenvector $\vec{u}^\mat{G}_j$ of $\mat{U}_{\mat{G}, k}$, 
for $j \notin I_{\mathcal{D}}$ and $j \in [k]$. Because this $\vec{u}^{\mat{G}}_j$ corresponds to some bigger eigenvalue, we can construct a new perturbed $\tilde{\mathcal{D}}$ matrix 
which contains an entry in its diagonal part that would make $\Tr(\tilde{\mathcal{D}} \mat{\Lambda}) > \Tr(\mathcal{D} \mat{\Lambda})$. Thus this results in 
decreased loss function, which basically shows that this is indeed a saddle point with decreasing direction. For example in \cite{baldi1989neural}, 
they have constructed this direction, by picking a column vector $\vec{u}^\mat{G}_p$ from $\mat{U}_{\mat{G}, I_{\mathcal{D}}}$, that $p \notin [k]$. 
Then, they have replaced this $\vec{u}^\mat{G}_p$ with $\tilde{\vec{u}}^\mat{G}_p = (1 + \epsilon^2)^{-\frac{1}{2}}(\vec{u}^\mat{G}_p + \epsilon \vec{u}^\mat{G}_j)$, 
for any $\epsilon > 0$, where $\vec{u}^\mat{G}_j$ is the eigenvector of $\mat{U}_k$ that corresponds to the $j$-th eigenvalue. 
Further details of can be found in \cite{baldi1989neural}. 
For $\lambda = 0$, we have multiple global minima, but all the critical points other than global minima are saddle points. 
This fact has been already proven in \cite[Thm. 1]{zhou2017criticalpointsneuralnetworks} and we refer to this work.

\rightline{$\square$}

\subsection{Equivalence to the Reduced-Rank Regression}\label{appendix:section2:equivalence_rrr}
We begin by examining the solution to a one-layer linear denoising autoencoder (DAE) with a rank constraint and no regularization. We then consider the corresponding minimum-norm solution.
This form of linear regression with a rank constraint is commonly known in the literature as \textit{reduced-rank regression}~\cite{mukherjee2011reduced,daviesTso}.
\paragraph*{Training Objective}
The training objective is defined as follows:
\begin{equation} \label{chap:rcla:one_layer:training_metric1} 
    \mat{W}_* := \argmin\limits_{\mat{W} \in \mathbb{R}^{d \times d}, \newline r_\mat{W} \leq k} \frac{1}{n}\|\mat{X} - \mat{W}(\mat{X} + \mat{A})\|_F^2.
\end{equation}
Since the loss function involves a non-smooth rank constraint, we derive the solution by following the method outlined in~\cite{xiang2012optimal}. Note that the result is applicable for the overparameterized setting.
\begin{theorem}\textnormal{(Solution of Rank-Constrained One-Layer Linear DAE).}
    \label{chap:rcla:one_layer:solution}
    Consider the one-layer linear denoising autoencoder with the training objective defined in (\ref{chap:rcla:one_layer:training_metric1}). 
    Let $\mat{X} + \mat{A} = \bar{\mat{U}} \bar{\mat{\Sigma}} \bar{\mat{V}}^T$ denote the singular value decomposition of $\mat{X + A}$.
    Then, for any $\mat{C} \in \mathbb{R}^{n \times (d - n)}$, the global minimizer $\mat{W}_*$ is given by \begin{equation}\label{chap:rcla:one_layer:solution_equation}
        \mat{W}_* = \begin{bmatrix} P_k(\mat{X})\bar{\mat{V}} \bar{\mat{D}}^{-1} & P_k(\mat{X})\bar{\mat{V}} \mat{C} \end{bmatrix} \bar{\mat{U}}^T.
    \end{equation}
\end{theorem}
\begin{remark}\textnormal{(Minimum-Norm Solution).}
    Note that this holds for any $\mat{C}$, and the minimum-norm solution corresponds to the case $\mat{C} = \mat{0}$. In this case, $\mat{W}_* = P_k(\mat{X})(\mat{X + A})^\dagger$, , which matches the minimum-norm solution derived in Theorem~\ref{appendix:section2:full_theorem}.
\end{remark}

\textit{Proof.\quad}
Let $\mat{Z} := \mat{X}\bar{\mat{V}}$ and $\mat{Y} := \mat{W}\bar{\mat{U}}$. Using the invariance of the Frobenius norm under the unitary transformations, we can write:
\begin{align}
    \|\mat{X} - \mat{W}(\mat{X} + \mat{A})\|_F^2 &= \|(\mat{X} - \mat{W}\bar{\mat{U}} \bar{\mat{\Sigma}} \bar{\mat{V}}^T )\bar{\mat{V}} \|_F^2 \nonumber\\
    &= \|\mat{X}\bar{\mat{V}} - \mat{W}\bar{\mat{U}}\bar{\mat{\Sigma}} \|_F^2 \nonumber\\
    &= \|\mat{Y} \bar{\mat{\Sigma}} - \mat{Z}\|_F^2 .
\end{align}
With $\mat{Y} \bar{\mat{\Sigma}} = \mat{Z}$, we have that
$\mat{Y} \bar{\mat{\Sigma}} = \begin{bmatrix} \mat{Y}_n & \mat{Y}_{d - n} \end{bmatrix}$
$
\begin{bmatrix}
  \bar{\mat{D}}_n \\
  \mat{0} \\
\end{bmatrix} = \mat{Y}_n\bar{\mat{D}}_n = \mat{Z}. $ 
This equivalence tells us that finding $\mat{Y}_n$ is equivalent to finding $\mat{W}$ that satisfies the rank constraint.
Thus, the problem of finding $W$ can be reduced to solving the following optimization problem:
\begin{align}
    \label{chap:rcla:one_layer:training_metric2}
\argmin_{\mat{Y}_n \in \mathbb{R}^{d \times n}} \|\mat{Y}_n\bar{\mat{D}}_n - \mat{Z}\|_F^2, \quad \text{s.t.} \quad rank(\mat{Y}_n\bar{\mat{D}}_n) \leq k.
\end{align}
Now applying the Eckart-Young Theorem\cite{eckart1936approximation} gives us that $\mat{Y}_{n}^* \bar{\mat{D}}_n := P_k(\mat{Z})$. 
Therefore, $\mat{Y}_{n}^* = P_k(\mat{Z}) \bar{\mat{D}}_n^{-1}$. \\ 
For $\mat{Y}_{d - n}^*$, there are no additional constraints other than the fact that this term cannot introduce additional rank, 
since all the rank $k$ were spent for $\mat{Y}_n$. In other words, columns of this matrix should be linear combinations of columns of 
the $\mat{Y}_n$. Thus there exists a coefficient matrix $\mat{C} \in \mathbb{R}^{n \times (d - n)}$, such that $\mat{Y}_{d - n}^* = \mat{Y}_{n}^* \mat{C}$. 
Substituting these results, we have that
\begin{align}
    \mat{W}_*  &= \mat{Y}^* \mat{U}^T \nonumber \\
    &= \begin{bmatrix} P_k(\mat{X}\bar{\mat{V}})\bar{\mat{D}}_n^{-1} & P_k(\mat{X}\bar{\mat{V}}) \mat{C} \end{bmatrix} \bar{\mat{U}}^T \nonumber \\
    &= \begin{bmatrix} P_k(\mat{X})\bar{\mat{V}}\bar{\mat{D}}_n^{-1} & P_k(\mat{X})\bar{\mat{V}} \mat{C} \end{bmatrix} \bar{\mat{U}}^T. \nonumber
\end{align}

\rightline{$\square$}

\section{Proofs and Bias–Variance Definition for Section~\ref{section:generalization}}
In this section, we first provide detailed proofs of the theorems stated in Section~\ref{section:generalization}.
Using these results, we then define a notion of bias and variance that applies to the two models studied in this paper, as presented in Appendix~\ref{appendix:section3:bias_variance_decomposition}.

\subsection{Proofs}\label{appendix:section3:proofs}
The main idea of the proof, originally introduced in~\cite{sonthalia2022training} and extended in~\cite{kausik2024doubledescentoverfittingnoisy}, is to first handle the pseudo-inverse term and then apply a concentration argument.  
To deal with the pseudo-inverse,~\cite{kausik2024doubledescentoverfittingnoisy} applies results from~\cite{WEI20011}, which allow expansion of the pseudo-inverse under certain rank conditions.

For the concentration step, assume $X$ and $Y$ are real-valued random variables that are concentrated around their means, with variances scaling as $o(1)$. In this case, we have the bound $|\mathbb{E}[XY] - \mathbb{E}[X] \mathbb{E}[Y]| \leq \sqrt{Var(X)Var(X)}$, which implies that $\mathbb{E}[XY] = \mathbb{E}[X] \mathbb{E}[Y] + o(1)$. A similar approximation holds for products involving more than two random variables, using results of~\cite{Bohrnstedt_Arthur} regarding the covariance of four random variables. Importantly, this concentration argument allows us to approximate expectations of products by products of expectations. For example,
$\mathbb{E}\left[\Tr(XY)\right] = \Tr\left(\mathbb{E}[X]\mathbb{E}[Y]\right) + o(1)$, which plays a critical role in the proof.
For further details, we refer the reader to~\cite{kausik2024doubledescentoverfittingnoisy}. 

We first prove Theorem \ref{theorem:test_risk_noskip}, followed by Theorem \ref{theorem:test_risk_skip}. Before presenting the proofs, we introduce the following notation, adapted from~\cite{kausik2024doubledescentoverfittingnoisy}. Let $\mat{\tilde{\mat{U}} D \tilde{\mat{V}}^\top}$ be the reduced SVD of $\mat{X}$. Then, we denote 
\begin{itemize}
    \item $\mat{P} := - (\mathbb{I} - \mat{A\mat{A}^{\dagger})\tilde{\mat{U}}D} \in \mathbb{R}^{d \times r}$
    \item $\mat{H} := \tilde{\mat{V}}^\top\mat{A}^{\dagger} \in \mathbb{R}^{r \times d}$
    \item $\mat{Z} := \mathbb{I} + \tilde{\mat{V}}^\top\mat{A}^{\dagger}\tilde{\mat{U}}\mat{D} \in \mathbb{R}^{r \times r}$
    \item $\mat{K}_1 := \mat{HH}^\top + \mat{Z}(\mat{P}^\top\mat{P})^{-1}\mat{Z}^\top \in \mathbb{R}^{r \times r}$.
\end{itemize}

\subsubsection{Proof of Theorem \ref{theorem:test_risk_noskip}} \label{appendix:section3:proof_thm1}
Observe the following decomposition of the test metric (Eq.~(\ref{test_metric})):
\begin{align}\label{appendix:decomposition_woskip}
     &\frac{1}{N_{\text{\text{tst}}}} \mathbb{E}_\mat{{A_{\text{\text{trn}}}, A_{\text{\text{tst}}}}}\left[\|\mat{X}_{\text{\text{tst}}} - \mat{W}_c (\mat{X}_{\text{\text{tst}}} + \mat{A}_{\text{\text{tst}}})\|_F^2 \right] \nonumber \\
     &= \frac{1}{N_{\text{\text{tst}}}}\mathbb{E}_{\mat{A}_{\text{\text{trn}}}, \mat{A}_{\text{\text{tst}}}}\left[\Tr((\mat{X}_{\text{\text{tst}}} - \mat{W}_c (\mat{X}_{\text{\text{tst}}} + \mat{A}_{\text{\text{tst}}}))(\mat{X}_{\text{\text{tst}}} - \mat{W}_c (\mat{X}_{\text{\text{tst}}} + \mat{A}_{\text{\text{tst}}})^\top))\right] \nonumber \\
     & = \frac{1}{N_{\text{\text{tst}}}}\mathbb{E}_{\mat{A}_{\text{\text{trn}}}, \mat{A}_{\text{\text{tst}}}}\left[\|\mat{X}_{\text{\text{tst}}} - \mat{W}_c\mat{X}_{\text{\text{tst}}}\|_F^2 + \|\mat{W}_c\mat{A}_{\text{\text{tst}}}\|_F^2\right] \nonumber \\
     &= \frac{1}{N_{\text{\text{tst}}}}\mathbb{E}_{\mat{A}_{\text{\text{trn}}}}\left[\|\mat{X}_{\text{\text{tst}}} - \mat{W}_c\mat{X}_{\text{\text{tst}}}\|_F^2\right] + \frac{\eta_{\text{\text{tst}}}^2}{d} \mathbb{E}_{\mat{A}_{\text{\text{trn}}}}\left[\|\mat{W}_c\|_F^2\right].
\end{align}
Based on this decomposition, our goal is to prove Lemmas~\ref{appendix:section3:wo_skip:lma3} (the first term) and~\ref{appendix:section3:wo_skip:lma4}(the second term), which together establish Theorem~\ref{theorem:solution_noskip}.

We begin by presenting the technical lemmas required for the proof. 
Recall that critical points satisfy $\mat{W}_c = P_{I^x}(\mat{X})(\mat{X + A})^{\dagger}$. 
By a slight abuse of notation, we denote $\mat{W}_c = \mat{X}_{I^x}(\mat{X + A})^{\dagger}$. 

\begin{lemma}\label{appendix:section3:wo_skip:lma1}\textnormal{(Expanding the Pseudo-Inverse Term for Critical Points).}
    In the case of $d \geq n + r$ and given the solution $\mat{W}_c = \mat{X}_{I^x}(\mat{X + A})^{\dagger}$, it holds that
    \begin{equation}
    \mat{W}_c = \tilde{\mat{U}}\mat{D}_{I^x}(\mat{P}^\top\mat{P})^{-1}\mat{Z}^\top\mat{K}_1^{-1}\mat{H} - \tilde{\mat{U}}\mat{D}_{I^x}\mat{Z}^{-1}\mat{HH}^\top\mat{K}_1^{-1}\mat{Z}\mat{P}^{\dagger}
    \end{equation}
\end{lemma}
\textit{Proof.\quad}
    As $r \leq d - n$, the above $\mat{P}$ matrix has full rank. Thus we can invoke Corollary 2.1 from \cite{WEI20011}, then we have
    \begin{align*}
    (\mat{A + \tilde{\mat{U}} D \tilde{\mat{V}}}^\top)^{\dagger} = \mat{A}^{\dagger} + \mat{A}^{\dagger}\mat{\tilde{\mat{U}} D P}^{\dagger} 
    &- (\mat{A}^{\dagger}\mat{H}^\top + \mat{A}^{\dagger}\tilde{\mat{U}} \mat{D} (\mat{P}^\top\mat{P})^{-1}\mat{Z}^\top)\mat{K}_1^{-1}(\mat{H} + \mat{Z}\mat{P}^{\dagger}).
    \end{align*}
    Then, multiplying $\mat{X}_{I^x} = \tilde{\mat{U}} \mat{D}_{I^x} \tilde{\mat{V}}^\top$ to the left side, we get
    \begin{align*}
        &\mat{X}_{I^x}(\mat{X} + \mat{A})^{\dagger} \\
        &= \mat{\tilde{\mat{U}}D}_{I^x}\tilde{\mat{V}}^\top(\mat{A}^{\dagger} + \mat{A}^{\dagger}\mat{U}\mat{D} \mat{P}^{\dagger} - (\mat{A}^{\dagger}\mat{H}^\top + \mat{A}^{\dagger}\tilde{\mat{U}}\mat{D} (\mat{P}^\top\mat{P})^{-1}\mat{Z}^\top)\mat{K}_1^{-1}(\mat{H} + \mat{Z}\mat{P}^{\dagger})) \\
        &= \tilde{\mat{U}}\mat{D}_{I^x}\tilde{\mat{V}}^\top \mat{A}^{\dagger} + \tilde{\mat{U}}\mat{D}_{I^x} \tilde{\mat{V}}^\top\mat{A}^{\dagger}\mat{U}\mat{D}\mat{P}^{\dagger} - \tilde{\mat{U}}\mat{D}_{I^x}\tilde{\mat{V}}^\top \mat{A}^{\dagger}\mat{H}^\top \mat{K}_1^{-1}(\mat{H} + \mat{Z}\mat{P}^{\dagger}) -\\
        & \tilde{\mat{U}}\mat{D}_{I^x}\tilde{\mat{V}}^\top\mat{A}^{\dagger}\tilde{\mat{U}}\mat{D} (\mat{P}^\top\mat{P})^{-1}\mat{Z}^\top \mat{K}_1^{-1}(\mat{H} + \mat{Z}\mat{P}^{\dagger}).
\end{align*}
The desired result can be obtained by successively substituting 
$\tilde{\mat{V}}^\top \mat{A} = \mat{H}$, 
$\mat{H}\tilde{\mat{U}}\mat{D} = \mat{Z} - \mathbb{I}$, and $\mat{H}\mat{H}^\top + \mat{Z}(\mat{P}^\top \mat{P})^{-1} \mat{Z}^\top = \mat{K}_1$.

\rightline{$\square$}

\begin{lemma}\label{appendix:section3:wo_skip:lma2}
    Let $\mat{T} \in \mathbb{R}^{r \times r}$ be a diagonal matrix which satisfies 
    \begin{align*}
        \mat{T}_{ii} = 
        \begin{cases}
            1 & \text{if } i \in I^x \\
            0 & \text{otherwise.}
        \end{cases}
    \end{align*}
     Then,
    $\mathbb{E}[\mat{D}_{I^x}(\mat{P}^\top\mat{P})^{-1}\mat{D}] = \frac{c}{c - 1}\mat{T} + O(d^{-1}).$
\end{lemma}
\textit{Proof.\quad} Observe that
    \begin{align*}
        \mathbb{E}[\mat{D}_{I^x}(\mat{P}^\top\mat{P})^{-1}\mat{\Sigma}]
        &= \mathbb{E}[\mat{D}_{I^x}\mat{D}^{-1}(\mat{D}(\mat{P}^\top\mat{P})^{-1}\mat{D})] \\
        &= \mat{D}_{I^x}\mat{D}^{-1}\mathbb{E}[\mat{D}(\mat{P}^\top\mat{P})^{-1}\mat{D}] \\
        &= \mat{T} \mathbb{E}[\mat{D}\mat{P}^\top\mat{P})^{-1}\mat{D}].
    \end{align*}
    By applying Lemma 6 of \cite{kausik2024doubledescentoverfittingnoisy}, 
    we have 
    \begin{align*}
        \mathbb{E}[\mat{D}_{I^x}(\mat{P}^\top\mat{P})^{-1}\mat{\Sigma}] = \frac{c}{c - 1}\mat{T} + O(d^{-1}).
    \end{align*}
Because $\mat{T}$ is a diagonal matrix with 0 or 1 entries on its diagonal part, the element-wise variance is still $O(d^{-1})$.
    
\rightline{$\square$}

Now we state lemmas that are directly related to the test metric. 

\begin{lemma}\label{appendix:section3:wo_skip:lma3}\textnormal{(Variance Term).}
    For $d \geq n + r$, 
    \begin{align*}
        \mathbb{E}[\|\mat{W}_c\|_F^2] &= \frac{c}{c - 1} \sum\limits_{j \in I^x} \frac{\sigma_j^2}{\eta_{\text{\text{trn}}}^2 + \sigma_j^2} + O\left(\frac{\|\mat{D}\|^2}{d}\right) + o(1).
    \end{align*}
\end{lemma}

\textit{Proof. \quad} 
    Note that $\|\mat{W}_c\|_F^2 = \Tr(\mat{W}_c^\top \mat{W}_c)$. Use Lemma \ref{appendix:section3:wo_skip:lma1} to expand this.
    \begin{align}\label{chap:tlla:proof:wo_skip:eq1}
    \mathbb{E}[\Tr(\mat{W}_c^{\top} \mat{W}_c)] = \mathbb{E}[\Tr(\mat{H}^\top \mat{K}_1^{-1}\mat{T}\mat{Z}(\mat{P}^\top \mat{P})^{-1}\mat{D}_{I^x}^2(\mat{P}^\top \mat{P})^{-1}\mat{Z}^\top\mat{K}_1^{-1}\mat{H}) \nonumber \\ 
    - 2\Tr(\mat{K}_1^{-1}\mat{Z}^\top\mat{D}^{-1}(\mat{D}(\mat{P}^\top \mat{P})^{-1}\mat{D}_{I^x}^\top \mat{Z}\mat{Z}^{-1}\mat{HH}^\top \mat{Z}\mat{P}^{\dagger}\mat{H}^\top)) \nonumber \\ 
    + \Tr(\mat{P}^{\dagger})^\top \mat{Z}^\top \mat{K}_1^{-1} \mat{H}\mat{H}^\top(\mat{Z}^{-1})^\top \mat{D}_{I^x}^\top  \tilde{\mat{U}}^\top \tilde{\mat{U}} \mat{D}_{I^x}\mat{Z}^{-1}\mat{HH}^\top \mat{K}_1^{-1}\mat{Z}\mat{P}^{\dagger})].
    \end{align}
    Using the cyclic invariance property of trace, the first term is equivalent to 
    $$
    \mathbb{E}[\Tr(\mat{D}_{I^x}(\mat{P}^\top\mat{P})^{-1}Z^\top\mat{K}_1^{-1}\mat{HH}^\top\mat{K}_1^{-1}\mat{Z}(\mat{P}^\top\mat{P})^{-1}\mat{D}_{I^x})]
    $$.
    This is equivalent to
    $$
    \mathbb{E}[\Tr(\mat{D}_{I^x}(\mat{P}^\top \mat{P})^{-1}\mat{D}(\mat{D}^{-1}\mat{Z}^\top)\mat{K}_1^{-1}\mat{H}\mat{H}^\top \mat{K}_1^{-1}(\mat{Z}\mat{D}^{-1})\mat{D}(\mat{P}^\top \mat{P})^{-1}\mat{D}_{I^x})] 
    $$
    We invoke Lemma \ref{appendix:section3:wo_skip:lma2}, and 4, 7, 8 from \cite{kausik2024doubledescentoverfittingnoisy}
    Each of the terms $\mat{D}_{I^x}(\mat{P}^\top \mat{P})^{-1}\mat{D}$, $\mat{D}^{-1}\mat{Z}^\top$, $\mat{K}_1^{-1}, \mat{H}\mat{H}^\top$ has element-wise variance of $O(d^{-1})$ which vanishes away. Thus, we have concentration around the product of expectations (more details in \cite{kausik2024doubledescentoverfittingnoisy}). 
    Applying the lemmas to each of these terms gives us that
    \begin{align*}
    & \mathbb{E}[\Tr(\mat{D}_{I^x}(\mat{P}^\top \mat{P})^{-1}\mat{D}(\mat{D}^{-1}\mat{Z}^\top)\mat{K}_1^{-1}\mat{H}\mat{H}^\top \mat{K}_1^{-1}(\mat{Z}\mat{D}^{-1})\mat{D}(\mat{P}^\top \mat{P})^{-1}\mat{D}_{I^x})] \\
    &=\frac{\eta_{\text{\text{trn}}}^2 c}{(c - 1)}\Tr(\mat{T}\mat{D}^{-1}(\eta_{\text{\text{trn}}}^2 \mat{D}^{-2} + \mathbb{I}_r)^{-2}\mat{D}^{-1}\mat{T}^\top) + o(1).
    \end{align*}
     This is equivalent to
    \begin{align*}
    &\frac{\eta_{\text{\text{trn}}}^2c}{(c - 1)}\Tr(\mat{D}_{I^x}^{-1}(\eta_{\text{\text{trn}}}^2\mat{D}_{I^x}^{-2} + \mathbb{I}_{I^x})^{-2}\mat{D}_{I^x}^{-1}) + o(1) \\
    &= \frac{\eta_{\text{\text{trn}}}^2c}{(c - 1)}\Tr(\mat{D}_{I^x}^{2}(\eta_{\text{\text{trn}}}^2 \mathbb{I}_{I^x} + \mat{D}_{I^x}^{2})^{-2}) + o(1).
    \end{align*} Recall that the $o(1)$ term accounts for the error introduced when replacing the expectation of a product with the product of expectations. 
    
    The second term of \eqref{chap:tlla:proof:wo_skip:eq1} directly follows from the argument of the fact that $\mat{P}^{\dagger}\mat{H}^\top = 0$. 
    Thus this term is $0$.
    For the third term of \eqref{chap:tlla:proof:wo_skip:eq1}, , we proceed as follows: \begin{align*}
    &\mathbb{E}[\Tr((\mat{P}^{\dagger})^\top \mat{Z}^\top(\mat{K}_1^{-1})^\top \mat{H}\mat{H}^\top(\mat{Z}^{-1})^\top \mat{D}_{I^x}^\top \mat{D}_{I^x}\mat{Z}^{-1}\mat{H}\mat{H}^\top \mat{K}_1^{-1}\mat{Z}\mat{P}^{\dagger})] \nonumber \\
    &= \mathbb{E}[\Tr((\mat{K}_1^{-1})^\top \mat{H}\mat{H}^\top(\mat{D}_{I^x} \mat{Z}^{-1})^\top \mat{D}_{I^x}\mat{Z}^{-1}\mat{H}\mat{H}^\top \mat{K}_1^{-1}\mat{Z}\mat{D}^{1}(\mat{D}(\mat{P}^\top \mat{P})^{-1}\mat{D})\mat{D}^{-1}\mat{Z}^\top)]
    \end{align*}
    Applying Lemma 4, 6, 8 of \cite{kausik2024doubledescentoverfittingnoisy}, and Lemma \ref{appendix:section3:wo_skip:lma2}, 
    it follows that
    \begin{align*}
    \frac{c}{(c - 1)}\mathbb{E}[\Tr((\eta_{\text{\text{trn}}}^2 \mat{D}^{-2} + \mathbb{I}_r)^{-1} \mat{T} (\mat{D} \mat{Z}^{-1})^\top \mat{T} (\mat{D}\mat{Z}^{-1})(\eta_{\text{\text{trn}}}^2 \mat{D}^{-2} + & \mathbb{I}_r)^{-1}\mat{Z}\mat{D}^{-1}\mat{D}^{-1}\mat{Z}^\top)] \\ 
    & + o(1)
    \end{align*}
    for $\mat{T}$ defined in Lemma \ref{appendix:section3:wo_skip:lma2}.
    Applying Lemma 7 of \cite{kausik2024doubledescentoverfittingnoisy} for $\mat{Z}^{-1}$ and $\mat{Z}$, we finally get
    \begin{align*}
        &\frac{c}{c - 1}\Tr(\mat{D}_{I^x}^2(\eta_{\text{\text{trn}}}^2 \mat{D}^{-2} + \mathbb{I}_r)^{-2}\mat{D}^{-2}) + O\left(\frac{\|\mat{D}\|^2}{d}\right) + o(1) \\
    &= \frac{c}{c - 1}\Tr(\mat{D}_{I^x}^4(\eta_{\text{\text{trn}}}^2 \mathbb{I}_{I^x} + \mat{D}_{I^x}^2)^{-2})) + O\left(\frac{\|\mat{D}\|^2}{d}\right) + o(1).
    \end{align*}
    Finally, combining all three terms gives us that
    \begin{align*}
    \mathbb{E}[\|\mat{W}_c\|_F^2] &= \frac{c}{c - 1} \Tr(\mat{D}_{I^x}^2( \eta_{\text{\text{trn}}}^2 \mathbb{I}_{I^x} + \mat{D}_{I^x}^2)^{-1}) + O\left(\frac{\|\mat{D}\|^2}{d}\right) + o(1) \\
    &= \frac{c}{c - 1} \sum\limits_{j \in I^x} \frac{\sigma_j^2}{\eta_{\text{\text{trn}}}^2 + \sigma_j^2} + O\left(\frac{\|\mat{D}\|^2}{d}\right) + o(1).
    \end{align*}
    
\rightline{$\square$}

\begin{lemma}\label{appendix:section3:wo_skip:lma4}\textnormal{(Bias Term).}
    For $d \geq n + r$, and given $\mat{X}_{\text{tst}} = \tilde{\mat{U}} \mat{L}$ for some $\mat{L} \in \mathbb{R}^{r \times N_{\text{tst}}}$, we have
    \begin{align*}
        &\frac{1}{N_{\text{tst}}}\mathbb{E}[\|\mat{X}_{\text{tst}} - \mat{W}_c \mat{X}_{\text{tst}}\|_F^{2}] \\
        &= \frac{1}{N_{\text{tst}}}\Tr(\mat{J} \mat{L}\mat{L}^{\top}) + o(1).
    \end{align*}
    $\mat{J} \in \mathbb{R}^{d \times d}$ is a diagonal matrix defined as 
    \begin{align*}
        \mat{J}_{ii} = \begin{cases}
            \left(\left(\frac{\sigma_i}{\eta_{\text{trn}}}\right)^2 + 1\right)^{-2} & \text{if } i \in I^x \\
            1 & \text{otherwise.}
        \end{cases}
    \end{align*}
\end{lemma}

\textit{Proof.\quad} 
    Using Lemma \ref{appendix:section3:wo_skip:lma1} to replace $\mat{W}_c$, we obtain that 
    \begin{align*}
        \mat{X}_{\text{tst}} - \mat{W}_c \mat{X}_{\text{tst}} &= \tilde{\mat{U}} \mat{L} - (\tilde{\mat{U}}\mat{D}_{I^x} (\mat{P}^{\top} \mat{P})^{-1} \mat{Z}^{\top} \mat{K}_1^{-1} \mat{H} - \tilde{\mat{U}} \mat{D}_{I^x} \mat{Z}^{-1} \mat{H}\mat{H}^{\top} \mat{K}_1^{-1} \mat{Z} \mat{P}^{\dagger})\tilde{\mat{U}} \mat{L}.
    \end{align*}
    With $\mat{P}^{\dagger} \tilde{\mat{U}} = - \mat{D}^{-1}$ and $\mat{H}\tilde{\mat{U}} = (\mat{Z} - \mathbb{I})\mat{D}^{-1}$, we have that
    \begin{align*}
        &\tilde{\mat{U}}\mat{L} - \tilde{\mat{U}} \mat{D}_{I^x} (\mat{P}^{\top} \mat{P})^{-1} \mat{Z}^{\top} \mat{K}_1^{-1} (\mat{Z} - \mathbb{I})\mat{D}^{-1} \mat{L} - \tilde{\mat{U}}\mat{D}_{I^x} \mat{Z}^{-1} \mat{H} \mat{H}^{\top} \mat{K}_1^{-1} \mat{Z} \mat{D}^{-1} \mat{L} \\
        &= \tilde{\mat{U}}\mat{L} - \tilde{\mat{U}} \mat{D}_{I^x} \mat{Z}^{-1} (\mat{Z} (\mat{P}^{\top} \mat{P})^{-1} \mat{Z}^{\top} \mat{K}_1^{-1} (\mat{Z} - \mathbb{I}) + \mat{H}\mat{H}^{\top} \mat{K}_1^{-1} \mat{Z}) \mat{D}^{-1} \mat{L} \\
        &= \tilde{\mat{U}}\mat{L} - \tilde{\mat{U}} \mat{D}_{I^x} \mat{Z}^{-1} ((\mat{K}_1 - \mat{H}\mat{H}^{\top}) \mat{K}_1^{-1} (\mat{Z} - \mathbb{I}) + \mat{H}\mat{H}^{\top} \mat{K}_1^{-1} \mat{Z}) \mat{D}^{-1} \mat{L} \\
        &= \tilde{\mat{U}}\mat{L} - \tilde{\mat{U}} \mat{D}_{I^x} \mat{Z}^{-1} (\mat{Z} - \mathbb{I} + \mat{H}\mat{H}^{\top}\mat{K}_1^{-1}) \mat{D}^{-1} \mat{L}. 
    \end{align*}
    The third inequality is due to $\mat{Z} (\mat{P}^{\top} \mat{P})^{-1} \mat{Z}^{\top} = \mat{K}_1 - \mat{H}\mat{H}^{\top}$. Expanding the parentheses, for $\mat{T}^c := \mathbb{I} - \mat{T}$ 
    ($\mat{T}$ from Lemma \ref{appendix:section3:wo_skip:lma2}), we obtain that
    \begin{align*}
        &\tilde{\mat{U}}\mat{L} - \tilde{\mat{U}}\mat{D}_{I^x} \mat{D}^{-1} \mat{L} + \tilde{\mat{U}} \mat{D}_{I^x} (\mat{P}^{\top} \mat{P})^{-1} \mat{Z}^{\top} \mat{K}_1^{-1} \mat{D}^{-1} \mat{L} \\
        &= \tilde{\mat{U}} \mat{T}^c \mat{L} + \tilde{\mat{U}} \mat{D}_{I^x} (\mat{P}^{\top} \mat{P})^{-1} \mat{Z}^{\top} \mat{K}_1^{-1} \mat{D}^{-1} \mat{L}.
    \end{align*}
    We consider the term $\|\mat{X}_{\text{tst}} - \mat{W}_c \mat{X}_{\text{tst}}\|_F^2$. 
   Using the equivalence between the Frobenius norm and the trace, this can be written as $\Tr((\mat{X}_{\text{tst}} - \mat{W}_c \mat{X}_{\text{tst}})^{\top} (\mat{X}_{\text{tst}} - \mat{W}_c \mat{X}_{\text{tst}}))$. Expanding the expression, and applying the cyclic invariance of the trace along with the identity $\mat{T}^c \mat{D}_{I^x} = 0$, 
    we obtain that, 
    \begin{align*}
        &\mathbb{E}[\Tr((\mat{X}_{\text{tst}} - \mat{W}_c \mat{X}_{\text{tst}})^{\top} (\mat{X}_{\text{tst}} - \mat{W}_c \mat{X}_{\text{tst}}))] = \\
        &= \Tr(\mat{T}^c \mat{L}\mat{L}^{\top}) + 
        \mathbb{E}[\Tr(\mat{D}^{-1} \mat{K}_1^{-1} \mat{Z} (\mat{P}^{\top}\mat{P})^{-\top} \mat{D}_{I^x}^2 (\mat{P}^{\top}\mat{P})^{-1} \mat{Z}^{\top} \mat{K}_1^{-1} \mat{D}^{-1} \mat{L} \mat{L}^{\top})].
    \end{align*}
    Using Lemma \ref{appendix:section3:wo_skip:lma2} and Lemma 7, 8 of \cite{kausik2024doubledescentoverfittingnoisy}, we get that the second term is equivalent to 
    \begin{align*}
        &\eta_{\text{trn}}^4 \Tr(\mat{D}^{-4} (\eta_{\text{trn}}^2 \mat{D}^{-2} + \mathbb{I}_r)^{-2} \mat{T} \mat{L}\mat{L}^{\top}) + o(1) \\
        &= \eta_{\text{trn}}^4 \Tr((\mat{D}^{2} + \eta_{\text{trn}}^2 \mathbb{I}_r)^{-2} T \mat{L}\mat{L}^{\top}) + O(d^{-1}) + o(1).
    \end{align*}
The $o(1)$ term refers to the error incurred when approximating the expectation of products by the product of expectations.
    To summarize the result, we have that 
    \begin{align*}
        &\frac{1}{N_{\text{tst}}}\mathbb{E}[\|\mat{X}_{\text{tst}} - \mat{W}_c \mat{X}_{\text{tst}}\|_F^{2}] \\
        &= \frac{1}{N_{\text{tst}}}\left(\Tr(\mat{T}^c \mat{L}\mat{L}^{\top}) + \Tr((\eta_{\text{trn}}^{-2} \mat{D}^2 + \mathbb{I}_r)^{-2} \mat{T} \mat{L}\mat{L}^{\top}) + O(d^{-1}) + o(1)\right) \\ 
        &= \frac{1}{N_{\text{tst}}}\Tr(\mat{J} \mat{L}\mat{L}^{\top}) + O(d^{-1}) + o(1).
    \end{align*}
    \rightline{$\square$}

Combining Lemmas~\ref{appendix:section3:wo_skip:lma3} and ~\ref{appendix:section3:wo_skip:lma4} within the decomposition in Equation~\eqref{appendix:decomposition_woskip} yields the desired result.

\subsubsection{Proof of Theorem \ref{theorem:test_risk_skip}}
We first decompose the test risk (\ref{test_metric_skip}):
\begin{align}\label{skip_decomposition}
    &\mathbb{E}\left[\frac{1}{N_{\text{tst}}}\|\mat{X}_{\text{tst}} - (\mat{W}^{\text{sc}}_c + \mathbb{I}_d)(\mat{X}_{\text{tst}} + \mat{A}_{\text{tst}})\|_F^2\right] \nonumber\\
    &= \frac{1}{N_{\text{tst}}}\mathbb{E}[\Tr(\mat{A}_{\text{tst}}\mat{A}_{\text{tst}}^{\top} + 2 \mat{A}_{\text{tst}}(\mat{X}_{\text{tst}} + \mat{A}_{\text{tst}})^{\top} (\mat{W}^{\text{sc}}_c)^{\top} + \nonumber\\
    &\mat{W}_c^{\text{sc}}(\mat{X}_{\text{tst}} + \mat{A}_{\text{tst}})(\mat{X}_{\text{tst}} + \mat{A}_{\text{tst}})^{\top} (\mat{W}^{\text{sc}}_c)^{\top})] \nonumber\\
    &= \frac{1}{N_{\text{tst}}}\mathbb{E}_{\mat{A}_{\text{trn}}}\left[\frac{N_{\text{tst}}\eta_{\text{tst}}^2}{d}\left(\Tr(\mathbb{I}_d) + \Tr(\mat{W}_c^{\text{sc}}) + \|\mat{W}^{\text{sc}}_c \|_F^2\right) + \|\mat{W}^{\text{sc}}_c \mat{X}_{\text{tst}} \|_F^2\right] \nonumber\\
    &= \eta_{\text{tst}}^2 + \mathbb{E}_{\mat{A}_{\text{trn}}}\left[\Tr\left(\frac{2 \eta_{\text{tst}}^2}{d} \mat{W}_c^{\text{sc}}\right) + \frac{\eta_{\text{tst}}^2}{d}\|\mat{W}_c^{\text{sc}}\|_F^2 + \frac{1}{N_{\text{tst}}}\|\mat{W}_c^{\text{sc}} \mat{X}_{\text{tst}} \|_F^2 \right]. 
\end{align}
We aim to prove Lemmas~\ref{appendix:section3:with_skip:lma7} (the first term), ~\ref{appendix:section3:with_skip:lma6}(the second term), and Lemma~\ref{appendix:section3:with_skip:lma8}, in order to establish Theorem~\ref{theorem:solution_noskip}.

We begin with couple of technical lemmas that are necessary for the proof of this theorem. Recall that critical points satisfy $\mat{W}^{\text{sc}}_c = - P_{I^a}(\mat{A})(\mat{X + A})^{\dagger}$. 
By a slight abuse of notation, we write $\mat{W}^{\text{sc}}_c = - \mat{A}_{I^a}(\mat{X + A})^{\dagger}$. 

\begin{lemma}\label{appendix:section3:with_skip:lma0}\textnormal{(\cite[Lemma 3]{kausik2024doubledescentoverfittingnoisy}).} Consider $\vec{a} \in \mathbb{R}^{d}$ and $\vec{b} \in \mathbb{R}^d$, and uniform random orthogonal matrix $\mat{Q} \in \mathbb{R}^{d \times d}$.
    If $\langle \vec{a}, \vec{b} \rangle = 0$, then $\mathbb{E}[(\mat{Q}\vec{a})_i(\mat{Q}\vec{b})_i] = 0$.
\end{lemma}
\textit{Proof.\quad}
Note that $\langle \mat{Q}\vec{a}, \mat{Q}\vec{b} \rangle = 0$. Then, $\sum\limits_{i=1}^d \mathbb{E}[(\mat{Q}\vec{a})_i (\mat{Q}\vec{b})_i] = 0$. 
Due to symmetry of $Q$, $\mathbb{E}[(\mat{Q}\vec{a})_i (\mat{Q}\vec{b})_i] = \mathbb{E}[(\mat{Q}\vec{a})_j (\mat{Q}\vec{b})_j] = 0$, for any $1 \leq i, j \leq d$.

\rightline{$\square$}

\begin{lemma}\label{appendix:section3:with_skip:lma1}
    Consider an unit vector $\vec{a} \in \mathbb{R}^{d}$, and an uniform random orthogonal matrix $\mat{Q} \in \mathbb{R}^{d \times d}$. 
    Then, $\mathbb{E}\left[ (\mat{Q} \vec{a})_i (\mat{Q} \vec{a})_i \right] = \frac{1}{d}$.
\end{lemma}

\textit{Proof.\quad}
Note that $\mathbb{E}\left[ \vec{a}^{\top} \mat{Q} \mat{Q}^{\top} \vec{a} \right] = 1$. 
Then, $\sum\limits_{i=1}^d \mathbb{E}\left[ \vec{a}^{\top} \vec{q}_i \vec{q}_i^{\top} \vec{a} \right] = 1$. 
Then the result follows from the symmetry of $\mat{Q}$.

\rightline{$\square$}

\begin{lemma}\label{appendix:section3:with_skip:lma2}
    For $c = \frac{d}{n}$, $\mathbb{E}[\mat{H} \mat{A}_{I^a} \mat{A}^{\dagger} \mat{H}^{\top}] = \frac{|I^a|}{n}\frac{1}{c - 1}\mathbb{I} + o(\frac{|I^a|}{n})$ with element wise variance of $O(\frac{k}{N^2})$.
\end{lemma}
\textit{Proof.\quad}
For $\mat{A} = \mat{U}_\mat{A} \mat{\Sigma}_\mat{A} \mat{V}_\mat{A}^{\top}$, 
we define a diagonal matrix $\mat{T}_a \in \mathbb{R}^{d \times d}$, such that 
$$(\mat{T}_a)_{ii} = \begin{cases}
    (\sigma^\mat{A}_i)^{-2} & \text{if } i \in I^a \\
    0 & \text{otherwise.}
\end{cases}
$$
Then, we can write
$\mathbb{E}[\mat{H}\mat{A}_{I^a}\mat{A}^{\dagger}\mat{H}^{\top}] = \mathbb{E}[\tilde{\mat{V}}^{\top}\mat{A}^{\dagger}\mat{A}_{I^a}\mat{A}^{\dagger}(\mat{A}^{\dagger})^{\top}\tilde{\mat{V}}] = \mathbb{E}[\tilde{\mat{V}}^{\top}\mat{V}_\mat{A} \mat{T}_a \mat{V}_\mat{A}^{\top}\tilde{\mat{V}}]$.
Observe $(\tilde{\mat{V}}^{\top}\mat{V}_\mat{A} \mat{T}_a \mat{V}_\mat{A}^{\top}\tilde{\mat{V}})_{ij} = \vec{v}_i^{\top}\mat{V}_\mat{A} \mat{T}_a \mat{V}_\mat{A}^{\top}\vec{v}_j$, which is equivalent to
$\vec{a}_i^{\top} \mat{T}_a \vec{a}_j$, for $\vec{a}_i := \vec{v}_i^{\top} \mat{V}_\mat{A}$.
For $i \neq j$, this expression evaluates to $0$ due to Lemma \ref{appendix:section3:with_skip:lma0}.
On the other hand, if $i = j$, then $\mathbb{E}\left[\vec{a}_i^{\top} \mat{T}_a \mathbf{a}_i\right] = \sum\limits_{l=1}^{|I^a|} \mathbb{E}\left[(\vec{a}_i)_l^2\right] \mathbb{E}\left[(\sigma^\mat{A}_l)^{-2}\right] 
= \frac{k}{n}\left(\frac{1}{c - 1} + o(1)\right)$. 
This follows from the fact that 
$\mat{A}$ is a Gaussian matrix, for which the matrix of singular values is independent of the singular vectors.
The final equality results from a direct evaluation of the Stieltjes transform of the Marchenko--Pastur distribution (\cite[Lemma 5]{sonthalia2022training}). 
\\
For the variance part, we consider 
$\mathbb{E}\left[\sum\limits_{m = 1}^{|I^a|}\sum\limits_{l = 1}^{|I^a|} (\vec{a}_i)_m (\vec{a}_j)_m (\vec{a}_i)_l (\vec{a}_j)_l (\sigma^\mat{A}_l)^{-2} (\sigma^\mat{A}_m)^{-2} \right]$,
which matches the variance computation in Lemma 4 of~\cite{kausik2024doubledescentoverfittingnoisy}, with $N$ replaced by $|I^a|$. As a result, the element-wise variance is of order $O(\frac{|I^a|}{n^2})$.

\rightline{$\square$}

\begin{lemma}\label{appendix:section3:with_skip:lma3}
    $\mat{A}_{I^a}\mat{A}^+(\mat{P}^{\dagger})^{\top} = 0$, $\Tr(\mat{A}_{I^a}\mat{A}^{\dagger}\tilde{\mat{U}}\mat{D} \mat{P}^{\dagger}) = 0$, and $\mat{P}^{\dagger} \mat{H}^{\top} = 0$.
\end{lemma}
\textit{Proof.\quad} For the first term, note that $\mat{A}_{I^a}\mat{A}^{\dagger}(\mat{P}^{\dagger})^{\top} = \mat{A}_{I^a}\mat{A}^{\dagger}\mat{P}(\mat{P}^{\top}\mat{P})^{-\top}$.
Also, it applies that $\mat{P} = - (\mathbb{I}_d - \mat{A}\mat{A}^{\dagger})\tilde{\mat{U}} \mat{D}$.
Then, 
\begin{align*} 
    \mat{A}_{I^a}\mat{A}^{\dagger}\mat{P} &= -\mat{A}_{I^a}\mat{A}^{\dagger}\tilde{\mat{U}}\mat{D} +  \mat{A}_{I^a}\mat{A}^{\dagger}\mat{A}\mat{A}^{\dagger}\tilde{\mat{U}}\mat{D} \\
&= - \mat{A}_{I^a}\mat{A}^{\dagger}\tilde{\mat{U}}\mat{D} +  \mat{A}_{I^a}\mat{A}^{\dagger}\tilde{\mat{U}}\mat{D} = 0.
\end{align*}
We can prove $\Tr(\mat{A}_{I^a}\mat{A}^{\dagger}\tilde{\mat{U}}\mat{D} \mat{P}^{\dagger}) = 0$ similarly. 
$\mat{P}^{\dagger} \mat{H}^{\top} = 0$ was already proved in Lemma 9 of \cite{kausik2024doubledescentoverfittingnoisy}.

\rightline{$\square$}

\begin{lemma}\label{appendix:section3:with_skip:lma4}
    For $c > 1$, $\mathbb{E}[\tilde{\mat{V}}^{\top} \mat{A}^{\dagger}\mat{A}_{I^a}\mat{A}^{\dagger}\tilde{\mat{U}}] = \mathbb{E}[\mat{H}\mat{A}_{I^a}\mat{A}^{\dagger}\tilde{\mat{U}}] = 0$, with element-wise variance $O(\frac{|I^a|}{dn})$.
\end{lemma}

\textit{Proof.} 
For notational convenience, we define 
$$
\mat{T}_a = \begin{cases}
    (\sigma^\mat{A}_i)^{-1} & \text{if } i \in I^a \\
    0 & \text{otherwise .}
\end{cases}
$$
Then, $\mathbb{E}[(\tilde{\mat{V}}^{\top}\mat{A}^{\dagger}\mat{A}_{I^a}\mat{A}^{\dagger}\tilde{\mat{U}})_{ij}] = \mathbb{E}[(\tilde{\mat{V}}^{\top} \mat{A}_{I^a}^{\dagger} \tilde{\mat{U}})_{ij}] = 
\mathbb{E}\left[\mathbf{v}_i^{\top} \mat{V}_\mat{A} \mat{T}_a \mat{U}_\mat{A}^{\top} \mathbf{u}_j\right]$.
This is equal to
$\mathbb{E}\left[\mathbf{a}^{\top} \mat{T}_a \mathbf{b}\right] = 
\mathbb{E}\left[\sum_{l=1}^{|I^a|} (\sigma^\mat{A}_l)^{-1}\mathbf{a}_l \mathbf{b}_l\right] = 0$, for $\mathbf{a} := \mathbf{v}_i^{\top} \mat{V}_\mat{A}$, $\mathbf{b} := \mathbf{u}_j^{\top} \mat{U}_\mat{A}$. 
For the variance part, if $i \neq j$,
\begin{align*}
\mathbb{E}\left[(\tilde{\mat{V}}^{\top}\mat{A}^{\dagger}\mat{A}_{I^a}\mat{A}^{\dagger}\tilde{\mat{U}})_{ij}^2\right] &= 
    \mathbb{E}\left[\sum_{l=1}^{|I^a|} (\sigma^\mat{A}_l)^{-2}\mathbf{a}_l^2 \mathbf{b}_l^2\right]\\
    &= \frac{|I^a|}{dn}\left(\frac{1}{c - 1} + o(1)\right) = O\left(\frac{|I^a|}{dn}\right).
\end{align*}
If $i = j$, 
\begin{align*}
    \mathbb{E}\left[(\tilde{\mat{V}}^{\top}\mat{A}^{\dagger}\mat{A}_{I^a}\mat{A}^{\dagger}\tilde{\mat{U}})_{ij}^2\right] &= 
    \mathbb{E}\left[\sum_{l=1}^{|I^a|} (\sigma^\mat{A}_l)^{-2}\mathbf{a}_l^4 \right]\\
    &= \frac{3|I^a|}{d(d + 2)}\left(\frac{1}{c - 1} + o(1)\right) = O\left(\frac{|I^a|}{d^2}\right).
\end{align*}

\rightline{$\square$}

\begin{lemma}\label{appendix:section3:with_skip:lma5}
    $\mathbb{E}[\tilde{\mat{U}}^{\top}\mat{A}_{I^a}\mat{A}^{\dagger}\tilde{\mat{U}}] = \frac{|I^a|}{d} \mathbb{I}_r$, with its element-wise variance $O(\frac{|I^a|^2}{d^{2}})$.
\end{lemma}
\textit{Proof.\quad} 
For notational convenience, we write
$$
\mat{T}_a = \begin{cases}
    1 & \text{if } i \in I^a \\
    0 & \text{otherwise.}
\end{cases}
$$
Then, $\mathbb{E}[(\tilde{\mat{U}}^{\top}\mat{A}_{I^a}\mat{A}^{\dagger}\tilde{\mat{U}})_{ij}] = \mathbb{E}[\mathbf{u}_i^{\top} \mat{U}_\mat{A} \mat{T}_a \mat{U}_\mat{A}^{\top} \mathbf{u}_j]$.
Thus with Lemma \ref{appendix:section3:with_skip:lma0}, this is $0$ if $i \neq j$.
On the other hand, if $i = j$, then this is $\frac{|I^a|}{d}$. For the variance part, first assume that $i \neq j$. Next, observe that for uniformly distributed random vectors $\mathbf{a} := \mat{U}_\mat{A}^{\top} \mathbf{u}_i, \mathbf{b} := \mat{U}_\mat{A}^{\top} \mathbf{u}_j$, we have the following:
$$\mathbb{E}\left[(\tilde{\mat{U}}^{\top}\mat{A}_{I^a}\mat{A}^{\dagger}\tilde{\mat{U}})_{ij}^2\right] = 
\mathbb{E}\left[(\mathbf{a}^{\top} \mat{T}_a \mathbf{b})^2\right]
= \mathbb{E}\left[\sum\limits_{l=1}^{|I^a|} \sum\limits_{m=1}^{|I^a|} \vec{a}_l\vec{b}_l \vec{a}_m \vec{b}_m\right].
$$
Note that $\langle \mathbf{a}, \mathbf{b} \rangle = 0$. 
This is because
\begin{align*}
0 \leq \sum\limits_{l=1}^{|I^a|} (a_lb_l) \sum\limits_{m=1}^{|I^a|} (\vec{a}_m \vec{b}_m) &\leq 
\sum\limits_{l=1}^d (\vec{a}_l \vec{b}_l) \sum\limits_{m=1}^d (\vec{a}_m \vec{b}_m) \\
&= \langle \mathbf{a}, \mathbf{b} \rangle^2 \\
&= 0.
\end{align*}
The first inequality satisties due to $\sum\limits_{l=1}^{|I^a|} (\vec{a}_l \vec{b}_l) = \sum\limits_{h=1}^{|I^a|} \mathbf{u}_i^{\top} (\mathbf{u}^\mat{A}_h)(\mathbf{u}^\mat{A}_h)^{\top} \mathbf{u}_j$, and
each $(\mathbf{u}^\mat{A}_h)(\mathbf{u}^\mat{A}_h)^{\top}$ is a positive semi-definite matrix, which means 
$\vec{u}_i^{\top} \vec{u}^\mat{A}_h (\vec{u}^\mat{A}_h)^{\top} \vec{u}_j \geq 0$. Thus, adding extra terms will only increase this. 
With this, the above term is $0$.
Now for $i = j$, this is the case where $\mathbf{a} = \mathbf{b}$.
Then, we have that
\begin{align*}
    \mathbb{E}\left[\sum\limits_{l=1}^{|I^a|} \sum\limits_{m=1}^{|I^a|} a_l^2 \vec{a}_m^2\right] &= 
    \mathbb{E}\left[\sum\limits_{l=1}^{|I^a|} \vec{a}_l^4] + \mathbb{E}[\sum\limits_{l \neq |I^a|} \vec{a}_l^2 \vec{a}_m^2\right] \\
    &= |I^a| \times \frac{3}{d(d + 2)} + |I^a|\left(|I^a| - 1\right) \times \frac{1}{d(d + 2)} \\
    &= O\left(\frac{|I^a|^2}{d^2}\right).
\end{align*}

\rightline{$\square$}

From the decomposition of the test metric, we have the following lemmas that are directly relevant to Theorem \ref{theorem:test_risk_skip}.
\begin{lemma}\label{appendix:section3:with_skip:lma6}\textnormal{($\mathbb{E}[\|\mat{W}^{\text{sc}}_c\|_F^2]$ Term).} \\
    For $c := \frac{d}{n}$ and $d \geq n + r$, we obtain that
    \begin{align*}
        &\mathbb{E}[\|\mat{W}^{\text{sc}}_c\|_F^2] \\
        &= |I^a| + \frac{|I^a|}{d} \frac{c}{c - 1} \sum\limits_{i=1}^d \frac{\sigma_i^2}{(\eta_{\text{trn}}^2 + \sigma_i^2)}
        + \frac{|I^a|}{n} \frac{1}{c} \sum\limits_{i=1}^d \frac{\eta_{\text{trn}}^2 \sigma_i^2}{(\eta_{\text{trn}}^2 + \sigma_i^2)} + o(1).
    \end{align*}
\end{lemma}
\textit{Proof.\quad} We evaluate $\mat{E}[\|\mat{W}_c^{\text{sc}}\|_F^2] = \mat{E}[\Tr(\mat{W}_c^{\text{sc}}(\mat{W}_c^{\text{sc}})^{\top})]$. Following \ref{appendix:section3:proof_thm1}, we approximate the term using the concentration argument.
By using Corollary 2.1 of \cite{WEI20011}, 
\begin{align*}
    \mat{W}^{\text{sc}}_c &= - \mat{A}_{I^a}(\mat{A} + \tilde{\mat{U}}\mat{D} \tilde{\mat{V}}^{\top})^{\dagger} \nonumber \\
    &= - \mat{A}_{I^a}\mat{A}^{\dagger} - \mat{A}_{I^a}\mat{A}^{\dagger} \tilde{\mat{U}}\mat{D}\mat{P}^{\dagger} + \mat{A}_{I^a}\mat{A}^{\dagger} \mat{H}^{\top} \mat{K}_1^{-1} \mat{H} + \mat{A}_{I^a}\mat{A}^{\dagger} \mat{H}^{\top} \mat{K}_1^{-1} \mat{Z} \mat{P}^{\dagger} + \\
    &\mat{A}_{I^a}\mat{A}^{\dagger} \tilde{\mat{U}} \mat{D} (\mat{P}^{\top} \mat{P})^{-1} \mat{Z}^{\top} \mat{K}_1^{-1} \mat{H} + \mat{A}_{I^a} \mat{A}^{\dagger} \tilde{\mat{U}} \mat{D} (\mat{P}^{\top} \mat{P})^{-1} \mat{Z}^{\top} \mat{K}_1^{-1} \mat{Z} \mat{P}^{\dagger}.
\end{align*}
We now examine how each of the six terms behaves after expanding them by multiplying with $(\mat{W}^{\text{sc}}_c)^{\top}$. 
\begin{enumerate}
    \item Distributing $-\mat{A}_{I^a}\mat{A}^{\dagger}$: For a notational convenience, we define
    $$
    \mat{T}_a = \begin{cases}
        1 & \text{if } i \in I^a \\
        0 & \text{otherwise.}
    \end{cases}
    $$
    Due to the fact that $\mat{A}_{I^a}\mat{A}^{\dagger}(\mat{A}_{I^a}\mat{A}^{\dagger})^{\top} = \mat{U}_\mat{A} \mat{T}_a\mat{U}_\mat{A}^{\top} = \mat{A}_{I^a}\mat{A}^{\dagger}$, we have that
    \begin{align*}
        &-\Tr(\mat{A}_{I^a}\mat{A}^{\dagger}(\mat{W}^{\text{\text{sc}}}_{\text{c}})^{\top}) = \Tr(\mat{A}_{I^a}\mat{A}^{\dagger}(\mat{A}_{I^a}\mat{A}^{\dagger})^{\top}) - \Tr(\mat{A}_{I^a}\mat{A}^{\dagger}\mat{H}^\top(\mat{K}_1)^{-1}\mat{H}) - \\ 
        &\Tr(\mat{A}_{I^a}\mat{A}^{\dagger} \mat{H}^\top \mat{K}_1^{-\top} \mat{Z}(\mat{P}^{\top} \mat{P})^{-\top} \mat{D} \tilde{\mat{U}}^{\top}).
    \end{align*} 
    Note that, due to Lemma~\ref{appendix:section3:with_skip:lma3}, certain terms evaluate to zero and therefore do not appear in the expression.
    The first term $\mathbb{E}\left[ \Tr(\mat{A}_{I^a}\mat{A}^{\dagger}(\mat{A}_{I^a}\mat{A}^{\dagger})^{\top})\right] = \mathbb{E}[\Tr(\mat{U}_{\mat{A}} \mat{T}_a \mat{U}_{\mat{A}}^{\top})] = |I^a|$. The second term, with Lemma \ref{appendix:section3:with_skip:lma2},
     and Lemma 8 of \cite{kausik2024doubledescentoverfittingnoisy}, 
    \begin{align*}
    \Tr(\mat{A}_{I^a}\mat{A}^{\dagger}\mat{H}^\top(\mat{K}_1)^{-1}\mat{H}) &= \Tr(\mat{H}\mat{A}_{I^a}\mat{A}^{\dagger}\mat{H}^\top(\mat{K}_1)^{-1}) \\
        &= \frac{|I^a|}{n}\frac{\eta_{\text{trn}}^2}{c}\Tr((\eta_{\text{trn}}^2\mat{D}^{-2} + \mathbb{I}_r)^{-1}) + o\left(\frac{|I^a|}{n}\right).
    \end{align*}
    Because two terms have element-wise variance of $O(\frac{|I^a|}{n^2})$, and $O(d^{-1})$  respectively, 
    the element-wise estimation error of the whole term would be $o(1)$ ($O\left(\frac{\sqrt{|I^a|}}{d\sqrt{d}}\right)$ to be more precise).  
    \\ The third term $\Tr(\mat{A}_{I^a}\mat{A}^{\dagger} \mat{H}^\top \mat{K}_1^{-\top} \mat{Z}(\mat{P}^{\top} \mat{P})^{-\top} \mat{D} \tilde{\mat{U}}^{\top})$ 
    has mean of $0$ due to Lemma \ref{appendix:section3:with_skip:lma4}. 
    Thus we only need to take a look at the element-wise variance. Due to the cyclic invariance of trace,
    \begin{align*}
        &\Tr(\mat{A}_{I^a}\mat{A}^{\dagger} \mat{H}^\top \mat{K}_1^{-\top} \mat{Z}(\mat{P}^{\top} \mat{P})^{-\top} \mat{D} \tilde{\mat{U}}^{\top}) \\
        &= \Tr( \tilde{\mat{U}}^{\top} \mat{A}_{I^a}\mat{A}^{\dagger} \mat{H}^\top \mat{K}_1^{-\top} \mat{Z}(\mat{P}^{\top} \mat{P})^{-\top} \mat{D}) \\
        &= \Tr(\tilde{\mat{U}}^{\top} \mat{A}_{I^a}\mat{A}^{\dagger} \mat{H}^\top \mat{K}_1^{-\top} \mat{Z}\mat{D}^{-1} (\mat{D}(\mat{P}^{\top} \mat{P})^{-\top} \mat{D}))
    \end{align*}.
     From \cite{kausik2024doubledescentoverfittingnoisy},
    the element-wise variances of $\mat{K}^{-1}, \mat{Z}\mat{D}^{-1}, \mat{D}(\mat{P}^{\top} \mat{P})^{-1} \mat{D}$ are all $O(d^{-1})$. 
    Thus applying the concentration argument 
    gives us the total element wise variance of $O(d^{-1})$. 
    
    To summarize the result of distributing $- \mat{A}_{I^a}\mat{A}^{\dagger}$,  this can be written down as $|I^a| + \frac{|I^a|}{n} \frac{\eta_{\text{trn}}^2}{c} \Tr((\eta_{\text{trn}}^2 \mat{D}^{-2} + \mathbb{I}_r)^{-1}) + O(d^{-1})$.
    
    \item Distributing $-\mat{A}_{I^a}\mat{A}^{\dagger} \tilde{\mat{U}} \mat{D} \mat{P}^{\dagger}$: This is
    \begin{align*}
        &-\Tr(\mat{A}_{I^a}\mat{A}^{\dagger}\tilde{\mat{U}} \mat{D} \mat{P}^{\dagger} (\mat{W}^{\text{sc}}_c)^\top) = \\ 
        &\Tr(\mat{A}_{I^a}\mat{A}^{\dagger} \tilde{\mat{U}} \mat{D} \mat{P}^{\dagger} (\tilde{\mat{U}} \mat{D} \mat{P}^{\dagger})^{\top})  - \Tr(\mat{A}_{I^a}\mat{A}^{\dagger} \tilde{\mat{U}} \mat{D} \mat{P}^{\dagger} (\mat{P}^{\dagger})^{\top} \mat{Z}^{\top} \mat{K}_1^{-\top} \mat{H}) - \\
        &\Tr(\mat{A}_{I^a}\mat{A}^{\dagger} \tilde{\mat{U}} \mat{D} \mat{P}^{\dagger} (\mat{P}^{\dagger})^{\top} \mat{Z}^{\top} \mat{K}_1^{-\top} \mat{Z} (\mat{P}^{\top}\mat{P})^{-\top} \mat{D} \tilde{\mat{U}}^{\top})
    \end{align*}
    Note that some terms do not appear here due to Lemma \ref{appendix:section3:with_skip:lma3}.
    The second term has also mean $0$ due to Lemma \ref{appendix:section3:with_skip:lma4}, 
    thus only the variance needs to be bounded for this term. Due to the cyclic invariance of trace,
    \begin{align*}
    &- \Tr(\mat{A}_{I^a}\mat{A}^{\dagger} \tilde{\mat{U}} \mat{D} \mat{P}^{\dagger} (\mat{P}^{\dagger})^{\top} \mat{Z}^{\top} \mat{K}_1^{-\top} \mat{H}) \\
    &= - \Tr(\mat{H}\mat{A}_{I^a}\mat{A}^{\dagger} \tilde{\mat{U}} \mat{D} \mat{P}^{\dagger} (\mat{P}^{\dagger})^{\top} \mat{Z}^{\top} \mat{K}_1^{-\top}) \\
    &= - \Tr(\mat{H}\mat{A}_{I^a}\mat{A}^{\dagger} \tilde{\mat{U}} (\mat{D} (\mat{P}^{\top} \mat{P})^{-1} \mat{D})(\mat{D}^{-1} \mat{Z}^{\top}) \mat{K}_1^{-\top}).
    \end{align*}
    Thus from Lemmas 6, 7, 8 from \cite{kausik2024doubledescentoverfittingnoisy}, $\mat{D} \mat{P}^{\dagger} (\mat{P}^{\dagger})^{\top} \mat{D}$, $\mat{D}^{-1} \mat{Z}^{\top}$, and $\mat{K}_1^{-1}$ have variance of $O(d^{-1})$. 
    From the concentration argument, this has an error rate of $O(d^{-1})$. \\
    The first term is $\Tr(\mat{A}_{I^a}\mat{A}^{\dagger} \tilde{\mat{U}} \mat{D} \mat{P}^{\dagger} (\mat{U} \mat{D} \mat{P}^{\dagger})^{\top}) = \Tr(\tilde{\mat{U}}^{\top} \mat{A}_{I^a}\mat{A}^{\dagger} \tilde{\mat{U}} \mat{D} \mat{P}^{\dagger} (\mat{P}^{\dagger})^{\top} \mat{D}) = \Tr(\tilde{\mat{U}}^{\top} \mat{A}_{I^a}\mat{A}^{\dagger} \tilde{\mat{U}} \mat{D} (\mat{P}^{\top} \mat{P})^{-1} \mat{D})$.
    From Lemma \ref{appendix:section3:with_skip:lma5}, and Lemma 6 from \cite{kausik2024doubledescentoverfittingnoisy}, we have that
    \begin{align}
        \Tr(\tilde{\mat{U}}^{\top} \mat{A}_{I^a}\mat{A}^{\dagger} \tilde{\mat{U}} \mat{D} (\mat{P}^{\top} \mat{P})^{-1} \mat{D}) = \frac{|I^a|}{d}\frac{c}{c - 1}\Tr(\mathbb{I}_r) + O\left(\frac{|I^a|}{dn}\right)
    \end{align}
    with element-wise variance $O\left(\frac{|I^a|}{d \sqrt{d}}\right)$.
    The final term is 
    \begin{align*}
    &\Tr(\mat{A}_{I^a}\mat{A}^{\dagger} \tilde{\mat{U}} \mat{D} \mat{P}^{\dagger}(\mat{P}^{\dagger})^{\top}\mat{Z}^{\top} \mat{K}_1^{-\top} \mat{Z}(\mat{P}^{\top} \mat{P})^{-\top} \mat{D} \tilde{\mat{U}}^{\top}) \\
    &= \Tr( (\tilde{\mat{U}}^{\top} \mat{A}_{I^a}\mat{A}^{\dagger} \tilde{\mat{U}}) (\mat{D}(\mat{P}^{\top} \mat{P})^{-1}\mat{D}) \mat{D}^{-1} \mat{Z}^{\top} \mat{K}_1^{-\top} (\mat{Z} \mat{D}^{-1}) (\mat{D} (\mat{P}^{\top} \mat{P})^{-\top} \mat{D})).
    \end{align*}.
    Using Lemma \ref{appendix:section3:with_skip:lma5}, and Lemmas 6, 7, 8 from \cite{kausik2024doubledescentoverfittingnoisy}, we have that
    \begin{align*}
        &- \Tr( (\tilde{\mat{U}}^{\top} \mat{A}_{I^a}\mat{A}^{\dagger} \tilde{\mat{U}}) (\mat{D} \mat{P}^{\top} \mat{P} \mat{D})^{-1} \mat{D}^{-1} \mat{Z}^{\top} \mat{K}_1^{-\top} (\mat{Z} \mat{D}^{-1}) (\mat{D} (\mat{P}^{\top} \mat{P})^{-\top} \mat{D})) \\
        &= - \frac{|I^a|}{d}\frac{\eta_{\text{trn}}^2 c}{c - 1}\Tr(\mat{D}^{-2} (\eta_{\text{trn}}^2 \mat{D}^{-2} + \mathbb{I}_r)^{-1}) + O\left(\frac{|I^a|}{d^2}\right)
    \end{align*}
    with its element-wise variance $O(d^{-1})$.
    Thus, in total, this can be written as 
    $$
    \frac{|I^a|}{d}\frac{c}{c - 1}\Tr(\mat{D}^2 (\eta_{\text{trn}}^2 \mathbb{I}_r + \mat{D}^2)^{-1}) + O(\frac{|I^a|}{d^2})
    $$
    \item Distributing $\mat{A}_{I^a}\mat{A}^{\dagger} \mat{H}^\top \mat{K}_1^{-1} 
    \mat{H}$: This is
    \begin{align*}
        &\Tr(\mat{A}_{I^a}\mat{A}^{\dagger} \mat{H}^\top \mat{K}_1^{-1} \mat{H} (\mat{W}^{\text{\text{sc}}}_{\text{c}})^{\top}) = - \Tr(\mat{A}_{I^a}\mat{A}^{\dagger} \mat{H}^\top \mat{K}_1^{-1} \mat{H})  \\
        &+ \Tr(\mat{A}_{I^a}\mat{A}^{\dagger} \mat{H}^\top \mat{K}_1^{-1} \mat{H} \mat{H}^\top \mat{K}_1^{-1} \mat{H}) + \\
        &\Tr(\mat{A}_{I^a}\mat{A}^{\dagger} \mat{H}^\top \mat{K}_1^{-1} \mat{H} \mat{H}^\top \mat{K}_1^{-1} \mat{Z} (\mat{P}^{\top} \mat{P})^{-1} \mat{D} \tilde{\mat{U}}^{\top}) 
    \end{align*}
    Again, using Lemma \ref{appendix:section3:with_skip:lma3}, some terms are filtered out. The first term is, from Lemma \ref{appendix:section3:with_skip:lma2}, and Lemma 8 of \cite{kausik2024doubledescentoverfittingnoisy},
    \begin{align*}
        -\Tr(\mat{A}_{I^a}\mat{A}^{\dagger} \mat{H}^\top \mat{K}_1^{-1} \mat{H})
        &= - \frac{|I^a|}{n}\frac{\eta_{\text{trn}}^2}{c}\Tr((\eta_{\text{trn}}^2 \mat{D}^{-2} + \mathbb{I}_r)^{-1}) + o\left(\frac{|I^a|}{n}\right).
    \end{align*},
    with element-wise variance of $O\left(\frac{\sqrt{|I^a|}}{N\sqrt{d}}\right)$. Due to Lemma \ref{appendix:section3:with_skip:lma2},
     and Lemma 4, 8 of \cite{kausik2024doubledescentoverfittingnoisy}, the second term is 
    \begin{align*}
        \Tr(\mat{A}_{I^a}\mat{A}^{\dagger} \mat{H}^\top \mat{K}_1^{-1} \mat{H} \mat{H}^\top \mat{K}_1^{-1} \mat{H}) &= \Tr((\mat{H}\mat{A}_{I^a}\mat{A}^{\dagger} \mat{H}^\top) \mat{K}_1^{-1} (\mat{H} \mat{H}^\top) \mat{K}_1^{-1})\\
        &= \frac{|I^a|}{n}\frac{\eta_{\text{trn}}^2}{c}\Tr((\eta_{\text{trn}}^2 \mat{D}^{-2} + \mathbb{I}_r)^{-2}) + o(\frac{|I^a|}{n}).
    \end{align*}
    with element-wise variance of $O(d^{-1})$. The final term is, due to Lemma \ref{appendix:section3:with_skip:lma4},
    has mean of $0$, thus only the variance needs to be bounded. 
    Due to the cyclic invariance of trace, this is
    \begin{align*}
        &\Tr(\mat{A}_{I^a}\mat{A}^{\dagger} \mat{H}^\top \mat{K}_1^{-1} \mat{H} \mat{H}^\top \mat{K}_1^{-1} \mat{Z} (\mat{P}^{\top} \mat{P})^{-1} \mat{D} \tilde{\mat{U}}^{\top}) \\
        &= \Tr(\tilde{\mat{U}}^{\top} \mat{A}_{I^a}\mat{A}^{\dagger} \mat{H}^\top \mat{K}_1^{-1} \mat{H} \mat{H}^\top \mat{K}_1^{-1} \mat{Z} (\mat{P}^{\top} \mat{P})^{-1} \mat{D}) \\
        &= \Tr((\tilde{\mat{U}}^{\top} \mat{A}_{I^a}\mat{A}^{\dagger} \mat{H}^\top) \mat{K}_1^{-1} (\mat{H} \mat{H}^\top) \mat{K}_1^{-1} (\mat{Z} \mat{D}^{-1}) (\mat{D} (\mat{P}^{\top} \mat{P})^{-1} \mat{D})).
    \end{align*}
    Thus, from Lemma \ref{appendix:section3:with_skip:lma4}, and Lemma 4, 6, 7, 8 of \cite{kausik2024doubledescentoverfittingnoisy},
    this term has element-wise variance of $O(d^{-1})$.
    Then we have in total, this term is $\frac{k}{n}\frac{\eta_{\text{trn}}^2}{c}\Tr((\eta_{\text{trn}}^2 \mat{D}^{-2} + \mathbb{I}_r)^{-2}) - \frac{|I^a|}{n}\frac{\eta_{\text{trn}}^2}{c}\Tr((\eta_{\text{trn}}^2 \mat{D}^{-2} + \mathbb{I}_r)^{-1}) + O(d^{-1})$.
    \item Distributing $\mat{A}_{I^a}\mat{A}^{\dagger} \mat{H}^\top \mat{K}_1^{-1} \mat{Z} \mat{P}^{\dagger}$: This is
    \begin{align*}
        &\Tr(\mat{A}_{I^a}\mat{A}^{\dagger} \mat{H}^\top \mat{K}_1^{-1} \mat{Z} \mat{P}^{\dagger} (\mat{W}^{\text{\text{sc}}}_{\text{c}})^{\top}) =  -\Tr(\mat{A}_{I^a}\mat{A}^{\dagger} \mat{H}^\top \mat{K}_1^{-1} \mat{Z} \mat{P}^{\dagger} (\mat{P}^{\dagger})^{\top} \mat{D} \mat{U}^{\top}) + \\
        & \Tr(\mat{A}_{I^a}\mat{A}^{\dagger} \mat{H}^\top \mat{K}_1^{-1} \mat{Z} \mat{P}^{\dagger} (\mat{P}^{\dagger})^{\top} \mat{Z}^{\top} \mat{K}_1^{-1} \mat{H}) + \\
        &\Tr(\mat{A}_{I^a}\mat{A}^{\dagger} \mat{H}^\top \mat{K}_1^{-1} \mat{Z} \mat{P}^{\dagger} (\mat{P}^{\dagger})^{\top} \mat{Z}^{\top} \mat{K}_1^{-1} \mat{Z} (\mat{P}^{\top} \mat{P})^{-1} \mat{D} \mat{U}^{\top})
    \end{align*}
From this point onward, since the proof follows a similar structure to previous arguments, we provide only a brief sketch.
The first term has zero mean, as shown in Lemma~\ref{appendix:section3:with_skip:lma4}.
Its variance can be bounded in the usual way by $O(d^{-1})$. 
    The second term is, by Lemma~\ref{appendix:section3:with_skip:lma2} and Lemmas 6, 7, and 8 of~\cite{kausik2024doubledescentoverfittingnoisy},
    \begin{align*}
        &\Tr(\mat{A}_{I^a}\mat{A}^{\dagger} \mat{H}^\top \mat{K}_1^{-1} \mat{Z} \mat{P}^{\dagger} (\mat{P}^{\dagger})^{\top} \mat{Z}^{\top} \mat{K}_1^{-1} \mat{H})\\
        &= \frac{|I^a|}{n}\frac{\eta_{\text{trn}}^4}{c}\Tr(\mat{D}^{-2} (\eta_{\text{trn}}^2 \mat{D}^{-2} + \mathbb{I}_r)^{-2}) + o\left(\frac{|I^a|}{n}\right).
    \end{align*}
    This has the element-wise variance of $O(d^{-1})$. The last term, by Lemma~\ref{appendix:section3:with_skip:lma4}, has zero mean, and applying the standard concentration argument yields an element-wise variance of $O(d^{-1})$.
    Combining all contributions, the total expression is $\frac{|I^a|}{n}\frac{\eta_{\text{trn}}^4}{c}\Tr(\mat{D}^{-2} (\eta_{\text{trn}}^2 \mat{D}^{-2} + \mathbb{I}_r)^{-2}) + O(d^{-1})$.
    \item Distributing $\mat{A}_{I^a}\mat{A}^{\dagger} \tilde{\mat{U}} \mat{D} (\mat{P}^{\top} \mat{P})^{-1} \mat{Z}^{\top} \mat{K}_1^{-1} \mat{H}$: This is
    \begin{align*}
        &\Tr(\mat{A}_{I^a}\mat{A}^{\dagger} \tilde{\mat{U}} \mat{D} (\mat{P}^{\top} \mat{P})^{-1} \mat{Z}^{\top} \mat{K}_1^{-1} \mat{H} (\mat{W}^{\text{\text{sc}}}_{\text{c}})^{\top}) = - \Tr(\mat{A}_{I^a}\mat{A}^{\dagger} \tilde{\mat{U}} \mat{D} (\mat{P}^{\top} \mat{P})^{-1} \mat{Z}^{\top} \mat{K}_1^{-1} \mat{H}) - \\
        &\Tr(\mat{A}_{I^a}\mat{A}^{\dagger} \tilde{\mat{U}} \mat{D} (\mat{P}^{\top} \mat{P})^{-1} \mat{Z}^{\top} \mat{K}_1^{-1} \mat{H} \mat{H}^\top \mat{K}_1^{-1} \mat{H}) + \\
        &\Tr(\mat{A}_{I^a}\mat{A}^{\dagger} \tilde{\mat{U}} \mat{D} (\mat{P}^{\top} \mat{P})^{-1} \mat{Z}^{\top} \mat{K}_1^{-1} \mat{H} \mat{H}^\top \mat{K}_1^{-1} \mat{Z} (\mat{P}^{\top} \mat{P})^{-1} \mat{D} \tilde{\mat{U}}^{\top})
    \end{align*}
The first and second terms have zero mean, due to Lemma~\ref{appendix:section3:with_skip:lma3}.
For the element-wise variance, Lemma~\ref{appendix:section3:wo_skip:lma3} and Lemmas 4, 6, 7, and 8 from~\cite{kausik2024doubledescentoverfittingnoisy} imply that it is of order $o(1)$. 
The final term, by the cyclic invariance of the trace, Lemma~\ref{appendix:section3:with_skip:lma5}, and again Lemmas 4, 6, 7, and 8 from~\cite{kausik2024doubledescentoverfittingnoisy}, satisfies that
    \begin{align*}
        &\Tr(\mat{A}_{I^a}\mat{A}^{\dagger} \tilde{\mat{U}} \mat{D} (\mat{P}^{\top} \mat{P})^{-1} \mat{Z}^{\top} \mat{K}_1^{-1} \mat{H} \mat{H}^\top \mat{K}_1^{-1} \mat{Z} (\mat{P}^{\top} \mat{P})^{-1} \mat{D} \tilde{\mat{U}}^{\top}) \\
        &= \frac{k}{d}\frac{\eta_{\text{trn}}^2 c}{c - 1}\Tr(\mat{D}^{-2} (\eta_{\text{trn}}^2\mat{D}^{-2} + \mathbb{I}_r)^{-2}) + o(\frac{k}{d}).
    \end{align*} This term has element-wise variance of $o(1)$.
    Therefore, the total contribution can be summarized as $\frac{|I^a|}{d}\frac{\eta_{\text{trn}}^2 c}{c - 1}\Tr(\mat{D}^{-2} (\eta_{\text{trn}}^2 \mat{D}^{-2} + \mathbb{I}_r)^{-2}) + o(1)$.
    \item Distributing $\mat{A}_{I^a}\mat{A}^{\dagger} \mat{U} \mat{D} (\mat{P}^{\top} \mat{P})^{-1} \mat{Z}^{\top} \mat{K}_1^{-1} \mat{Z} \mat{P}^{\dagger}$: This is
    \begin{align*}
        &\Tr(\mat{A}_{I^a}\mat{A}^{\dagger} \tilde{\mat{U}} \mat{D} (\mat{P}^{\top} \mat{P})^{-1} \mat{Z}^{\top} \mat{K}_1^{-1} \mat{Z} \mat{P}^{\dagger} (\mat{W}^{\text{\text{sc}}}_{\text{c}})^{\top}) = \\
        &- \Tr(\mat{A}_{I^a}\mat{A}^{\dagger} \tilde{\mat{U}} \mat{D} (\mat{P}^{\top} \mat{P})^{-1} \mat{Z}^{\top} \mat{K}_1^{-1} \mat{Z} \mat{P}^{\dagger} (\mat{P}^{\dagger})^{\top} \mat{D} \tilde{\mat{U}}^{\top}) + \\
        &\Tr(\mat{A}_{I^a}\mat{A}^{\dagger} \tilde{\mat{U}} \mat{D} (\mat{P}^{\top} \mat{P})^{-1} \mat{Z}^{\top} \mat{K}_1^{-1} \mat{Z} \mat{P}^{\dagger} (\mat{P}^{\dagger})^{\top} \mat{Z}^{\top} \mat{K}_1^{-1} \mat{H}) + \\
        &\Tr(\mat{A}_{I^a}\mat{A}^{\dagger} \tilde{\mat{U}} \mat{D} (\mat{P}^{\top} \mat{P})^{-1} \mat{Z}^{\top} \mat{K}_1^{-1} \mat{Z} \mat{P}^{\dagger} (\mat{P}^{\dagger})^{\top} \mat{Z}^{\top} \mat{K}_1^{-1} \mat{Z} (\mat{P}^{\top} \mat{P})^{-1} \mat{D} \tilde{\mat{U}}^{\top})
    \end{align*}
The second term has zero mean due to Lemma~\ref{appendix:section3:wo_skip:lma4}.
The first term, by Lemma~\ref{appendix:section3:with_skip:lma5} and Lemmas 6, 7, and 8 of~\cite{kausik2024doubledescentoverfittingnoisy}, evaluates to
    \begin{align*}
        &-\Tr(\mat{A}_{I^a}\mat{A}^{\dagger} \tilde{\mat{U}} \mat{D} (\mat{P}^{\top} \mat{P})^{-1} \mat{Z}^{\top} \mat{K}_1^{-1} \mat{Z} \mat{P}^{\dagger} (\mat{P}^{\dagger})^{\top} \mat{D} \tilde{\mat{U}}^{\top}) \\
        &= - \frac{|I^a|}{d}\frac{\eta_{\text{trn}}^2 c}{c - 1}\Tr(\mat{D}^{-2} (\eta_{\text{trn}}^2 \mat{D}^{-2} + \mathbb{I}_r)^{-1}) + o\left(\frac{|I^a|}{d}\right).
    \end{align*}
The last term, again by Lemma~\ref{appendix:section3:with_skip:lma5} and Lemmas 6, 7, and 8 of~\cite{kausik2024doubledescentoverfittingnoisy}, becomes
    \begin{align*}
        &\Tr(\mat{A}_{I^a}\mat{A}^{\dagger} \tilde{\mat{U}} \mat{D} (\mat{P}^{\top} \mat{P})^{-1} \mat{Z}^{\top} \mat{K}_1^{-1} \mat{Z} \mat{P}^{\dagger} (\mat{P}^{\dagger})^{\top} \mat{Z}^{\top} \mat{K}_1^{-1} \mat{Z} (\mat{P}^{\top} \mat{P})^{-1} \mat{D} \tilde{\mat{U}}^{\top}) \\
        &= \frac{|I^a|}{d}\frac{\eta_{\text{trn}}^4 c}{c - 1}\Tr(\mat{D}^{-4} (\eta_{\text{trn}}^2 \mat{D}^{-2} + \mathbb{I}_r)^{-2}) + o\left(\frac{|I^a|}{d}\right).
    \end{align*}
    All of these terms have variance of order $o(1)$.
    Thus, in total, we have
    \begin{align*}
        \frac{|I^a|}{d}\frac{\eta_{\text{trn}}^2 c}{c - 1}\Tr(\eta_{\text{trn}}^2 \mat{D}^{-4} (\eta_{\text{trn}}^2 \mat{D}^{-2} + \mathbb{I}_r)^{-2} - \mat{D}^{-2} (\eta_{\text{trn}}^2 \mat{D}^{-2} + \mathbb{I}_r)^{-1}) + o(1).  
    \end{align*}
\end{enumerate}
Now that all terms have been computed, we organize them to obtain the final expression for $\mathbb{E}[\|\mat{W}^{\text{sc}}_c\|_F^2]$.
\begin{align*}
    &\mathbb{E}[\|\mat{W}^{\text{sc}}_c\|_F^2] =\\
    &|I^a| + \frac{|I^a|}{d} \frac{c}{c - 1} \Tr(\mat{D}^2 (\eta_{\text{trn}}^2 \mathbb{I}_r + \mat{D}^2)^{-1}) 
    + \frac{|I^a|}{n} \frac{1}{c} \Tr((\mat{D}^{-2} + \eta_{\text{trn}}^{-2} \mathbb{I}_r)^{-1}) + o(1).
\end{align*}

\rightline{$\square$}

\begin{lemma}\label{appendix:section3:with_skip:lma7}\textnormal{($\mathbb{E}[\Tr(\mat{W}^{\text{sc}}_c)]$ Term.)} \\
    For $c := \frac{d}{n}$ and $d \geq n + r$,
    $$\mathbb{E}[\Tr(\mat{W}^{\text{\text{sc}}}_{\text{c}})] = - |I^a| + \frac{|I^a|}{n} \frac{1}{c} \sum\limits_{i=1}^d \frac{\eta_{\text{trn}}^2 \sigma_i^2}{(\eta_{\text{trn}}^2 + \sigma_i^2)} + O(d^{-1}).$$
\end{lemma}
\textit{Proof.\quad}
By using corollary 2.1 of \cite{WEI20011} of expanding $(\mat{X} + \mat{A})^{\dagger}$, we have that
\begin{align*}
    \Tr(\mat{W}^{\text{\text{sc}}}_{\text{c}}) &= -\Tr(\mat{A}_{I^a}\mat{A}^{\dagger}) - \Tr(\mat{A}_{I^a}\mat{A}^{\dagger} \tilde{\mat{U}} \mat{D} \mat{P}^{\dagger}) + \Tr(\mat{A}_{I^a}\mat{A}^{\dagger} \mat{H}^\top \mat{K}_1^{-1} \mat{H}) + \\
    &\Tr(\mat{A}_{I^a}\mat{A}^{\dagger} \mat{H}^\top \mat{K}_1^{-1} \mat{Z} \mat{P}^{\dagger}) + \Tr(\mat{A}_{I^a}\mat{A}^{\dagger} \tilde{\mat{U}} \mat{D} (\mat{P}^{\top} \mat{P})^{-1} \mat{Z}^{\top} \mat{K}_1^{-1} \mat{H}) + \\
    &\Tr(\mat{A}_{I^a}\mat{A}^{\dagger} \tilde{\mat{U}} \mat{D} (\mat{P}^{\top} \mat{P})^{-1} \mat{Z}^{\top} \mat{K}_1^{-1} \mat{Z} \mat{P}^{\dagger})
\end{align*}
Note that the first term $- \Tr(\mat{A}_{I^a} \mat{A}^{\dagger}) = - |I^a|$. The second, the fourth, and the sixth terms are $0$ 
due to lemma \ref{appendix:section3:with_skip:lma3}. The fifth term has mean of $0$ due to Lemma \ref{appendix:section3:with_skip:lma4}.
Finally, the third term is 
\begin{align*}
    \Tr(\mat{A}_{I^a}\mat{A}^{\dagger} \mat{H}^\top \mat{K}_1^{-1} \mat{H}) &= \Tr((\mat{H}\mat{A}_{I^a}\mat{A}^{\dagger} \mat{H}^\top) \mat{K}_1^{-1}) \\
    &= \frac{|I^a|}{n}\frac{\eta_{\text{trn}}^2}{c}\Tr((\eta_{\text{trn}}^2 \mat{D}^{-2} + \mathbb{I}_r)^{-1}) + o\left(\frac{|I^a|}{n}\right).
\end{align*}
Using standard concentration arguments, each of these terms exhibits an element-wise variance of order $O(d^{-1})$.

\rightline{$\square$}

\begin{lemma}\label{appendix:section3:with_skip:lma8}\textnormal{($\|\mat{W}^{\text{\text{sc}}}_{\text{c}}\mat{X}_{\text{tst}}\|_F^2$ Term).}
    \begin{align*}
        &\mathbb{E}\|(\mat{W}^{\text{\text{sc}}}_{\text{c}}) \mat{X}_{\text{tst}}\|_F^2 \\
        &= \frac{|I^a|}{d} \Tr\left(((c - 1) \mat{D}^2 + \mathbb{I}_d)(\mathbb{I}_r + \eta_{\text{trn}}^{-2} \mat{D}^2)^{-2} \mat{L}\mat{L}^{\top}\right) + O(d^{-1}). 
    \end{align*}
\end{lemma}

\textit{Proof.\quad}
From Corollary 2.1 of \cite{WEI20011}, we have that 
\begin{align}
    \mat{W}^{\text{\text{sc}}}_{\text{c}} &= - \mat{A}_{I^a}(\mat{A} + \tilde{\mat{U}}\mat{D} \tilde{\mat{V}}^{\top})^{\dagger} \nonumber \\
    &= - \mat{A}_{I^a}(\mat{A}^{\dagger} + \mat{A}^{\dagger}\tilde{\mat{U}}\mat{D} \mat{P}^{\dagger} - (\mat{A}^{\dagger}\mat{H}^\top + \mat{A}^{\dagger} \tilde{\mat{U}} \mat{D} (\mat{P}^{\top}\mat{P})^{-1} \mat{Z}^{\top}) \mat{K}_1^{-1} (\mat{H} + \mat{Z}\mat{P}^{\dagger})) \nonumber \\
    &= - \mat{A}_{I^a}\mat{A}^{\dagger} - \mat{A}_{I^a}\mat{A}^{\dagger} \tilde{\mat{U}}\mat{D} \mat{P}^{\dagger} + \mat{A}_{I^a} \mat{A}^{\dagger} \mat{H}^\top \mat{K}_1^{-1} \mat{H} + \mat{A}_{I^a} \mat{A}^{\dagger} \mat{H}^\top \mat{K}_1^{-1} \mat{Z} \mat{P}^{\dagger} + \nonumber \\
    &\mat{A}_{I^a}\mat{A}^{\dagger} \tilde{\mat{U}} \mat{D} (\mat{P}^{\top} \mat{P})^{-1} \mat{Z}^{\top} \mat{K}_1^{-1} \mat{H} + \mat{A}_{I^a} \mat{A}^{\dagger} \tilde{\mat{U}} \mat{D} (\mat{P}^{\top} \mat{P})^{-1} \mat{Z}^{\top} \mat{K}_1^{-1} \mat{Z} \mat{P}^{\dagger}
\end{align}

Using $\mat{X}_{\text{tst}} = \tilde{\mat{U}} \mat{L}$, with the fact that $\mat{P}^{\dagger} \tilde{\mat{U}} = - \mat{D}^{-1}$ and $\mat{H}\tilde{\mat{U}} = (\mat{Z} - \mathbb{I})\mat{D}^{-1}$, 
we have that, 
\begin{align*}
    \mat{W}^{\text{\text{sc}}}_{\text{c}} \mat{X}_{\text{tst}} &= - (\mat{A}_{I^a} \mat{A}^{\dagger} \mat{H}^\top \mat{K}_1^{-1} \mat{D}^{-1} + \mat{A}_{I^a} \mat{A}^{\dagger} \tilde{\mat{U}} \mat{D} (\mat{P}^{\top} \mat{P})^{-1} \mat{Z}^{\top} \mat{K}_1^{-1} \mat{D}^{-1}) \mat{L}.
\end{align*}
Now consider $\|\mat{W}^{\text{\text{sc}}}_{\text{c}} \mat{X}_{\text{tst}}\|_F^2 = \Tr(\mat{X}_{\text{tst}}^{\top} (\mat{W}_{\text{c}}^{\text{\text{sc}}})^{\top} \mat{W}_{\text{c}}^{\text{\text{sc}}} \mat{X}_{\text{\text{tst}}}).$ This is expanded as follows.
\begin{align*}
    &\Tr(\mat{D}^{-1} \mat{K}_1^{-1}( \mat{H} (\mat{A}^{\dagger})^{\top} \mat{A}_{I^a}^{\top} \mat{A}_{I^a} \mat{A}^{\dagger} \mat{H}^\top + \mat{H} (\mat{A}^{\dagger})^{\top} \mat{A}_{I^a}^{\top} \mat{A}_{I^a} \mat{A}^{\dagger} \tilde{\mat{U}}  \mat{D} (\mat{P}^{\top} \mat{P})^{-1} \mat{Z}^{\top} \\
    & + \mat{Z} (\mat{P}^{\top}\mat{P})^{-1} \mat{D} \tilde{\mat{U}}^{\top} (\mat{A}^{\dagger})^{\top} \mat{A}_{I^a}^{\top} \mat{A}_{I^a} \mat{A}^{\dagger} \mat{H}^\top \\ 
    &+ \mat{Z} (\mat{P}^{\top}\mat{P})^{-1} \mat{D} \tilde{\mat{U}}^{\top} (\mat{A}^{\dagger})^{\top} \mat{A}_{I^a}^{\top} \mat{A}_{I^a} \mat{A}^{\dagger} \tilde{\mat{U}}  \mat{D} (\mat{P}^{\top} \mat{P})^{-1} \mat{Z}^{\top} ) \mat{K}_1^{-1} \mat{D}^{-1})
\end{align*}
Note that the second and third term have mean of $0$ due to Lemma \ref{appendix:section3:with_skip:lma4}. 
Thus both terms are $0$, and since their element-wise variances are of order $O(d^{-1})$, the error introduced by this approximation is also $O(d^{-1})$. 
According to Lemma~\ref{appendix:section3:with_skip:lma2} and Lemma 8 of~\cite{kausik2024doubledescentoverfittingnoisy},
the first term has mean $\eta_{\text{trn}}^4 \frac{|I^a|}{n}\frac{c - 1}{c} (\eta_{\text{trn}}^2 \mat{D}^{-2} + \mathbb{I}_r)^{-2} \mat{D}^{-2}$, with the element-wise variance of 
$O(d^{-1})$. Similarly, the fourth term has mean $\eta_{\text{trn}}^4 \frac{|I^a|}{d} \mat{D}^{-4} (\eta_{\text{trn}}^2 \mat{D}^{-2} + \mathbb{I}_r)^{-2}$ with  element-wise variance of $O(d^{-1})$. This is due to Lemma \ref{appendix:section3:with_skip:lma5}, and Lemmas 6, 7, 8 of \cite{kausik2024doubledescentoverfittingnoisy}.
Organizing the terms, we have that 
\begin{align*}
    &\mathbb{E}\|\mat{W}^{\text{\text{sc}}}_{\text{c}} \mat{X}_{\text{\text{tst}}}\|_F^2 \\
    &= \frac{|I^a|}{d} \Tr\left(((c - 1) \mat{D}^2 + c \mathbb{I}_d)(c \mathbb{I}_r + c \eta_{\text{trn}}^{-2} \mat{D}^2)^{-2} \mat{L}\mat{L}^{\top}\right) + O(d^{-1}).
\end{align*}

\rightline{$\square$}

Substituting Lemmas~\ref{appendix:section3:with_skip:lma6}, ~\ref{appendix:section3:with_skip:lma7}, and \ref{appendix:section3:with_skip:lma8} into the decomposition in Equation~\eqref{appendix:decomposition_woskip} yields the desired result.

\subsection{Bias-Variance Decomposition}\label{appendix:section3:bias_variance_decomposition}
For the model with skip connection, we defined the term $\|\mat{X}_{\text{tst}} - \mat{W}_c \mat{X}_{\text{tst}}\|_F^2$ as the bias term, and $N_{\text{tst}}^{-1}\|\mat{W}^{\text{sc}}_{c}\|_F^2$ as the variance term in Section~\ref{section:generalization}. From Lemma~\ref{appendix:section3:wo_skip:lma4}, the bias term is indeed asymptotically equal to $\frac{1}{N_{\text{tst}}} \Tr(\mat{J}\mat{L} \mat{L}^T)$. On the other hand, Lemma~\ref{appendix:section3:wo_skip:lma3} shows that the variance term satisfies $\frac{\eta_{\text{tst}}^2}{d}\mathbb{E}[\|\mat{W}_c\|_F^2] \approx \frac{\eta_{\text{tst}}^2}{d} \frac{c}{c - 1} \sum\limits_{j \in I^x} \frac{\sigma_j^2}{\eta_{\text{trn}}^2 + \sigma_j^2}$.
These bias and variance expressions exhibit a trade-off behavior as the bottleneck dimension varies, closely resembling the classical bias–variance relationship. Motivated by this observation, we propose the following definition.

\begin{definition}\textnormal{(Bias and Variance in Two-Layer Linear DAEs).} \label{appendix:bias_variance_definition}
The bias term in the under-complete linear DAE
is defined as the component of the test error that \textbf{decreases} as the model complexity
(i.e., the bottleneck dimension k) increases. Conversely, the variance term is defined
as the component of the test error that \textbf{increases} as the model complexity grows.
\end{definition}

Unlike the model without skip connections, the skip-connected model does not admit a clear decomposition that allows for straightforward interpretation. Nevertheless, in Remark~\ref{remark:variance_skip}, we defined $\|\mat{W}_{c}^{\text{sc}}\|_F^2$ as a variance term, following the definition provided in Definition~\ref{appendix:bias_variance_definition}. Lemma~\ref{appendix:section3:with_skip:lma6} supports this interpretation by showing that this quantity captures the variance behavior described therein. Especially, among the various variance contributions, the term involving the $(c - 1)^{-1}$ factor becomes dominant as $c$ gets closer to $1$, and this definition includes this term.

\paragraph{On the Bias Term of the Model with Skip Connections}
We have seen that the bias term of the skip connection model includes $\eta_{\text{tst}}^2$, which is relatively large compared to 
 the model without a skip connection, unless we have a very high signal-to-noise ratio (low $\eta_{\text{tst}}$). The decomposition early in this subsection (\ref{skip_decomposition}) shows that the constant $\eta_{\text{tst}}^2$ originates from 
 $$\frac{1}{N_{\text{tst}}}\mathbb{E}_{\mat{A}_{\text{tst}}}\left[ \Tr(\mat{A}_{\text{tst}} \mat{A}_{\text{tst}}^T) \right].$$ 
 This occurs because incorporating a skip connection in two-layer model 
 makes the target changes from low-rank target $\mat{X}_{\text{tst}}$ to full-rank noisy target $\mat{A}_{\text{tst}}$.
 In contrast, the model we consider has fixed rank budget $k$. We believe this is due to of the limitation of having skip connection in two-layer linear models. 
 Things could be different, for instance, in \textit{four-layer linear} models, 
 with the skip connection exists between the two hidden layers in the middle. 
 In this case, the target is no longer the full rank noisy matrix $A_{\text{tst}}$. To illustrate this, let us examine the decomposition of this model.
 For a skip connection between the middle hidden layers, the network structure is defined as 
 $$\mat{W} := \mat{W}_4(\mat{W}_3\mat{W}_2 + \mathbb{I})\mat{W}_1.$$
Then, the decomposition of the test metric for a four-layer linear model is given by
 \begin{align*}
     &\frac{1}{N_{\text{tst}}}\mathbb{E}\left[ \|\mat{X}_{\text{tst}} - \mat{W}_4(\mat{W}_3\mat{W}_2 + \mathbb{I})\mat{W}_1(\mat{X}_{\text{tst}} + \mat{A}_{\text{tst}})\|_F^2 \right] \\
     &= \frac{1}{N_{\text{tst}}}(\|\mat{X}_{\text{tst}}\|_F^2 - 2 \Tr(\mat{W}\mat{X}_{\text{tst}}\mat{X}_{\text{tst}}^T) + \frac{\eta_{\text{tst}}^2 N_{\text{tst}}}{d} \|\mat{W}\|_F^2).
 \end{align*}
Now there is no full-rank noisy target $\mat{A}_{\text{tst}}$ contributing to the constant $\eta_{\text{tst}}^2$. Thus, there will be no large bias term 
 arising from this.
 It is left to future work to characterize the exact bias term for this model and compare it with the two-layer linear models with skip connections.

\section{Proofs and Supporting Results for Section~\ref{section:rmt}}\label{appendix:rmt_results}
In this section, we first identify the eigenvalue locations of the model described in Definition~\ref{def:rank1_add_model}, which implies that its eigenvalue distribution follows the Marchenko--Pastur law (Theorem \ref{appendix:rmt_background:mp_law}). We then show that the eigenvalue distribution of \textit{information-plus-noise} model, which is used throughout the paper, follows also the Marchenko--Pastur. These results suggest that the intuition developed from the simplified model (Definition~\ref{def:rank1_add_model}) may extend to the information-plus-noise setting. This analysis demonstrates that the peak observed near $c \approx 1$ arises from the accumulation of small eigenvalues near zero, whose number increases as $c \rightarrow 1$. This supports the argument presented in  Section~\ref{section:rmt}. Finally, in Subsection~\ref{appendix:rmt_results:main_proof}, we provide the proof of Theorem~\ref{theorem:alignment} in the main text.

\subsection{Understanding the Peak Near $c \approx 1$}
First, we identify the location of eigenvalues of $\mat{S}$ that was described in Definition \ref{def:rank1_add_model}. Recall that $\lambda_1, \vec{u}_1$ denotes the eigenvalue and eigenvector of the rank-1 $\mat{X}\mat{X}^T$, respectively.
\begin{lemma}\label{appendix:section4:lma1}
    \textnormal{(Location of Eigenvalues for Rank-1 Additive Model).}
    Let $\lambda^{\mat{S}}$ be any non-zero eigenvalue of $\mat{S}$. Then, it satisfies that $\vec{u}_1^T \mat{Q}_\mat{A}(\lambda^\mat{S}) \vec{u}_1 = - \frac{1}{\lambda_1}$. Furthermore, let $\lambda_m^{\mat{S}}$ be a $m$-th eigenvalue of $\mat{S}$, for $m \in \{2, \dots, n\}$. Then, $\lambda_m^{\mat{S}} \in (\lambda_m^{\mat{A}}, \lambda_{m - 1}^{\mat{A}})$.
\end{lemma}

\textit{Proof.\quad} To identify the eigenvalue information, we observe $det(\mat{S} - \alpha \mathbb{I}_d)$.
For $\lambda^{\mat{S}}_j, j \in \{1, \dots, d \}$, we now derive an equivalent condition for $det(\mat{S} - \lambda^{\mat{S}}_j \mathbb{I}_d) = 0$. 
Observe that,
\begin{align*}
    &det(\mat{S} - \lambda^{\mat{S}} \mathbb{I}_d) \\
    &= det(\mat{XX^T} + \mat{A}\mat{A}^T - \lambda^{\mat{S}} \mathbb{I}_d) \\
    &= det(\mat{Q}_{\mat{A}}^{-1})det(\mathbb{I} + \lambda_1 \mat{Q}_\mat{A} \vec{u}_1 \vec{u}_1^T) \\
    &= det(\mat{Q}_\mat{A}^{-1})det(1 + \lambda_1 \vec{u}_1^T \mat{Q}_\mat{A} \vec{u}_1).
\end{align*}
The last equality is due to Lemma \ref{appendix:rmt_background:lemmata}-(5).
Thus we have the equivalent condition that,
\begin{equation}\label{appendix:rmt_results:determinant_scm}
    det(\mat{Q}_\mat{A}^{-1})det(1 + \lambda_1 \vec{u}_1^T \mat{Q}_\mat{A} \vec{u}_1) = 0.
\end{equation}
Thus we have either $det\left(\mat{Q}_\mat{A}^{-1}(\lambda^\mat{S}_j)\right) = 0$, or 
$det\left(1 + \lambda_1 \vec{u}_1^T \mat{Q}_\mat{A}(\lambda^\mat{S}) \vec{u}_1\right) = 0$. The former one is not possible, as 
$\lambda^\mat{S}_j$ is not an eigenvalue of $\mat{A}\mat{A}^T$. 
The latter one is equivalent to finding $\alpha \in \mathbb{R}_+ \backslash \{\lambda^\mat{A}_i, \dots, \lambda^\mat{A}_d\}$,
 such that $\vec{u}_1^T \mat{Q}_\mat{A}(\alpha) \vec{u}_1 = - \frac{1}{\lambda_1}$. Note that $f(\alpha) := \vec{u}_1^T \mat{Q}_\mat{A}(\alpha) \vec{u}_1$ is increasing in every interval of 
 $(\lambda^\mat{A}_{k}, \lambda^\mat{A}_{k - 1})$, for $k \in \{2, \dots, n\}$, as $f'(\alpha) := \vec{u}_1^T \mat{Q}_\mat{A}^2(\alpha) \vec{u}_1 > 0$. In addition to this,
 $\lim\limits_{\alpha \downarrow \lambda^A_{k}} f(\alpha) = - \infty, \lim\limits_{\alpha \uparrow \lambda^A_{k - 1}} f(\alpha) = \infty$, thus 
 this $f(\alpha)$ is monotonically increasing from $-\infty$ to $\infty$. This means that we have eigenvalue $\lambda^\mat{S}_k$ inside every interval 
 $(\lambda^\mat{A}_{k}, \lambda^\mat{A}_{k - 1})$, $\forall k \in \{2, \dots, n\}$. Thus we have exactly $n$ eigenvalues, including $\lambda^\mat{S}_1 \in (\lambda_{1}, \infty)$. 

\rightline{$\square$}
This result shows that the empirical spectral distributions of $\mat{S}$ and $\mat{A}\mat{A}^\top$ are essentially identical in the limit.
A similar property holds for the \textit{information-plus-noise} model: the empirical eigenvalue distribution converges weakly to the Marchenko–Pastur law. 
This is formalized in the following theorem.

\begin{theorem}\label{appendix:section4:thm1}
    \textnormal{($\mu_{N}$ converges weakly to $\mu_{MP}$).} \\
    For $\alpha \in \mathbb{C} \backslash \mathbb{R}_+$, let $\mu_{N}$ be the empirical spectral measure (Def \ref{appendix:rmt_background:definitions}) of $(\mat{X} + \mat{A})(\mat{X} + \mat{A})^T$, 
    for $\mat{X}$ and $\mat{A}$ satisfying Assumption~\ref{data_assumption}.
    In addition to this, assume that for some constant $C_1 > 0$, 
    it satisfies that $\|\vec{x}_i \|_2 \leq \frac{C_1}{\sqrt{N}}$, for $i \in \{ 1, \dots, N\}$.
    Let $\mu_{MP}$ be a version of Marchenko-Pastur distribution, where 
    \begin{align*}
        \mu_{MP}(\alpha) = \begin{cases}
            \frac{\sqrt{(c \alpha - \eta^2 (\sqrt{c} - 1)^2)(\eta^2 (\sqrt{c} + 1)^2 - c \alpha)}}{2 \pi \alpha c \eta^2} & \text{if } \alpha \in [\frac{\eta^2}{c}(\sqrt{c} - 1)^2, \frac{\eta^2}{c}(\sqrt{c} + 1)^2] \\
            1 - \frac{1}{c}  & \text{else}.
        \end{cases}
    \end{align*}
    Then, $\mu_{N}$ converges weakly to $\mu_{MP}$.
\end{theorem}

Note that the assumption regarding the norm of the columns of $\mat{X}$ is natural, 
given that we have assumed $\|\mat{X}\|_2$ scales as $\Theta(1)$. 
This implies the Frobenius norm of $X$ also scales as $\Theta(1)$, since $\mat{X}$ has fixed rank $r$. 
If each data point $\vec{x}_i$ scales at the same rate, 
then each individual data point would scale as $\Theta(N^{-1 \slash 2})$. Thus, this assumption is equivalent to stating that 
\textit{each data point $\mathbf{x}_i$ follows the same scaling behavior}. \\
 
\textit{Proof.\quad}
It is sufficient to show $m_{\mu_N} \xrightarrow{a.s} m_{\mu_{MP}}$ to reach the conclusion, 
based on the fact of that $\mathbb{P}(\mu_{N} \rightarrow \mu_{MP}\, \text{weakly}) = 1 \Leftrightarrow m_{\mu_N} \xrightarrow{a.s} m_{\mu_{MP}}$ \cite[Exercise 2.4.10]{tao2012topics}.
To establish this convergence, we first show that $m_{\mu_N} \xrightarrow{a.s} \mathbb{E}[m_{\mu_N}]$ and then show $\mathbb{E}[m_{\mu_N}] \xrightarrow{a.s} m_{\mu_{MP}}$, 
to conclude $m_{\mu_N} \xrightarrow{a.s} m_{\mu_{MP}}$. \\
Firstly, in order to show $m_{\mu_N} \xrightarrow{a.s} \mathbb{E}[m_{\mu_N}]$, 
we follow the standard approach outlined in Lemma 2.12 and 2.13 of \cite{bai2010spectral}.
In essence, we first construct a Martingale Difference Sequence to show that $\frac{1}{d}\Tr(\mat{Q}_{-i})$ converges to $\frac{1}{d}\Tr(\mat{Q})$ almost surely.
Then, we use Lemma 2.12 \cite{bai2010spectral} to find an upper bound and finish with the Borel-Cantelli Lemma to establish the almost sure convergence.
A key step in the proof is demonstrating that $\left|\frac{1}{d}\Tr(\mat{Q}_{-i}) - \frac{1}{d}\Tr(\mat{Q}) \right|$ is sufficiently small. 
We cannot directly use Lemma \ref{appendix:rmt_background:lemmata}-(4) however, 
since $\mat{Q} = (\mat{Z}\mat{Z}^T - \alpha \mathbb{I}_d)^{-1}$, and $\vec{z}_i = \vec{x}_i + \vec{a}_i$ is not just a mean-zero gaussian.
Nonetheless, leveraging the \textit{low-rank} property of $\mat{X}$, we can still show that this term remains small, as stated in the following lemma.
\begin{lemma}\label{appendix:section4:lma2}
    Consider the setting of Theorem \ref{appendix:section4:thm1}. Let $\mat{Q} := (\mat{Z}\mat{Z}^T - \alpha \mathbb{I}_d)^{-1}$, where $\mat{Z} = \mat{X} + \mat{A}$.
    Then,
    $\left|\frac{1}{d} \Tr(\mat{Q}) - \frac{1}{d} \Tr(\mat{Q}_{-j})\right| = O(n^{-1})$.
\end{lemma}

\textit{Proof of Lemma \ref{appendix:section4:lma2}.\quad}
Due to the Sherman-Morrison Lemma, (Lemma \ref{appendix:rmt_background:lemmata}-(2)), we have that $\mat{Q} = \mat{Q}_{-j} - \frac{\mat{Q}_{-j}\vec{z}_j \vec{z}_j^T \mat{Q}_{-j}}{1 + \vec{z}_j^T \mat{Q}_{-j} \vec{z}_j}$. 
Thus $\frac{1}{d} \Tr(\mat{Q}) - \frac{1}{d} \Tr(\mat{Q}_{-j}) = - \frac{1}{d} \Tr\left(\frac{\vec{z}_j^T \mat{Q}_{-j}^2 \vec{z}}{1 + \vec{z}_j^T \mat{Q}_{-j} \vec{z}}\right)$. 
Note that $\vec{z}_j^T \mat{Q}_{-j} \vec{z}_j = \vec{x}_j^T \mat{Q}_{-j} \vec{x}_j + \vec{x}_j^T \mat{Q}_{-j} \vec{a}_j + \vec{a}_j^T \mat{Q}_{-j} \vec{x}_j + \vec{a}_j^T \mat{Q}_{-j} \vec{a}_j$. 
The first term is $\vec{x}^T \mat{Q}_{-j} \vec{x}_j \leq \|\vec{x}_j\|_2^2 \|\mat{Q}_{-j}\|_2 = O(n^{-1})$, as $\|\mat{Q}_{-j}\|_2$ is bounded and $\|\vec{x}_j\|_2 = O(\frac{1}{\sqrt{N}})$.
$\vec{x}_j^T \mat{Q}_{-j} \vec{a}_j$ has mean of $0$, and variance is $\mathbb{E}\left[\vec{x}_j^T \mat{Q}_{-j} \vec{a}_j \vec{a}_j^T \mat{Q}_{-j} \vec{x}_j\right] \leq \frac{\eta^2}{d} \|\vec{x}_j\|_2^2 \|\mat{Q}_{-j}\|_2^2 = O(n^{-2})$. 
Thus applying the Borel-Cantelli lemma will give us that $\vec{x}_j^T \mat{Q}_{-j} \vec{a}_j \xrightarrow{a.s} 0$. 
Therefore, we have that $\vec{z}_j^T \mat{Q}_{-j} \vec{z}_j \xrightarrow{a.s} \vec{a}_j^T \mat{Q}_{-j} \vec{a}_j$. With this, we conclude that
\begin{align*}
    \left|\frac{1}{d} \Tr(\mat{Q}) - \frac{1}{d} \Tr(\mat{Q}_{-j})\right| &= \left|\frac{1}{d} \Tr\left(\frac{\vec{z}_j^T \mat{Q}_{-j}^2 \vec{z}}{1 + \vec{z}_j^T \mat{Q}_{-j} \vec{z}}\right)\right| \\
    &\simeq \left|\frac{1}{d} \Tr\left(\frac{\vec{a}_j^T \mat{Q}_{-j}^2 \vec{a}_j}{1 + \vec{a}_j^T \mat{Q}_{-j} \vec{a}_j}\right)\right| \\
    &\stackrel{\ref{appendix:rmt_background:lemmata}.(3)}{\simeq} \left|\frac{1}{d} \Tr\left(\frac{\frac{\eta^2}{d}\Tr(\mat{Q}_{-j}^2)}{1 + \frac{\eta^2}{d} \Tr(\mat{Q}_{-j})}\right)\right| \\
    &= O(d^{-1}) = O(n^{-1}).
\end{align*}

\rightline{$\square$}

Now back to the original proof, recall that $m_{\mu_N} = \frac{1}{d} \Tr(\mat{Q})$. 
Then, we construct the martigale difference sequence, which is
$$
m_{\mu_N} - \mathbb{E}[m_{\mu_N}] = \sum\limits_{j = 1}^N \left(\mathbb{E}_j \left[\frac{1}{d} \Tr(\mat{Q})\right] - \mathbb{E}_{j - 1}\left[\frac{1}{d} \Tr(\mat{Q})\right]\right),
$$ 
for $\mathbb{E}_j\left[\frac{1}{d} \Tr(\mat{Q})\right] := \mathbb{E}\left[\frac{1}{d} \Tr(\mat{Q});\vec{z}_1, \dots, \vec{z}_j\right]$, and $\mathbb{E}_0[m_{\mu_N}] := \mu_{N}$. 
This is by construction a martingale difference sequence, since
$$
\mathbb{E}\left[\left(\mathbb{E}_j - \mathbb{E}_{j - 1}\right)\left[\frac{1}{d}\Tr(\mat{Q})\right] ; \vec{z}_1, \dots, \vec{z}_{j - 1}\right] = 0.
$$ 
This is due to the fact that $\mathbb{E}\left[\mathbb{E}_j\left[\frac{1}{d}\Tr(\mat{Q})\right] ; \vec{z}_1, \dots, \vec{z}_{j - 1}\right] = \mathbb{E}_{j - 1}\left[\frac{1}{d}\Tr(\mat{Q})\right]$ (\cite[Theorem 4.1.13]{durrett2019probability}). \\
Now, observe that $\mathbb{E}_j\left[\frac{1}{d} \Tr(\mat{Q}_{-j})\right] = \mathbb{E}_{j - 1}\left[\frac{1}{d} \Tr(\mat{Q}_{-j})\right]$, 
then we have 
\begin{align*}
    &\sum\limits_{j = 1}^N \left(\mathbb{E}_j \left[\frac{1}{d} \Tr(\mat{Q})\right] - \mathbb{E}_{j - 1}\left[\frac{1}{d} \Tr(\mat{Q})\right]\right) \\
    &= \sum\limits_{j = 1}^N \left(\mathbb{E}_j \left[\frac{1}{d} \Tr(\mat{Q}) - \frac{1}{d} \Tr(\mat{Q}_{-j})\right] - \mathbb{E}_{j - 1}\left[\frac{1}{d} \Tr(\mat{Q}) - \frac{1}{d} \Tr(\mat{Q}_{-j})\right]\right). 
\end{align*}
With Lemma \ref{appendix:section4:lma2}, 
we have that $\left(\mathbb{E}_j - \mathbb{E}_{j - 1}\right)\left[\frac{1}{d} \Tr(\mat{Q})\right] = O(n^{-1})$. 
Applying \cite[Lemma 2.12]{bai2010spectral}, for some constant $K_2 > 0$, we have that
\begin{align*}
    \mathbb{E}\left[\left|m_{\mu_N} - \mathbb{E}[m_{\mu_N}]\right|^4\right] &= \mathbb{E}\left[\sum\limits_{j=1}^N (\mathbb{E}_j - \mathbb{E}_{j - 1})\left[\frac{1}{d} \Tr(\mat{Q}) - \frac{1}{d} \Tr(\mat{Q}_{-i})\right]\right] \\
    &\leq K_2 \mathbb{E}\left[\left(\sum\limits_{j=1}^N \left|(\mathbb{E}_j - \mathbb{E}_{j - 1})\left[\frac{1}{d} \Tr(\mat{Q}) - \frac{1}{d} \Tr(\mat{Q}_{-i})\right] \right|^2 \right)^2\right] \\
    &= O(n^{-2}).
\end{align*}
It follows that, for any $\epsilon > 0$,
\begin{align*}
    \mathbb{P}(\left|m_{\mu_N} - \mathbb{E}[m_{\mu_N}]\right| > \epsilon) &\leq \frac{\mathbb{E}\left[m_{\mu_N} - \mathbb{E}[m_{\mu_N}]|^4\right]}{\epsilon^4} \\
    &= O(n^{-2}).
\end{align*}
Applying the standard Borel-Cantelli lemma, we obtain $m_{\mu_N} \xrightarrow{a.s} \mathbb{E}[m_{\mu_N}]$. 
This completes the first step. From this point onward, we denote $m := \mathbb{E}[m_{\mu_N}]$. 

Since we have established the convergence of $m_{\mu_N} \xrightarrow{a.s} m$, our goal is now to show $m \xrightarrow{a.s} m_{\mu_{MP}}$. 
For this, the key idea is to find a fixed-point equation, leveraging the close asymptotical relationship between $\mat{Q}$ and $\mat{Q}_{-j}$.
From $\mat{Q} = \mat{Q}_{-j} - \frac{\mat{Q}_{-j}\vec{z}_j \vec{z}_j^T \mat{Q}_{-j}}{1 + \vec{z}_j^T \mat{Q}_{-j} \vec{z}_j}$ (Lemma~\ref{appendix:rmt_background:lemmata}-(2)), 
it holds that $\vec{z}_j^T \mat{Q} \vec{z}_j = \frac{\vec{z}_j^T \mat{Q}_{-j} \vec{z}_j}{1 + \vec{z}_j^T \mat{Q}_{-j} \vec{z}_j}$. 
From the proof of Lemma \ref{appendix:section4:lma2}, we already established that 
$$\vec{z}_j^T \mat{Q} \vec{z}_j \xrightarrow{a.s} \vec{a}_j^T \mat{Q}_{-j} \vec{a}_j.$$ 
Using this result, it satisfies that
$\vec{z}_j^T \mat{Q} \vec{z}_j \simeq \frac{\eta^2}{d} \frac{\Tr(\mat{Q}_{-j})}{1 + \frac{\eta^2}{d} \Tr(\mat{Q}_{-j})} \simeq \frac{\eta^2}{d} \frac{\Tr(\mat{Q})}{1 + \frac{\eta^2}{d} \Tr(\mat{Q})}$. 
Since this holds for all $j \in [n]$, it follows that $\sum\limits_{j = 1}^n \vec{z}_j^T Q \vec{z}_j \simeq n \frac{\eta^2}{d} \frac{\Tr(\mat{Q})}{1 + \frac{\eta^2}{d} \Tr(\mat{Q})}$.
Using the identity $\sum\limits_{j = 1}^n \vec{z}_j^T \mat{Q} \vec{z}_j = \Tr(\mat{Z}\mat{Z}^T \mat{Q})$, and the fact that 
$\mat{Z}\mat{Z}^T = (\mat{Q}^{-1} + \alpha \mathbb{I}_d)$, we obtain $\sum\limits_{j = 1}^n \vec{z}_j^T \mat{Q} \vec{z}_j = \Tr(\mathbb{I}_d + \alpha \mat{Q})$. 
Thus, we arrive at $d + \alpha \Tr(\mat{Q}) \simeq n \frac{\eta^2}{d} \frac{\Tr(\mat{Q})}{1 + \frac{\eta^2}{d} \Tr(\mat{Q})}$. In the limit case where $d, n \rightarrow \infty$, 
this simplifies to the fixed point equation 
 $$1 + \alpha m = \frac{\eta^2 c m}{1 + \eta^2 m}$$, 
 for $c = \frac{d}{n}$. 
 After rearranging, we obtain the quadratic equaion
\begin{align*}
    \alpha c \eta^2 m^2 + (\alpha c + c \eta^2 - \eta^2)m + c = 0.
\end{align*}
This is precisely the quadratic equation for $m_{\mu_{MP}}$, which proves $m \xrightarrow{a.s} m_{\mu_{MP}}$. 
To see this more clearly, note that
\begin{align*}
    m(\alpha) = \frac{\eta^2 - c \eta^2 - \alpha c}{2 \alpha c \eta^2} \pm \frac{\sqrt{(c \alpha - \eta^2 (\sqrt{c} - 1)^2)(c \alpha - \eta^2 (\sqrt{c} + 1)^2 )}}{2 \alpha c \eta^2}.
\end{align*}
Using the Inverse Stieltjes Transform(See \cite[Theorem 2.4]{couillet_liao_2022}), we find that 
for all $\alpha \in \mathbb{C} \backslash \{0\}$,
\begin{align*}
    &\mu_{MP}(\alpha) = \frac{1}{\pi} \lim\limits_{\epsilon \rightarrow 0} Im(m (\alpha + i \epsilon)) \\
    &= \frac{\sqrt{(c \alpha - \eta^2 (\sqrt{c} - 1)^2)(c \alpha - \eta^2 (\sqrt{c} + 1)^2 )}}{2 \pi \alpha c \eta^2} \text{, for } \alpha \in \left[\frac{\eta^2}{c}(\sqrt{c} - 1)^2, \frac{\eta^2}{c}(\sqrt{c} + 1)^2\right].
\end{align*}
For $\alpha = 0$, we have that
\begin{align*}
    \mu_{MP}(\{0\}) &= - \lim\limits_{\epsilon \downarrow 0} i \epsilon m(i \epsilon) \\
    &= \begin{cases}
        0 & \text{if } c < 1 \\
        1 - \frac{1}{c} & \text{if } c \geq 1.
    \end{cases}.
\end{align*}
Therefore $m \xrightarrow{a.s} m_{\mu_{MP}}$. Then, $m_{\mu_N} \xrightarrow{a.s} m_{\mu_{MP}}$ follows and this leads to $\mu_{N} \xrightarrow{a.s} \mu_{MP}$,
which concludes the proof.

\rightline{$\square$}

\subsection{Proof of Theorem \ref{theorem:alignment}}\label{appendix:rmt_results:main_proof}

We begin by establishing the condition for the location of eigenvalues of $\mat{S}$. Then, we apply the Cauchy integral formula as introduced in Subsection~\ref{appendix:rmt_background:subspace_information}.
We denote the resolvent of $\mat{S}$ as $\mat{Q}_\mat{S}(\alpha) = (\mat{S} - \alpha \mathbb{I}_d)^{-1}$ 
and the resolvent of $\mat{A}\mat{A}^T$ as $\mat{Q}_\mat{A}(\alpha) = (\mat{A}\mat{A}^T - \alpha \mathbb{I}_d)^{-1}$. 

We aim to analyze the following quantity, 
where $\Gamma_{\lambda^\mat{S}_j}$ is a closed, positive oriented contour that \textit{only} encompasses $\lambda^\mat{S}_j$.
\begin{equation}
\label{chap:understand_peak:further_arguments:main_object_scm_no_skip}
    \langle \vec{u}_1, \vec{u}^\mat{S}_j \rangle^2 = - \frac{1}{2 \pi i} \int_{\Gamma_{\lambda^\mat{S}_j}} \vec{u}_1^T \mat{Q}_{\mat{S}} \vec{u}_1 \,d\alpha.
\end{equation}
We first pull $\mat{Q}_{\mat{A}}$ out of $\mat{Q}_{\mat{S}}$ using the Woodbury Identity (Lemma \ref{appendix:rmt_background:lemmata}-(6)). 
\begin{align*}
    \mat{Q}_{\mat{S}} &= (\mat{Q}_\mat{A}^{-1} + \lambda_1 \vec{u}_1 \vec{u}_1^T)^{-1} \\
    &= \mat{Q}_{\mat{A}} - \frac{\lambda_1}{1 + \vec{u}_1^T \mat{Q}_{\mat{A}} \vec{u}_1} \mat{Q}_{\mat{A}} \vec{u}_1 \vec{u}_1^T \mat{Q}_{\mat{A}}.
\end{align*}
With this, we have that
\begin{align*}
    \langle \vec{u}_1, \vec{u}^\mat{S}_j \rangle^2 &= - \frac{1}{2 \pi i} \int_{\Gamma_{\lambda^\mat{S}_j}} \vec{u}_1^T \mat{Q}_{\mat{S}} \vec{u}_1 \,d\alpha \\
    &= - \frac{1}{2 \pi i} \int_{\Gamma_{\lambda^\mat{S}_j}} \vec{u}_1^T \mat{Q}_{\mat{A}} \vec{u}_1 \,d\alpha \, + \\
    & \frac{1}{2 \pi i} \int_{\Gamma_{\lambda^\mat{S}_j}} \frac{1}{1 + \vec{u}_1^T \mat{Q}_{\mat{A}} \vec{u}_1} \vec{u}_1^T \mat{Q}_{\mat{A}} \vec{u}_1 \vec{u}_1^T \mat{Q}_{\mat{A}} \vec{u}_1 \,d\alpha.
\end{align*}
Note that the first integral is $0$, since there is no singularity inside the contour. 
For the second integral, we have the singularity at $\lambda^\mat{S}_j$ from Lemma \ref{appendix:section4:lma1}.
Then, using the residue calculus, it follows that
\begin{align*}
    &\frac{1}{2 \pi i} \int_{\Gamma_{\lambda^\mat{S}_j}} \frac{1}{1 + \vec{u}_1^T \mat{Q}_{\mat{A}} \vec{u}_1} \vec{u}_1^T \mat{Q}_{\mat{A}} \vec{u}_1 \vec{u}_1^T \mat{Q}_{\mat{A}} \vec{u}_1 \,d\alpha \\
    &= \lim\limits_{\alpha \rightarrow \lambda^\mat{S}_j} (\alpha - \lambda^\mat{S}_j) \frac{1}{1 + \vec{u}_1^T \mat{Q}_{\mat{A}} \vec{u}_1} \vec{u}_1^T \mat{Q}_{\mat{A}} \vec{u}_1 \vec{u}_1^T \mat{Q}_{\mat{A}} \vec{u}_1 \\
    &= \frac{1}{\lambda_1^2} \lim\limits_{\alpha \rightarrow \lambda^\mat{S}_j} (\alpha - \lambda^\mat{S}_j) \frac{1}{1 + \vec{u}_1^T \mat{Q}_{\mat{A}} \vec{u}_1}.
\end{align*}
The last equality is due to Lemma \ref{appendix:section4:lma1}. \\
We denote $f(\alpha) := \vec{u}_1^T \mat{Q}_{\mat{A}}(\alpha) \vec{u}_1$, then we have that 
\begin{align*}
    \lim\limits_{\alpha \rightarrow \lambda^\mat{S}_j} (\alpha - \lambda^\mat{S}_j) \frac{1}{1 + \vec{u}_1^T \mat{Q}_{\mat{A}} \vec{u}_1} = \frac{1}{f'(\lambda^\mat{S}_j)}.
\end{align*}
Therefore, we conclude that
\begin{align}
    \langle \vec{u}_1, \vec{u}^\mat{S}_j \rangle^2 = \frac{1}{\lambda_1^2 f'(\lambda^\mat{S}_j)}.
\end{align}
Now we work on $\langle \vec{u}^\mat{A}_i, \vec{u}^\mat{S}_j \rangle^2$. We are interested in:
\begin{equation}\label{chap:understand_peak:further_arguments:main_object_scm_skip}
    \langle \vec{u}^\mat{A}_i, \vec{u}^\mat{S}_j \rangle^2 = - \frac{1}{2 \pi i} \int_{\Gamma_{\lambda^\mat{S}_j}} (\vec{u}^\mat{A}_i)^T \mat{Q}_{\mat{S}} \vec{u}^\mat{A}_i \,d\alpha.
\end{equation}
Following the same steps as above, we have the non-trivial term:
\begin{align*}
    &\frac{1}{2 \pi i} \int_{\Gamma_{\lambda^\mat{S}_j}} \frac{1}{1 + \vec{u}_1^T \mat{Q}_{\mat{A}} \vec{u}_1} (\vec{u}^\mat{A}_i)^T \mat{Q}_{\mat{A}} \vec{u}_1 \vec{u}_1^T \mat{Q}_{\mat{A}} \vec{u}^\mat{A}_i \,d\alpha \\
    &= \lim\limits_{\alpha \rightarrow \lambda^\mat{S}_j} (\alpha - \lambda^\mat{S}_j) \frac{1}{1 + \vec{u}_1^T \mat{Q}_{\mat{A}} \vec{u}_1} (\vec{u}^\mat{A}_i)^T \mat{Q}_{\mat{A}} \vec{u}_1 \vec{u}_1^T \mat{Q}_{\mat{A}}\vec{u}^\mat{A}_i \\
    &= \frac{\langle \vec{u}^\mat{A}_i, \vec{u}_1\rangle^2}{(\lambda^\mat{A}_i - \lambda^\mat{S}_j)^2} \lim\limits_{\alpha \rightarrow \lambda^\mat{S}_j} (\alpha - \lambda^\mat{S}_j) \frac{1}{1 + \vec{u}_1^T \mat{Q}_{\mat{A}} \vec{u}_1} \\
    &= \frac{\langle \vec{u}^\mat{A}_i, \vec{u}_1\rangle^2}{(\lambda^\mat{A}_i - \lambda^\mat{S}_j)^2} \frac{1}{f'(\lambda^\mat{S}_j)}.
\end{align*}
Thus, we have the relative proportion of the alignments as:
\begin{align*}
    \frac{\langle \vec{u}^\mat{A}_i, \vec{u}^\mat{S}_j \rangle^2}{\langle \vec{u}_1, \vec{u}^\mat{S}_j \rangle^2} = \frac{\langle \vec{u}^\mat{A}_i, \vec{u}_1\rangle^2}{(\lambda^\mat{A}_1 - \lambda^\mat{S}_j)^2} \frac{1}{f'(\lambda^\mat{S}_j)} \lambda_1^2 f'(\lambda^\mat{S}_j) = \lambda_1^2 \frac{\langle \vec{u}^\mat{A}_i, \vec{u}_1\rangle^2}{(\lambda^\mat{A}_i - \lambda^\mat{S}_j)^2}.
\end{align*}
Note that from Lemma~\ref{appendix:section4:lma1}, for $j \in [2,n] \backslash \{i-1, i\}$, $(\lambda^\mat{A}_i - \lambda^\mat{S}_j)^2 = \Theta\left((\lambda^\mat{A}_i - \lambda^\mat{A}_j)^2\right)$. Thus 
$$
\frac{\langle \vec{u}^\mat{A}_i, \vec{u}^\mat{S}_j \rangle^2}{\langle \vec{u}_1, \vec{u}^\mat{S}_j \rangle^2} = \Theta\left(\lambda_1^2 \frac{\langle \vec{u}^\mat{A}_i, \vec{u}_1\rangle^2}{(\lambda^\mat{A}_i - \lambda^\mat{A}_j)^2}. \right)
$$


Due to the fact that $\vec{u}^\mat{A}_i$ is an uniform random vector,  
it follows from Lemma \ref{appendix:section3:with_skip:lma1}, that $\mathbb{E}\left[\langle \vec{u}^\mat{A}_i, \vec{u}_1 \rangle^2\right] = d^{-1}$. Moreover, under our assumptions, $\lambda_1 = \Theta(1)$. Using the fact that the eigenvectors of a Gaussian random matrix are independent of its eigenvalues, for $i \in [k]$ and $j \in [2,n] \backslash \{i-1, i\}$, we obtain 
$$
\mathbb{E}\left[\frac{\langle \vec{u}^\mat{A}_i, \vec{u}^\mat{S}_j \rangle^2}{\langle \vec{u}_1, \vec{u}^\mat{S}_j \rangle^2}\right] = \Theta\left(\frac{1}{d(\lambda_i^{\mat{A}} - \lambda_j^{\mat{A}})^2}\right) = \Theta\left(\frac{1}{d(\lambda_i^{\mat{A}} - \lambda_j^{\mat{S}})^2}\right).
$$
\rightline{$\square$}

\begin{remark}\textnormal{(The case of $j = 1$).}
    Note that the above result also applies for $j = 1$, since $\lambda_j^{\mat{S}}$ converges almost surely to some constant. This can be proven utilizing the tools introduced in \cite{baik2004eigenvalueslargesamplecovariance}. We omit the proof for the brevity.
\end{remark}

\begin{remark}\textnormal{(For $j$ that Corresponds to Small Eigenvalues and Their Influence on Variance).}
    As it was proved in Lemma~\ref{appendix:section4:lma1}, the location of eigenvalues of $\mat{S}$ follows that of $\mat{A}\mat{A}^\top$. According to \cite{rudelson2009smallest}, the smallest eigenvalues of $\mat{A}\mat{A}^\top$ scale $O(n^{-2})$,  and these terms dominate the variance contribution to $\|\mat{W}_c\|_F^2$. For eigenvectors corresponding to these small eigenvalues, which scale at the same rate,  the theorem becomes $\Theta(d^{-1})$, since $(\lambda_i^{\mat{A}} - \lambda_j^{\mat{S}})^2 = \Theta(1)$. Moreover, as $c \rightarrow 1$, the number of such small eigenvalues increases, further amplifying their influence on the variance.
\end{remark}

\section{Additional Details on Numerical Results}\label{appendix:experiments}
\subsection{Data and Test Setting}
We used the CIFAR-10 dataset \cite{krizhevsky2009learning} throughout the main text. Training and test data were sampled from disjoint splits. Each data point was reshaped into a $3072$-dimensional vector. The number of test samples was fixed at $4500$. Since the dataset has a fixed ambient dimension $d$, our numerical experiments focused on varying the number of training samples $n$.
To generate Figure~\ref{fig:compare_test_risk} and the left and center plots of Figure~\ref{fig:skip_connection}, we set the data rank to $r = 100$ and the bottleneck dimension to $k = 50$. To obtain low-rank representations, we performed singular value decomposition (SVD) and retained the top $r$ components.
All figures were produced using appropriately scaled data to ensure a signal-to-noise ratio of approximately $\frac{\|\mat{X}\|_2}{\|\mat{A}\|_2} \approx 30$, where $\|\cdot\|_2$ denotes the operator norm.

Although computing the test error as described in Eq.~(\ref{test_metric}) ideally requires multiple trials to reduce variance, we conducted only a single trial. The results demonstrated strong agreement with theoretical expectations, likely due to concentration effects.
Our numerical experiments were conducted using a T4 GPU on Google Colab. Generating Figure~\ref{fig:bottleneck} took approximately 5 hours, using a stride of $20$ and starting from $n = 2568$. Figure~\ref{fig:skip_connection} required approximately 2 hours to compute.

\subsection{Solutions from Existing Methods Used in Plots}
To set the clear line from our denoising setting
to other settings, in Figure~\ref{fig:compare_test_risk} we generated generalization error curves for other models in the overparameterized setting. For this, we used the regularized expressions for the critical points from Section~\ref{section:preliminaries}, and utilized the minimum-norm solutions. 
For the underparameterized solution depicted in Figure~\ref{fig:skip_connection}, we directly used the analytical solutions provided by \cite{baldi1989neural}, adapting them to our setting as needed.

\end{document}